\documentclass[journal,10pt]{IEEEtran}
\ifCLASSOPTIONcompsoc
\else
\fi
\ifCLASSINFOpdf
\else
\fi
\usepackage[ruled,vlined]{algorithm2e}
\usepackage{epsfig}
\usepackage{graphicx}
\usepackage{amsmath}
\usepackage{amssymb}
\usepackage[font=small]{caption,subfig}
\usepackage{subfig}
\usepackage{color}
\usepackage{nomencl}
\usepackage{relsize}
\usepackage{array}
\usepackage{url}
\usepackage{booktabs}
\usepackage{bigints}
\usepackage{multirow}
\newtheorem{theorem}{Theorem}[section]

\newtheorem{proposition}[theorem]{Proposition}

\newcommand{\qed}{\nobreak \ifvmode \relax \else
      \ifdim\lastskip<1.5em \hskip-\lastskip
      \hskip1.5em plus0em minus0.5em \fi \nobreak
      \vrule height0.75em width0.5em depth0.25em\fi}

\DeclareMathOperator*{\argmax}{argmax}
\DeclareMathOperator*{\argmin}{argmin}

\begin{document}
\title{Matching-Constrained Active Contours}

\author{Junyan~Wang*,~\IEEEmembership{Member,~IEEE,}
        Kap~Luk~Chan,~\IEEEmembership{Member,~IEEE}
\thanks{Junyan Wang and Kap Luk Chan are with School of Electrical \& Electronic Engineering, Nanyang Technological University, Singapore.\protect\\
E-mail: \{wa0009an,eklchan\}@ntu.edu.sg}
}


\markboth{}%
{}

%\IEEEpubid{\makebox[\columnwidth]{\hfill 0000--0000/00/\$00.00~\copyright~2007 IEEE}%


% PRIOR to the title within the \IEEEcompsoctitleabstractindextext IEEEtran
% command as these need to go into the title area created by
\maketitle
%\IEEEcompsoctitleabstractindextext{%
\begin{abstract}
%\boldmath
In object segmentation by active contours, the initial contour is often required. Conventionally, the initial contour is provided by the user. This paper extends the conventional active contour model by incorporating feature matching in the formulation, which gives rise to a novel matching-constrained active contour. The numerical solution to the new optimization model provides an automated framework of object segmentation without user intervention. The main idea is to incorporate feature point matching as a constraint in active contour models. To this effect, we obtain a mathematical model of interior points to boundary contour such that matching of interior feature points gives contour alignment, and we formulate the matching score as a constraint to active contour model such that the feature matching of maximum score that gives the contour alignment provides the initial feasible solution to the constrained optimization model of segmentation. The constraint also ensures that the optimal contour does not deviate too much from the initial contour. Projected-gradient descent equations are derived to solve the constrained optimization. In the experiments, we show that our method is capable of achieving the automatic object segmentation, and it outperforms the related methods.
\end{abstract}
% not want either math or citations to appear in the abstract.


%\maketitle

% is not selected <OR> if conference mode is selected - because compsoc
% conference papers position the abstract like regular (non-compsoc)

%\IEEEdisplaynotcompsoctitleabstractindextext

% \fi
\IEEEpeerreviewmaketitle

\section{Introduction}
% The reasons why the formulation of object segmentation is required.

%1. Point distribution cannot represent the object shape.

%2. Matching may not be accurate.
%%%%%%%Structure%%%%%%%%%
Automatic object segmentation is desirable in many higher level computer vision tasks, such as object detection, tracking, scene understanding etc. Active contour model is one of the well-known models for object segmentation. The active contour tries to find the boundary contour of the target object. However, it generally requires the user to input a contour curve sufficiently close to the object boundary as the initial contour. Hence, the existing framework of active contours are generally semi-automatic. This paper presents a new contour optimization model of object segmentation based on active contours and feature matching. The numerical optimization of our model leads to an automatic object segmentation algorithm.

%Is the method deploying any prior knowledge about the shape and the inner appearance of the object that is detected?
%- If yes, how is such a prior knowledge acquired?
%- How is the training performed?
%- How such a pre-knowledge is used in the cost function to be minimized?
%- Is the active contour algorithm proposed here used for the [object detection] or for the [object segmentation] stage?
%- Could the difference between these two stages be illustrated with a good figure?
%- What is the role of affine-invariance in the whole process?

\subsection{Related works}
Almost all the active contours fall into two categories, i.e. the edge-based and region-based active contour models. The edge-based active contour models, e.g. the Geodesic Active Contour \cite{caselles97GAC}, requires the initial contours provided by the user to be sufficiently close to the boundaries. This is because edges are local image feature distributed over the entire image domain, and it is crucial to determine which of the edges are of interest. Region-based active contours, e.g. the Chan-Vese model \cite{ChanVese01ActiveCon}, are often insensitive to initialization. However, several image regions may share similar regional property, and the discriminating regional property of an object of interest against various backgrounds in the image is difficult to be modeled exactly beforehand. The difficulty in the modeling may reduce the suitability of the region-based active contour for object segmentation in real images. It is generally recognized that the two type of models suit for different images. In this work, we will be focusing on the edge-based active contours.

Relaxation of the user initialization in edge-based active contours for object segmentation has become a research topic. Xu and Prince in \cite{GVF98} proposed the gradient vector field (GVF) to extend the localized gradients such that the large gradients at the boundary edges can influence the active contour far away from the edges. This allows the initial contours to be far from the edges. However, the active contour in GVF cannot find the boundary of interest when there are boundaries of other objects.  Paragios et al. \cite{GVFGAC04} also applied the GVF to the level set based active contours to extract multiple objects. However, they reported that their method may not perform well when the initialization is arbitrary. Li et al. \cite{LiSnake05Split} proposed to split the contour for extracting multiple objects by evolving the curve in a segmented GVF. Xie and Mirmehdi \cite{Xie08MAC} proposed a curve evolution formulation based on edge detection, which allows more flexible contour initialization, as an alternative to the conventional edge-based active contours. In our previous work \cite{Wang2008}, we have also addressed the problem of restrictive initialization according to the observations on the geometry of the gradient field. However, even though the initial contours can be more flexible, the indication of the object of interest by the users is still required. Previous attempt in \cite{Li08AutoInitAC} for automatic initialization of the edge-based active contours selects the initial contours that have approximately the minimum energy. In other words, it is assumed that most of the edges in the image are of interest so that they should be considered in the optimization. This assumption is valid for the images studied in \cite{Li08AutoInitAC} but it cannot be generalized.

The global optimization techniques for some region-based and edge-based active contour models have been proposed in \cite{Cremers08GlobalSP} \cite{Schoenemann10}. In \cite{Cremers08GlobalSP}, Cremers et al. proposed a branch-and-bound method for approximating the exhaustive search in a region-based active contour with shape prior modeling. Schoenemann and Cremers in \cite{Schoenemann07} and \cite{Schoenemann10} proposed a functional ratio energy for characterizing the object boundaries. The optimization is achieved via minimum ratio cycle algorithm by Lawler \cite{Lawler1966MRC}. The global search methods do not require initial contour. Hence, these methods achieve automatic object segmentation. The idea behind these methods is to approximate the exhaustive search of the globally optimal contour curve over the entire image domain. These methods provide efficacious numerical solutions to global optimization of many active contour models. However, it can sometimes be observed that the object of interest is one of the local optimal solutions to the active contour models, as shown in the experiment section of this paper.

\subsection{Methodology}
\begin{figure}
\centering
  \includegraphics[width=0.7\columnwidth]{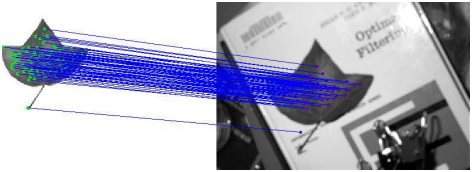}\\
  \caption{Illustration of the feature point matching (with errors).}\label{FIG:FPM}
\end{figure}
We address the automatic object segmentation following a divide-and-conquer strategy. We consider the problem of automatic object segmentation as a unity of two subproblems: the object detection and the boundary location. Unlike the global search methodology in \cite{Cremers08GlobalSP} \cite{Schoenemann10}, the object detection can provide the coarse location of the object with a relatively high confidence. We expect to obtain the segmentation with the initialization provided by the detector. The philosophy is that \emph{if the object detection cannot be done properly, then the object segmentation is also hopeless}. We aim at extending the formulation of active contour models by using object detection. There exist many effective object detectors based on classification, e.g. the Haar-like feature based detectors \cite{Viola04RealTimeFD} and the HOG descriptor based detectors \cite{Dalal05HOG} as well as object matching \cite{Lowe04SIFT}.

Object matching based on locally discriminative feature matching finds a set of points on the object of interest in the image \cite{Lowe04SIFT} \cite{Jiang09ScaRotMat} \cite{Li10ObjMatchLocalAffine}, achieving coarse object localization. The principle of the feature point matching is illustrated in Fig. \ref{FIG:FPM}, where the dots in two images are the feature points, and the lines across the images visualize the matching correspondences. In object matching, the feature point matching often starts from an interest point detection on a pair of images as in \cite{Lowe04SIFT}. Then the similarity between all possible pairs of matches are computed. Finally, the matching algorithm will assign a set of feature points in one image, possibly a subset of the entire feature set, to a set of feature points in the other image according to the point-wise similarity. A correspondence is often represented by using binary variable, i.e. $1$ means matched $0$ means unmatched.
%\begin{figure}[!t]
%\centering
%  % Requires \usepackage{graphicx}
%  \includegraphics[width=0.9\columnwidth]{Imgs/showtask3.eps}\\
%  \caption{Object segmentation through active contour with shape prior initialized by feature point matching.}\label{FIG:taskshow}
%\end{figure}

In this chapter, the formulation of active contour model is extended by using the formulations of object matching for automatic object segmentation. A difficulty is that the scatter of matched feature points cannot directly determine the object boundary or even approximate to the boundary reasonably. We propose to estimate the boundary contour of the object of interest by using the feature point matching. The boundary estimation can be used as the initial contour for active contours, and the initial contour is to be optimized in the active contour framework for object segmentation. This framework is also formulated as a constrained optimization model. In the experiment, we show that our method is capable of achieving accurate object extraction when the assumption in \cite{Cremers08GlobalSP} \cite{Schoenemann07} \cite{Schoenemann10} does not hold true, i.e., the object of interest is not the global optimal solution to active contour models.

\subsection{Contributions}
Our contributions mainly lie in three aspects. a) We obtain a mathematical model of the points-to-shape relation. We assume that such relation is invariant to affine transformation. We use this model to estimate the boundary contour given the matched interior points. b) Using the model of points-to-shape relation, we also obtain an affine active contour in which the contour motion is determined by the motions on the inner points, and we assume the point motion is due to affine transformation to preserve the shape initialized by the matching. c) These two contributions have readily led to the segmentation framework, and we further formulate the framework as a novel constrained optimization problem, which is called matching-constrained active contour model. The initial contour generated by using point matching is the initial feasible solution to the constrained optimization. We derive the projected-gradient descent equations for solving the constrained optimization.

\subsection{Organization}
The rest of the paper is organized as follows. In Section \ref{SEC:Shape_Modeling}, we introduce our mathematical model of the affine-invariant interior-points-to-shape relation, which leads up to the affine points-to-shape alignment. In Section \ref{SEC:MCAC}, we present the unified constrained optimization model of the framework, namely the matching-constrained active contour, and the projected gradient descent algorithm. In Section \ref{SEC:Exp}, we demonstrate our method for automatic object segmentation on real images of cluttered scenes, we compare our method with the state-of-the-art methods on example based object segmentation and we present the associated quantitative analysis with discussions. We conclude the paper with discussions on future directions in Section \ref{SEC:Concl}.

\section{Modeling affine-invariant interior-points-to-shape relation}\label{SEC:Shape_Modeling}
In this section, we present our mathematical model of the relation between interior points and the outlining contour shape, which enables the contour alignment based on feature matching.

\subsection{Points-to-shape relation as a binary classifier}
The reference object shape can be represented by its silhouette. The shape silhouette of an object can be defined as a binary function $H_o(x,y)$ as follows:
\begin{equation}\label{EQ:Def_H_o}
H_o(x,y) = \left\{\begin{array}{lr}
                1,& [x,y]\in\Omega\\
                0,& [x,y]\in\overline{\Omega}
             \end{array}\right.
\end{equation}
where $\Omega$ is the object region and $\overline{\Omega}$ is the non-object region. This definition can be used to define the object boundary contour $C$ as follows:
\begin{equation}\label{EQ:Def_C}
C = \left\{x,y\big|\|\nabla H_o(x,y)\| \neq 0\right\}
\end{equation}
where $H_o = H(\phi(x,y))$, $H$ is a Heaviside function and $\phi$ is a signed distance function. Generally, $\|\nabla H_o(x,y)\|=\delta(\phi(x,y))$, where $\delta$ is a Dirac delta function. In other words, we employ the implicit shape representation in this work, which is motivated by the level set method \cite{OsherSethian88Fronts}. The contour in the reference image is a set of discrete points and we propose to estimate the continuous contour shape by function approximation as follows:
\begin{equation}\label{EQ:Shape_Approx}
\begin{split}
H_e^* &= \argmin_{H_e\in \mathcal{S}}~\mathcal{E}(H_e)\\
\mathcal{E}(H_e) &= \int_\mathcal{D} |H_o-H_e|^2 dxdy
\end{split}
\end{equation}
where $\mathcal{D}$ is the entire image domain, $H_e$ is the approximate of $H_o$ and $\mathcal{S}$ denotes the solution space of $H_e$.

\begin{figure}[thb]
\centering
\includegraphics[width=0.7\columnwidth]{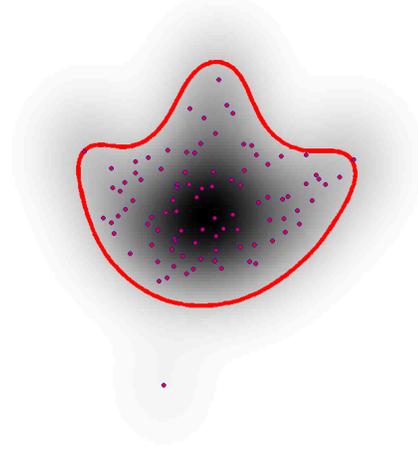}
  \caption{A 2D visualization of the trained shape decision function overlayed the corresponding implicit contour (in red). The larger the intensity of the image is the more likely the point belongs to the shape region.}\label{FIG:show_shape}
\end{figure}

In our context, $H_e$ is generated by the set of the position vectors of feature points $\{\vec{\mathbf{p}}\} = \{\vec{p}_1, \vec{p}_2,..., \vec{p}_N\}$, i.e. $H_e = H_e(\{\vec{p}\})$. If we consider the feature points $\{\vec{\mathbf{p}}\}$ as randomly distributed points, we may adopt the radial basis function (RBF) for the function approximation:
\begin{equation}\label{EQ:RBF_H_e}
H_e(\vec{z}) = H\left(\sum_{i=1}^N \alpha_i \psi(\|\vec{z}-\vec{p}_i\|) + \beta\right)
\end{equation}
where $\vec{z}=[x,y]^T$, $H$ is a Heaviside function and we may call $\phi(\vec{z}) = \sum_{i=1}^N \alpha_i \psi(\|\vec{z}-\vec{p}_i\|) + \beta$ the \emph{shape decision function} in which $\psi(\cdot)$ is a kernel function, $\{\alpha_i\},\beta$ are the weights and bias to be determined. The points $\{\vec{\mathbf{p}}\}$ are the center points of the kernel functions. The sign of shape decision function determines whether a point belongs to the shape. An example of the model is shown in Fig. \ref{FIG:show_shape}.

Combining the previous formulations in Eqs. (\ref{EQ:Def_H_o}), (\ref{EQ:Shape_Approx}) and (\ref{EQ:RBF_H_e}), we obtain a binary classification problem based on RBF neural network in which the decision function is formed by only some of the positive samples, i.e. the feature points, but the training is accomplished by using both positive and negative samples over the entire image domain. We consider that the training is done by direct minimization of the fitting error in (\ref{EQ:Shape_Approx}) with respect to the parameters $\{\alpha_i\}$ and $\beta$. The gradient descent equations for learning the parameters are given in \ref{APD-A}. The detailed learning strategy is presented in the section of experiment.

%The gradient descent equations for learning the parameters $\{\alpha_i\}$ and $\beta$ are as follows.
%\begin{equation*}
%\begin{split}
%{\partial\alpha_i\over\partial t} =& -\int_\mathcal{D} 2(H_e-H_o)\psi(\|\vec{z}-\vec{p}_i\|) H'dxdy,\\
%&i=1,2,\ldots,N\\
%\end{split}
%\end{equation*}
%\begin{equation*}
%{\partial\beta\over\partial t} = -\int_\mathcal{D} 2(H_e-H_o) H'dxdy
%\end{equation*}
%where $H'$ is the first order derivative of $H$. The trained contour curve is defined by Eq. (\ref{EQ:Def_C}) in which $H_o$ is replaced by the trained $H_e$.

%Note that the training based on support vector machine could be faster and more reliable, but the RBF network is sufficient for our problem of 2D function approximation with dense training sample.

\subsection{Imposing affine invariance in the points-to-shape relation}

\begin{figure}[thb]
\centering
{\includegraphics[width=0.7\columnwidth]{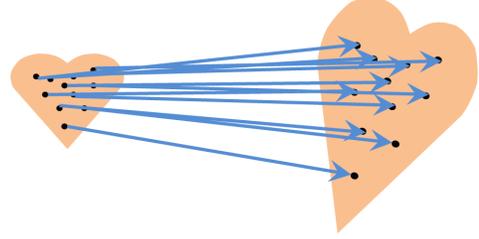}}
  \caption{Illustration of the need of affine invariance. The point correspondences should lead to the shape alignment.}\label{FIG:Need_Aff_Invar}
\end{figure}

In real images, the object shape and the positions of the interior feature points are often different from those given in the training session. Given the matched set of feature points in the image, we are left to determine the object shape. We assume that the contour shape of the object undergoes the transformation the same as that of the interior feature points, and we consider the affine transformation especially. This property is illustrated in Fig. \ref{FIG:Need_Aff_Invar}. This assumption is common in vision. The behavior that the shape and points undergo the same affine transformation is the \emph{affine invariance property}. However, we have not examined whether the implicit contour of the trained $H_e$, or the shape decision function $\phi(\cdot)$, have the affine invariance property. In the following, we will discuss about this issue.

We choose the Gaussian function as the RBF kernel. Thus, the shape decision function can be written as follows.
\begin{equation}\label{EQ:Sim_Shape_Dec}
\phi(\vec{z}) = \sum_{i=1}^N {\alpha_i}\exp\left(-{\|\vec{z}-\vec{p}_i\|^2\over\sigma^2}\right) + \beta.
\end{equation}
The contour curves can be defined by the shape decision function according to $C=\{x,y|\phi(x,y)=0\}$, which is equivalent to Eq. (\ref{EQ:Def_C}) where $H_o$ is replaced by $H_e$.

Let us consider the affine transformation of the kernel centers $\vec{p}_i$ as follows:
\begin{equation}\label{EQ:Trans_P}
\vec{q}_i = \mathbf{A}\vec{p}_i + \vec{b}, \hbox{~for all~} i,
\end{equation}
where $\mathbf{A}$ is an invertible matrix, $\vec{b}$ is a translation vector. The corresponding shape decision function in terms of the transformed kernel centers can be written as follows:
\begin{equation}\label{EQ:Shape_Dec_Trans}
\phi(\vec{z},\{\mathbf{A},\vec{b}\}) = \sum_{i=1}^N {\alpha_i}\exp\left(-{\|\vec{z}-\mathbf{A}\vec{p}_i-\vec{b}\|^2\over\sigma^2}\right) + \beta
\end{equation}

%We wonder whether the points on the implicit contour defined by the shape decision function generated by the transformed kernel centers is identical to the contour generated by direct transformation of the original contour.
The affine transformation of the contour points can be represented by $\vec{z}_s = \mathbf{A}\vec{z}_c + \vec{b}$, where $\vec{z}_c\in C$, such that $\phi(\vec{z}_c)=0$ in (\ref{EQ:Sim_Shape_Dec}). By substituting $\vec{z}_s$ into Eq. (\ref{EQ:Shape_Dec_Trans}), we obtain the following:
\begin{equation*}
\phi(\vec{z}_s,\{\mathbf{A},\vec{b}\})= \sum_{i=1}^N {\alpha_i}\exp\left(-{\|\mathbf{A}\vec{z}_c -\mathbf{A}\vec{p}_i\|^2\over\sigma^2}\right) + \beta.
\end{equation*}

Regarding the affine invariance property, the above leads to the following.
\begin{proposition}\label{Prop:NAffInV}
There exists infinitely many $\vec{z}_c$ and $\vec{p}_i$, such that for each $i$, we have $\exp\left(-{\|\mathbf{A}\vec{z}_c -\mathbf{A}\vec{p}_i\|^2\over\sigma^2}\right)\neq\exp\left(-{\|\vec{z}_c -\vec{p}_i\|^2\over\sigma^2}\right)$, if $\mathbf{A}$ is not an orthogonal matrix.
\end{proposition}
The proof is deferred to \ref{APD-B}.

This means that the affine transformation of the contour curve defined by the trained shape decision function, $\phi(\vec{z})$, may not be the contour curve defined by the transformed shape decision function $\phi(\vec{z},\{\mathbf{A},\vec{b}\})$, which is an violation of the aforementioned affine invariance property. To address this problem, we propose a revised shape representation $\phi_S$ as follows.
\begin{equation}\label{EQ:phi_S}
\begin{split}
&\phi_S(\vec{z},\{\mathbf{A}^{-1},\vec{b}\})\\
%&=\sum_{i=1}^N \alpha_i {\exp\left(-{\|\mathbf{A}^{-1}(\vec{z} -\mathbf{A}\vec{p}_i-\vec{b})\|^2}\right)} + \beta\\
&=\sum_{i=1}^N \alpha_i {\exp\left(-{\|\mathbf{A}^{-1}(\vec{z} -\mathbf{A}\vec{p}_i-\vec{b}))\|^2}\over\sigma^2\right)} + \beta.
\end{split}
\end{equation}
Substituting $\vec{z}_s$ into (\ref{EQ:phi_S}), we can verify that for general $\mathbf{A}$ and $\vec{b}$, we have the following affine invariance.
\begin{equation*} \phi_S(\vec{z}_s,\{\mathbf{A}^{-1},\vec{b}\})=\phi_S(\vec{z}_c,\{\mathbf{I}^{-1},\vec{0}\})=\phi(\vec{z}_c) = 0
\end{equation*}
where $\mathbf{I}=\left(\begin{array}{cc}1&0\\0&1\end{array}\right)$ and $\vec{0}$ defines an identity transformation. We consider that $\phi_S$ is parameterized by $\mathbf{A}^{-1}$ rather than $\mathbf{A}$ to simplify the later derivation for optimizing $\mathbf{A}$.
%The shape representation defined by Eq. (\ref{EQ:phi_S}) is static. We require a dynamic deformable shape model in this subsection for segmentation. The dynamic contour shape based on the affine-invariant shape decision function in Eq. (\ref{EQ:phi_S}) can be obtained by parameterizing the transformation in $\phi_S$ with an artificial time $t$ as follows.
%\begin{equation}\label{EQ:Dyn_C}
%C(t)=\{x,y|\phi(\vec{z}_s,\{\mathbf{A}(t),\vec{b}(t)\})=0\}
%\end{equation}

\begin{figure}[thb]
\centering
  \subfloat[]{\includegraphics[width=0.4\columnwidth]{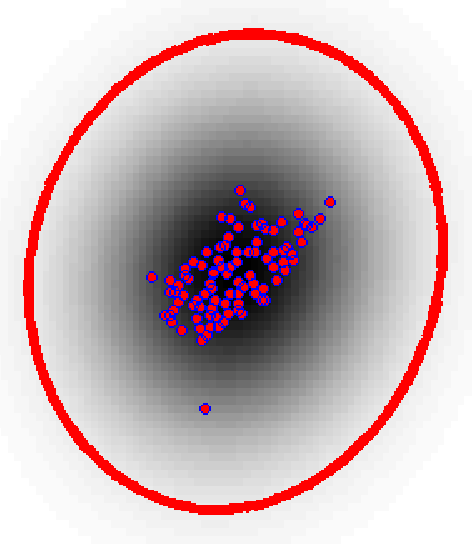}}
  \subfloat[]{\includegraphics[width=0.4\columnwidth]{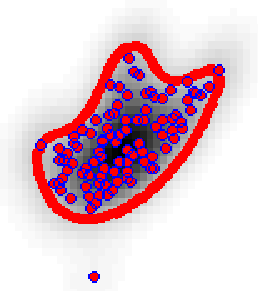}}
  \caption{Significance of affine invariance of the shape representation. The dots are the transformed feature points. The outlining curves are the implicit shape contours.}\label{FIG:Aff_Invar}
\end{figure}

\begin{figure}[!t]
\centering
  \includegraphics[width=0.7\columnwidth]{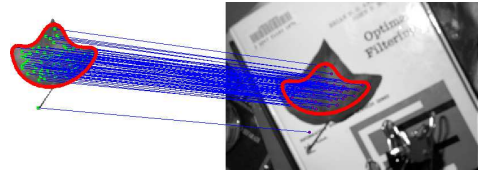}\\
  \caption{A demo of points-to-shape alignment.}\label{FIG:p2s_alignment}
\end{figure}

We may further appreciate the significance of the affine invariance property by the example shown in Fig. \ref{FIG:Aff_Invar}. Suppose we are given the pair of initial points and shape in Fig. \ref{FIG:show_shape}, the interior feature points of the leaf are then transformed by a predetermined affine transformation. The implicit contour curve defined by Eq. (\ref{EQ:Shape_Dec_Trans}) is shown in Fig. \ref{FIG:Aff_Invar}(a). Note that Eq. (\ref{EQ:Shape_Dec_Trans}) does not have the affine invariance property. The resultant shape, which tends to be a circle, differs significantly from a shape of a leaf. With the same transformation of the feature points, the implicit contour curves defined by the affine invariant shape decision function in Eq. (\ref{EQ:phi_S}) is shown in Fig. \ref{FIG:Aff_Invar}(b). Obviously, the affine-invariant implicit contour in Fig. \ref{FIG:Aff_Invar}(b) is like a leaf transformed resulting from the same transformation applied to the feature points, which is the affine-invariance property. With such an affine invariant interior-points-to-shape relation in Eq. (\ref{EQ:phi_S}), we can achieve the affine points-to-shape alignment. A result of the points-to-shape alignment is shown in Fig. \ref{FIG:p2s_alignment} based on the feature matching presented in Fig. \ref{FIG:FPM}.

Lastly, we enumerate some of the advantages of the established interior-points-to-shape model over the explicit transformation of the contour shape given in the object template.
\begin{enumerate}
  \item The affine invariant interior-points-to-shape relation provides continuous implicit contours without reparametrization, which can simplify the implementation.
  \item Sometimes, the entire region of object shape, namely the shape silhouette, is of interest. However, the explicit transformation generally cannot provide a uniform region of shape. The proposed interior-points-to-shape relation provides the shape silhouette by definition.
\end{enumerate}

\section{The matching-constrained active contour}\label{SEC:MCAC}
%The framework of joint object matching and segmentation presented in the last section may be practically useful.
%Previously, we have shown that with the affine-invariant interior-points-to-shape relation we can estimate the object boundary. With the estimation, we can initialize an affine shape prior based active contour for object segmentation without user intervention. In this section, we formulate a constrained optimization model that unifies the initialization and segmentation to be a single optimization problem.
The proposed model of affine-invariant interior-points-to-shape relation can be used for contour alignment given point matching. The implicit contours defined by the affine-invariant interior-points-to-shape relation are continuous and they do not require reparametrization when the shape changes. Such continuous geometrical shape representation is also desired by the active contour framework for segmentation. Therefore, we can reformulate the conventional active contour model by using our affine invariant points-to-shape relation. Due to the affine invariance, the solution to the reformulated active contour model will be an affine transformation of the initial contour. Hence, an affine shape prior adheres to the reformulated active contour model.

In  this section, we propose a general active contour model based on the affine-invariant interior-points-to-shape relation, with a constraint of matching.

\subsection{Active contour model with matching constraint}

Given the interior-point-to-shape relation, we are able to achieve affine contour alignment by affine point alignment. The affine point alignment can be obtained by using the matching correspondences. The aligned contour can then be used to initialize the curve evolution in the affine active contour. These all form an algorithmic framework. In this section, we propose an optimization model to unify the matching, alignment and curve evolution in a unified optimization framework. The motivation of our constrained optimization model lies in the notion of \emph{initial feasible solution}. Specifically, the contour curve given by the feature points matching and alignment provides the initial feasible solution to the constrained optimization model of active contour. The local features matching can also be used to constrain the optimization. The model is as follows:
\begin{equation}\label{EQ:MCAC_1}
\begin{split}
C^* &= \argmin_{C} J(C(\{\vec{\mathbf{q}}\}))\\
&\hbox{s.t.:~} E(\{\mathbf{q}\})\leq \tau,~~\vec{q}_i = \mathbf{A}\vec{p}_i + \vec{b},
\end{split}
\end{equation}
where $\{\vec{\mathbf{p}}\}=\{\vec{p}_1,\vec{p}_2,...,\vec{p}_N\}$ and  $\{\vec{\mathbf{q}}\}=\{\vec{q}_1,\vec{q}_2,...,\vec{q}_N\}$ are the feature points on the template and the transformed sets of these feature points, $C(\{\vec{\mathbf{q}}\})$ denotes the implicit relation between the interior points and the contour shape defined by (\ref{EQ:phi_S}). $J(C)$ is the abstract form of the active contour energy, $E(\cdot)$ is the abstract form of the cost of joint matching and alignment, and $\tau$ is a tolerance level. Note that the set of $\{\vec{\mathbf{q}}\}$ that corresponds to the minimal value of $E(\cdot)$ must be feasible to the inequality if there exists at least one feasible solution. By this formulation, we assume that the optimal contour can be obtained by transforming the boundary of the template object, and we assume that the transformation is affine.
%The feasibility of the optimally matched and registered points can hold true for sufficiently large tolerance level. For example, if $E(\{\vec{p}\})$ measures the optimality of the point set $\{\vec{p}\}$ in terms of matching and registration, and we set
%\begin{equation}
%\tau =\sqrt{(\min_{\{\vec{q}\}}E(\{\vec{q}\})+\epsilon^2},
%\end{equation}
%where $\epsilon\neq0$. Then, we have that $E(\{\vec{p}_o\})\leq\tau^2$, if $\{\vec{p}_o\}=\argmin_{\{\vec{q}\}}E(\{\vec{q}\})$. $\{\vec{p}_o\}$ is the set of optimally matched and registered points.

\subsection{Projected gradient decent algorithm}
Without knowing the actual formulations of the objective functional $J$ and the constraint $E$, we are still able to derive the solution to the abstract optimization problem based on the idea of projected gradient descent method \cite{Luenberger08BookLP&NLP}. Our projected-gradient descent equations are as follows:
\begin{equation}\label{EQ:PG_A}
{d\mathbf{A}\over dt} = \nabla_{\mathbf{A}}J-\left\langle\nabla_{\mathbf{A}}J,{\nabla_{\mathbf{A}}E\over\|\nabla_{\mathbf{A}}E\|}\right\rangle{\nabla_{\mathbf{A}}E\over\|\nabla_{\mathbf{A}}E\|},
\end{equation}
\begin{equation}\label{EQ:PG_b}
{d\vec{b}\over dt} = \nabla_{\vec{b}}J-\left\langle\nabla_{\vec{b}}J,{\nabla_{\vec{b}}E\over\|\nabla_{\vec{b}}E\|}\right\rangle{\nabla_{\vec{b}}E\over\|\nabla_{\vec{b}}E\|}.
\end{equation}
Note that these equations are not exactly the same as the ones in conventional projected gradient descent method in which inversion of a large matrix is needed. The rationale of these equations lies in the following property.
%\begin{equation}\label{EQ:PG_KKT_Ab}
%\begin{split}
%\left(\begin{array}{l}
%{d\mathbf{A}\over dt}\\
%{d\vec{b}\over dt}
%\end{array}\right)=&
%-\left(\begin{array}{l}
%\nabla_{\mathbf{A}}J\\
%\nabla_{\vec{b}}J
%\end{array}\right)\\
%&+\left(\begin{array}{l}
%\left\langle\nabla_{\mathbf{A}}J,{\nabla_{\mathbf{A}}E\over\|\nabla_{\mathbf{A}}E\|}\right\rangle\\
%+\left\langle\nabla_{\vec{b}}J,{\nabla_{\vec{b}}E\over\|\nabla_{\vec{b}}E\|}\right\rangle
%\end{array}\right)
%\left(\begin{array}{l}
%{\nabla_{\mathbf{A}}E\over\|\nabla_{\mathbf{A}}E\|}\\
%{\nabla_{\vec{b}}E\over\|\nabla_{\vec{b}}E\|}
%\end{array}\right)
%\end{split}
%\end{equation}
%where $\nabla_{\mathbf{A}}J$ can be obtained by (\ref{EQ:OBJ_GD_invA}), numerically we may update $\mathbf{A}^{-1}$ and compute its inverse and estimate $\nabla_{\mathbf{A}}J$.  $\nabla_{\mathbf{A}}E$, $\nabla_{\vec{b}}J$ and $\nabla_{\vec{b}}E$ can be obtained by Eqs. (\ref{EQ:OBJ_GD_b}), (\ref{EQ:FXP_A}) and (\ref{EQ:FXP_b}). Substituting the above into $E$ and $J$,

%\begin{equation}
%{dE\over dt}=\nabla_{\mathbf{A}}E {d\mathbf{A}\over dt}+\nabla_{\vec{b}}E {d\vec{b}\over dt}\leq 0
%\end{equation}
%Besides, we may also validate the following.
%\begin{equation}
%{dJ\over dt}=\nabla_{\mathbf{A}}J {d\mathbf{A}\over dt}+\nabla_{\vec{b}}J {d\vec{b}\over dt}\leq 0
%\end{equation}
\begin{equation}\label{EQ:dE/dt_dJ/dt}
{dE\over dt} = 0\hbox{~~and~~}{dJ\over dt}\leq 0.
\end{equation}
The derivation of this property is deferred to \ref{APD-C}.
%\begin{equation}\label{EQ:dE/dt}
%\begin{split}
%{dE\over dt}=&-\nabla_{\mathbf{A}}E \left(\nabla_{\mathbf{A}}J-\left\langle\nabla_{\mathbf{A}}J,{\nabla_{\mathbf{A}}E\over\|\nabla_{\mathbf{A}}E\|}\right\rangle{\nabla_{\mathbf{A}}E\over\|\nabla_{\mathbf{A}}E\|}\right)\\
%&=0
%\end{split}
%\end{equation}
%
%\begin{equation}\label{EQ:dJ/dt}
%\begin{split}
%{dJ\over dt}=&-\nabla_{\mathbf{A}}J \left(\nabla_{\mathbf{A}}J-\left\langle\nabla_{\mathbf{A}}J,{\nabla_{\mathbf{A}}E\over\|\nabla_{\mathbf{A}}E\|}\right\rangle{\nabla_{\mathbf{A}}E\over\|\nabla_{\mathbf{A}}E\|}\right)\\
%&\leq0
%\end{split}
%\end{equation}
The above indicates that the projected gradient descent algorithm governed by Eqs. (\ref{EQ:PG_A}) and (\ref{EQ:PG_b}) can reduce $J$ while leaving $E$ unchanged. This might be too strong for the inequality constrained optimization. In fact, we only require $E$ to be smaller than a predefined tolerance $\tau$. Therefore, we implement the original gradient descent if $E<\tau$, and we implement the full projected gradient algorithm, if $E\approx\tau$. To implement the projected gradient descent algorithm, we require the explicit form of the gradients $\nabla_{\mathbf{A}}J$.  $\nabla_{\mathbf{A}}E$, $\nabla_{\vec{b}}J$ and $\nabla_{\vec{b}}E$.

\subsection{Gradient of active contour parameterized by affine motion of interior points}
The projected gradient descent algorithm requires the explicit form of the gradients of the active contour energy, i.e.  $\nabla_{\mathbf{A}}J$ and $\nabla_{\vec{b}}J$. The derivation of the explicit form of $\nabla_{\mathbf{A}}J$ is complex. Alternatively, we can obtain $\nabla_{\mathbf{A}}J$ numerically by using $\nabla_{\mathbf{A}^{-1}}J$. The explicit form of $\nabla_{\mathbf{A}^-1}J$ and $\nabla_{\vec{b}}J$ can be written as follows:
\begin{equation}\label{EQ:OBJ_GD_invA}
\nabla_{\mathbf{A}^-1}J = -{\nabla J^T\vec{N}\over\vec{N}^T\nabla_{\vec{z}}\phi_S}{D\phi_S\over D\mathbf{A}^{-1}},
\end{equation}
\begin{equation}\label{EQ:OBJ_GD_b}
{\nabla_{\vec{b}}J} = -{\nabla J^T\vec{N}\over\vec{N}^T\nabla_{\vec{z}}\phi_S}{D\phi_S\over D\vec{b}},
\end{equation}
in which $\vec{N}$ is the normal of the contour and $\nabla_{\vec{z}}\phi_S$ can be computed either numerically or exactly. The closed form expressions of ${D\phi_S\over D\mathbf{A}^{-1}}$ and ${D\phi_S\over D\vec{b}}$ are needed for the computation, which are as follows:
\begin{equation*}
{D\phi_S\over D\mathbf{A}^{-1}} = -\sum_{i=1}^N 2w_i(\vec{z}){\vec{v}_i(\vec{z})\vec{z}^{T}\over\sigma^2},
\end{equation*}
\begin{equation*}
{D\phi_S\over D\vec{b}} = -\sum_{i=1}^N 2w_i(\vec{z}){(\mathbf{A}^{-1})^{T}\vec{v}_i(\vec{z})\over\sigma^2},
\end{equation*}
where
\begin{equation*}
w_i(\vec{z})=\alpha_i\exp\left(-{\|\mathbf{A}^{-1}\vec{z} -\vec{p}_i-\mathbf{A}^{-1}\vec{b})\|^2\over\sigma^2}\right),
\end{equation*}
\begin{equation*}
\vec{v}_i(\vec{z})=\mathbf{A}^{-1}\vec{z} -\vec{p}_i-\mathbf{A}^{-1}\vec{b}.
\end{equation*}
The derivation is deferred to \ref{APD-D}. Our derivation is not restricted to any specific active contour model. For the Geodesic Active Contour (GAC) \cite{caselles97GAC}, which is a well-known edge-based active contour and can locate the object boundary accurately given a good initialization, the functional gradient, $\nabla J$, is the following:
\begin{equation*}
\nabla J_{_{GAC}} = \langle\nabla g,\vec{N}\rangle\vec{N}-g\kappa\vec{N},
\end{equation*}
where $g$ is an edge indicator function in which the stronger edge corresponds to smaller value, $\vec{N}$ is the normal of the contour, $\kappa$ is the contour curvature.

%\subsection{Local Feature Matching}
%%%%%%%%%%%%%%%%%%%%%%%%%%%%%%%%%%%%%%%%%%%%%

\subsection{Joint formulation of point matching and alignment}
To obtain the explicit form of $\nabla_{\mathbf{A}}E$ and $\nabla_{\vec{b}}E$, we require the explicit formulation of $E$ in the constraint. $E$ is a measure of the optimality of feature matching and point alignment. Our formulation is a slight variate of the simplest form of the linear model of of feature matching \cite{Jiang07LinMatch}. Our joint optimization model of matching and alignment can be written as follows:
\begin{equation}\label{EQ:JF_match_regi}
\begin{split}
\left\{{\vec{\mathbf{q}^{*}}}\right\}&=\argmin_{\left\{{\vec{\mathbf{q}}}\right\}} ~E({\vec{\mathbf{q}}}) = \sum_{ij}c_{ij}\delta_\epsilon(r_{ij})\\
&\hbox{s.t. }\forall i,\sum_{j} \delta_\epsilon(r_{ij}) = 1,
\end{split}
\end{equation}
where $c_ij$ is the so-called matching cost measuring the distance between all pairs of features across the two images. We use the SIFT interest point detector and the SIFT feature \cite{Lowe04SIFT}. $\delta_\epsilon$ is the approximation of the Dirac delta with a parameter $\epsilon$, $r_{ij} = \|{\vec{q_i}}-\vec{q_j^t}\|$ and $\{\vec{\mathbf{q^t}}\}$ is the set of target feature points in the target image, $c_{ij}$ is the cost of the matching between ${\vec{q_i}}$ and $\vec{q_j^t}$. $\delta_\epsilon(r_{ij})$ is the relaxed matching indicator. $\delta_\epsilon(r_{ij})$ has the following property:
\begin{equation*}
\delta_\epsilon(r_{ij}) \approx \left\{\begin{array}{lr}
                                  M, & r_{ij} = {0}\\
                                  0, & r_{ij}\neq {0}
                         \end{array}\right.,
\end{equation*}
where $M$ is a constant. In this model, the optimal $\delta_\epsilon$ is determined by the closeness between ${\vec{q_i}}$ and $\vec{q^t_j}$. Thus, by aligning ${\vec{q_i}}$ toward a proper $\vec{q^t_j}$, the matching cost can be minimized. However, the above model is easily trapped by degenerate solutions where $\delta_\epsilon(r_{ij})=0$. To avoid such degeneration, we have the following reformulation:
\begin{equation}\label{EQ:JF_match_regi1}
\begin{split}
\left\{\vec{\mathbf{q}^*}\right\}&=\argmax _{\left\{\vec{\mathbf{q}}\right\}} ~\mathcal{E}(\{\vec{\mathbf{q}}\}) = \sum_{ij}e^{-c_{ij}}\delta_\epsilon(r_{ij})\\
&\hbox{s.t. }\forall i,\sum_{j} \delta_\epsilon(r_{ij}) = 1.
\end{split}
\end{equation}
The constraint in terms of $\mathcal{E}$ is therefore $E = -\mathcal{E}\leq\tau\leq0$.
%The conventional framework of matching and registration is theoretically sound. It is capable of providing the initial inner points and transformation parameters that determine the shape contour according to the shape decision function in (\ref{EQ:phi_S}) for object segmentation.

%We consider that the point correspondences and point locations provided by the given algorithms of matching and registration are near the global optimal solution of (\ref{EQ:JF_match_regi1}).
%To obtain our initial feasible solution to (\ref{EQ:MCAC_1}), we require an algorithm to produce a good solution to (\ref{EQ:JF_match_regi1}), e.g. the global optimal solution, such that $E$ is ensured less than or equal to a predetermined $\tau$. In this work, we consider such solution is attainable through a state-of-the-art matching algorithm followed by standard affine registration mechanism. The good solution still may not be the optimal solution of (\ref{EQ:JF_match_regi1}). Hence, we used the gradients of the joint matching-registration model in (\ref{EQ:JF_match_regi1}) for a local optimization of to arrive at the optimal solution.

The gradients of the energy function in (\ref{EQ:JF_match_regi1}) are as follows:
\begin{equation}\label{EQ:FixPT_A}
\nabla_{\mathbf{A}} E\big(\mathbf{A},\vec{b}\big) = -\sum_{ij} \hat{c}_{ij} g_{ij}(\mathbf{A}\vec{p_i}+\vec{b}-\vec{q^t_j})\vec{p_i}^T,
\end{equation}
\begin{equation}\label{EQ:FixPT_b}
\nabla_{\vec{b}} E\big(\mathbf{A},\vec{b}\big) = -\sum_{ij} \hat{c}_{ij} g_{ij}(\mathbf{A}\vec{p_i}+\vec{b}-\vec{q^t_j}),
\end{equation}
where $\hat{c}_{ij} = e^{-c_{ij}}$. We adopt the Gaussian function to approximate the Dirac delta. Thus, $g_{ij}=\delta_{\epsilon}(r_{ij})$. We also normalize the Gaussian functions according to the constraint $\sum_{i}\delta_\epsilon(r_{ij})=1$.

\begin{figure}
\centering
  \includegraphics[width=0.45\textwidth]{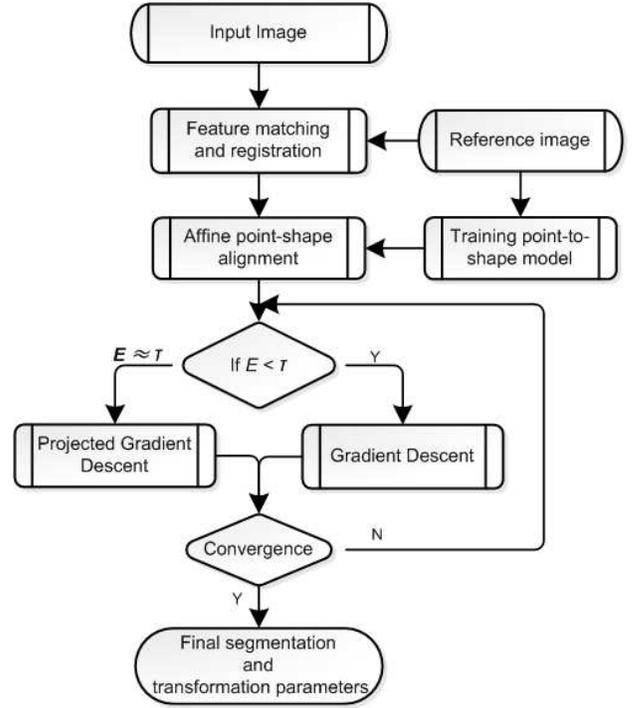}\\
  \caption{Flow chat of matching-constrained active contour framework.}\label{FIG:flowchart}
\end{figure}

Given the detailed formulations presented previously, we present the entire framework of the matching-constrained active contour in the pseudo code form in Algorithm \ref{Alg:PGAC}. We also visualize the algorithm in a diagram in Fig. \ref{FIG:flowchart}.
\begin{algorithm}
\DontPrintSemicolon
\LinesNumberedHidden
\SetKwInOut{Inputalg}{Input}\SetKwInOut{Outputalg}{Output}
\Inputalg{Input image $I$, source points $\{\vec{p}^s\}$, initial $\mathbf{A}_o,\vec{b}_o$, trained $\phi_S(\cdot)$}
\Outputalg{$H_e^*$}
\Begin{
\tcc{~~~~Points-to-shape alignment}
 Solving $\big[\mathbf{A}_o~ \vec{b}_o\big]\big[\mathbf{P}~ \vec{1}\big]^T=\mathbf{Q}^t$ w.r.t. $\mathbf{A}_o,\vec{b}_o$\;
    \Repeat{Convergence}{
        Update $\mathbf{A}_o,\vec{b}_o$ by Eqs. (\ref{EQ:FixPT_A}) and (\ref{EQ:FixPT_b})\;
            }
    $\vec{p_i}^{t_0} \longleftarrow \mathbf{A}_o\vec{p_i}^s+\vec{b}_o$\;
    Evaluate $\phi_S(\vec{z},\{\mathbf{A}_o^{-1},\vec{b}_o\})$ on image domain by Eq. (\ref{EQ:phi_S})\;
\tcc{~~~~Constrained optimization}
    $k \longleftarrow 1$\;
    \Repeat{Convergence}{
        \uIf{$E(\{\vec{p_i}^{t_{k-1}}\})<\tau$}{
            Update $\mathbf{A}_k$ and $\vec{b}_k$ by using  (\ref{EQ:OBJ_GD_invA}) and (\ref{EQ:OBJ_GD_b})\;
               }
        \Else{
            Update $\mathbf{A}_k$ and $\vec{b}_k$ by Eqs. (\ref{EQ:PG_A}) and (\ref{EQ:PG_b})\;
        }
        Evaluate $\phi_S(\vec{z},\{\mathbf{A}_k^{-1},\vec{b}_k\})$ image domain by Eq. (\ref{EQ:phi_S})\;
        $k \longleftarrow k + 1$\;
        }
}
-----------------------------------------------------------------------\;
\footnotesize{$\mathbf{P}=[\vec{p}_1,\vec{p}_2,...,\vec{p}_N]^T$ is an $N\times2$ matrix of feature vectors in the template image, $\mathbf{Q}^t=[\vec{q}^t_1,\vec{q}^t_2,...,\vec{q}^t_N]^T$ is an $N\times2$ matrix of the corresponding feature vectors in the target image.}\;
\caption{Projected gradient descent active contour}
\label{Alg:PGAC}
\end{algorithm}

The initial $\mathbf{A}_o$ and $\vec{b}_o$ are computed by using established point correspondences. We adopt the locally affine matching \cite{Li10ObjMatchLocalAffine} to provide the initial point correspondences. The matching score due to the locally affine matching is guaranteed to be high enough and the locally affine matching is robust to outliers.

\section{Experiments}\label{SEC:Exp}
In our implementation, we use the SIFT interest point detector and the SIFT feature \cite{Lowe04SIFT}, and we adopt the locally affine matching for producing the initial point correspondences. We are not confined to this choice of feature representation and matching algorithm. The $\tau$ in (\ref{EQ:MCAC_1}) is set in relation to the matching score $E$ from initial object matching.

We experiment on the real images taken from Mikolajczyk's homepage \footnote{\url{http://lear.inrialpes.fr/people/mikolajczyk/}}, Caltech computer vision archive \footnote{\url{http://www.vision.caltech.edu/html-files/archive.html}} and the ETHZ Toys dataset to evaluate our method for automatic object segmentation.

\subsection{Affine invariant shape modeling}
\begin{figure*}
\centering
\subfloat[]{\includegraphics[height=0.9in]{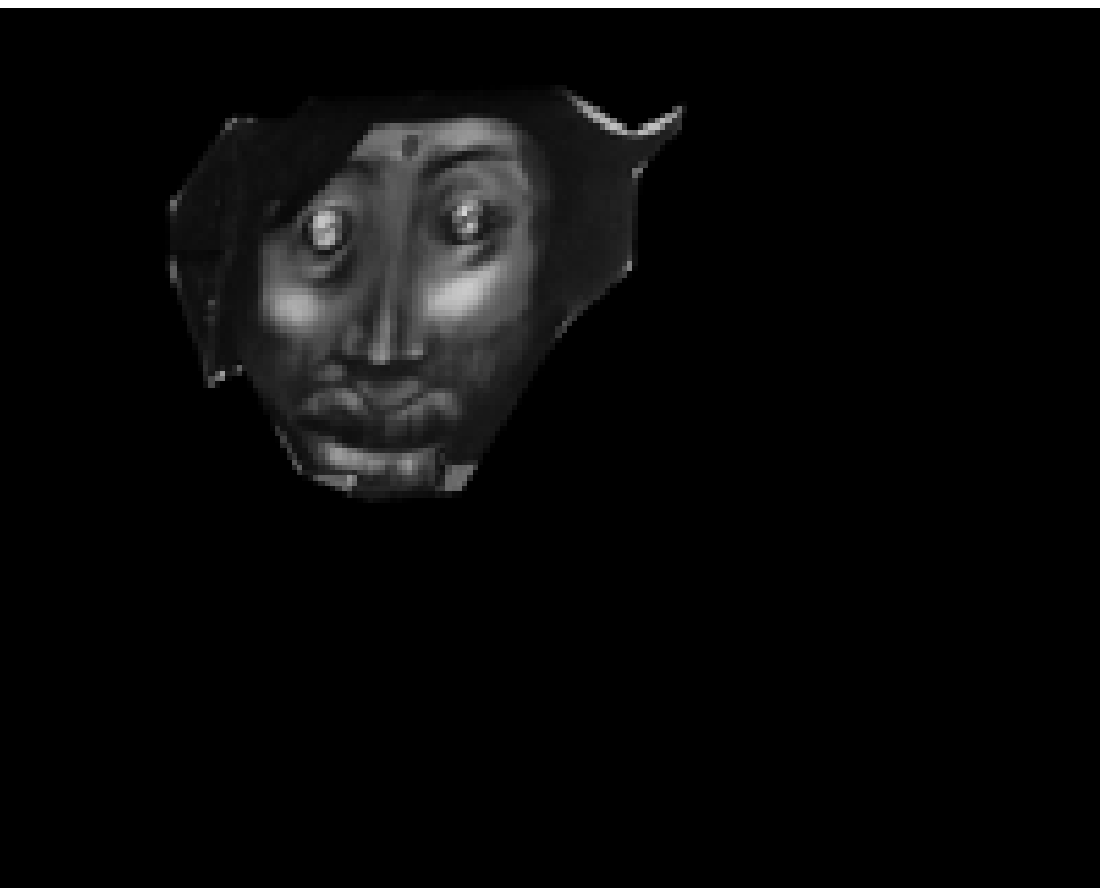}
\includegraphics[height=0.9in]{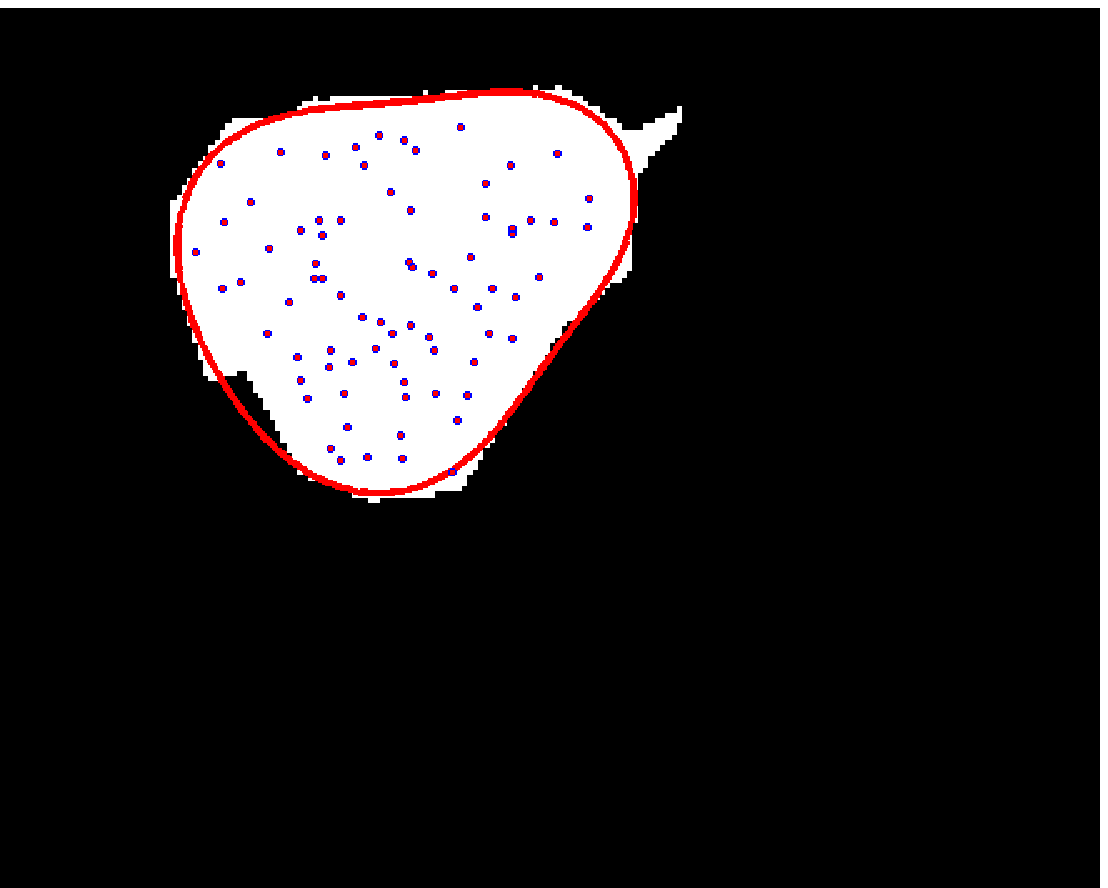}}~ \subfloat[]{\includegraphics[height=0.9in]{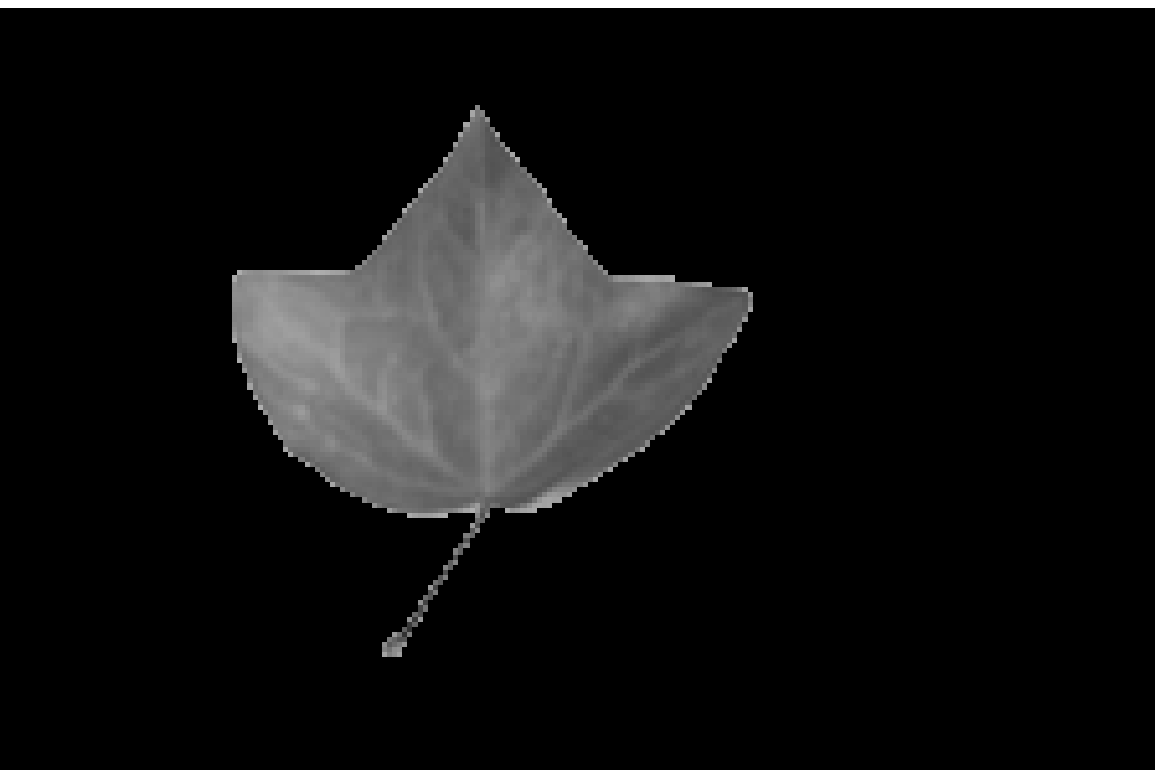}
\includegraphics[height=0.9in]{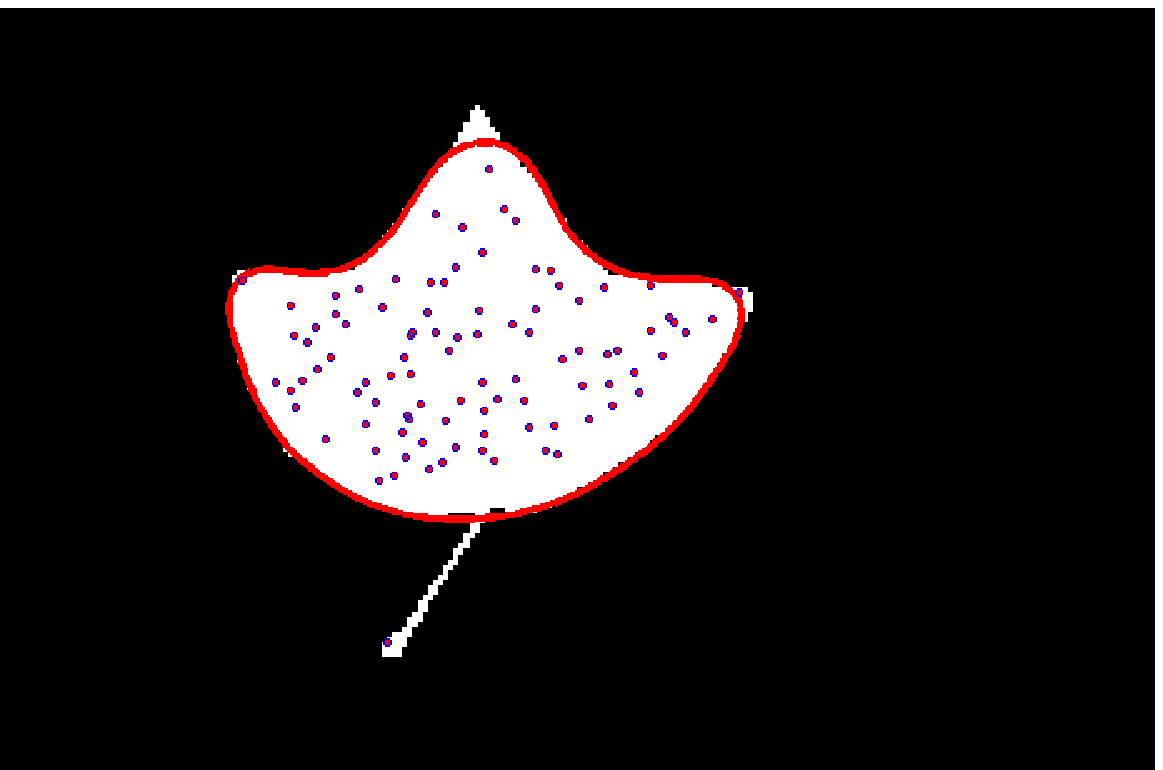}}~
\subfloat[]{\includegraphics[height=0.9in]{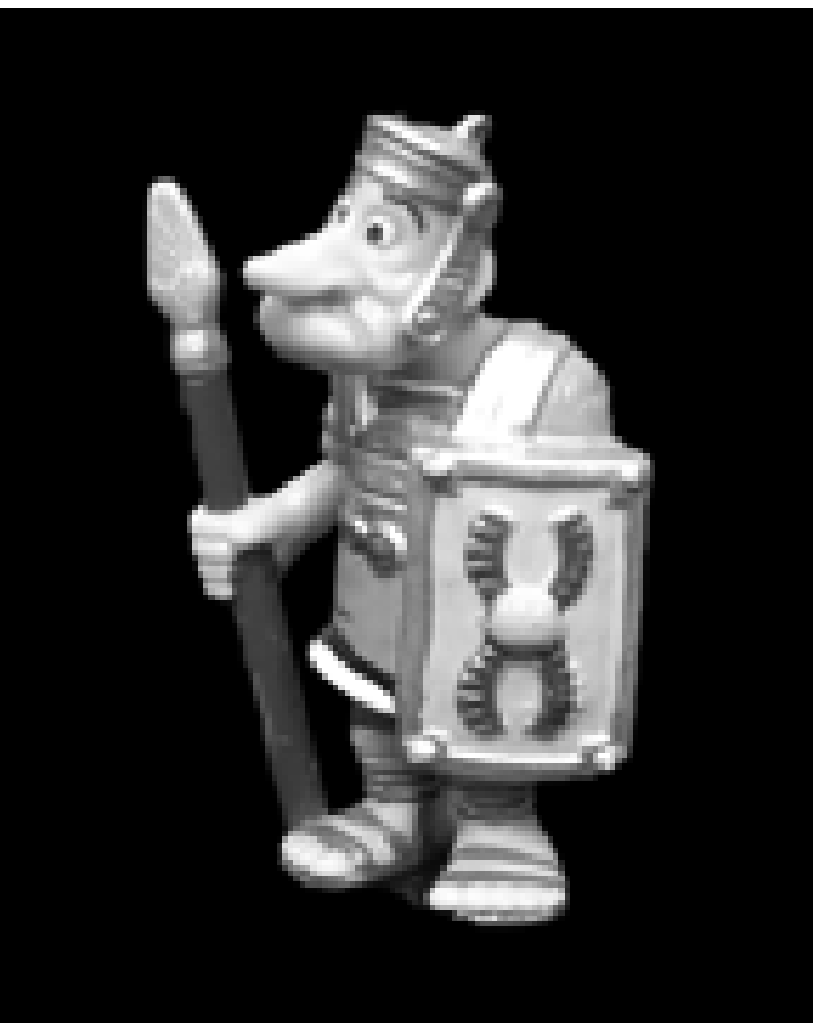}
\includegraphics[height=0.9in]{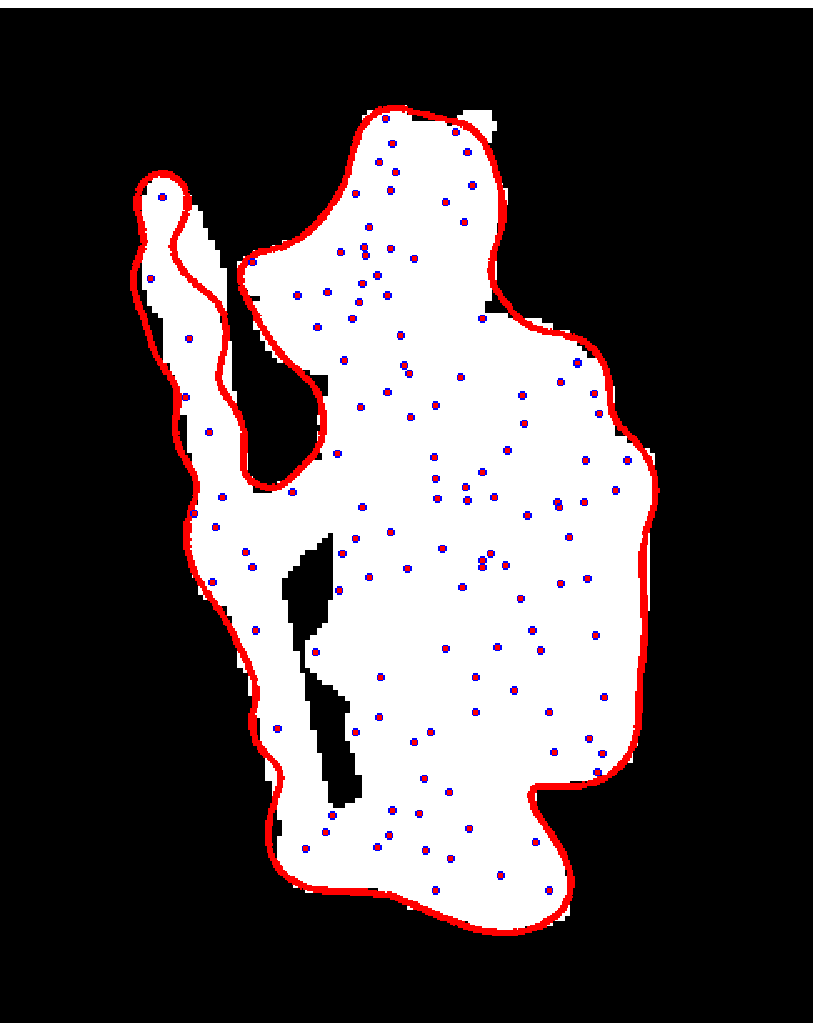}}
\caption{Three pairs of template objects (left) and the corresponding trained shape contours overlaying the training shape (right) }\label{FIG:Obj_and_Shape}
\end{figure*}

Following the shape training process presented in Section \ref{SEC:Shape_Modeling}, we can obtain the affine invariant points-to-shape model. We present the template objects and the trained shape contours in Fig. \ref{FIG:Obj_and_Shape}. We also present the fitting errors during the training of the RBF for the shapes in Fig. \ref{FIG:Obj_and_Shape}.The training of the shape models converges stably.
\begin{figure}[htb]
\centering
\subfloat[]{\includegraphics[width=0.166\textwidth]{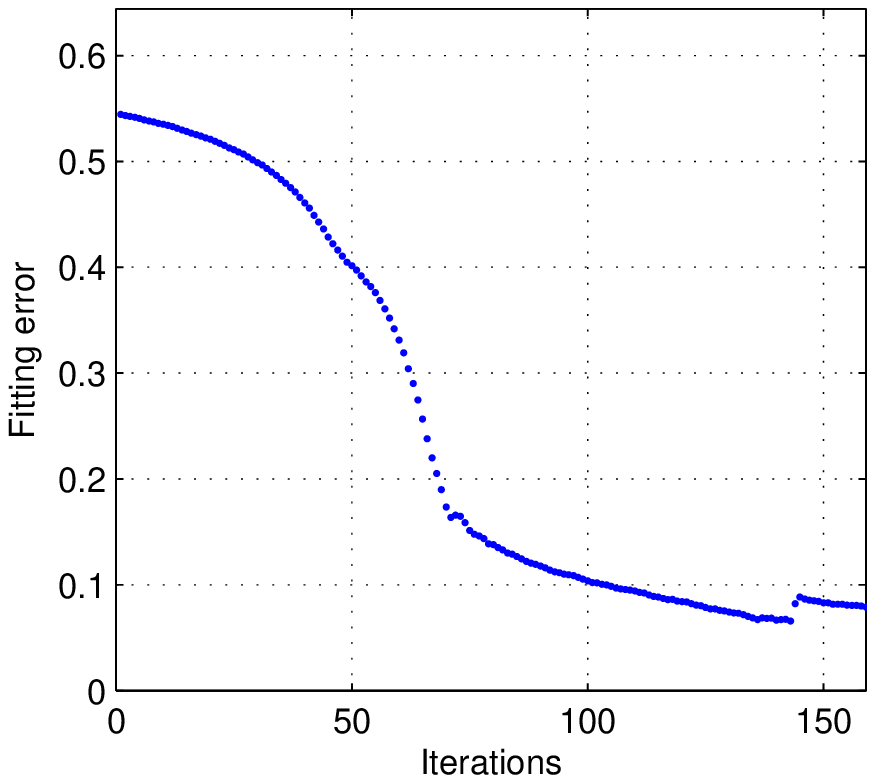}}
\subfloat[]{\includegraphics[width=0.166\textwidth]{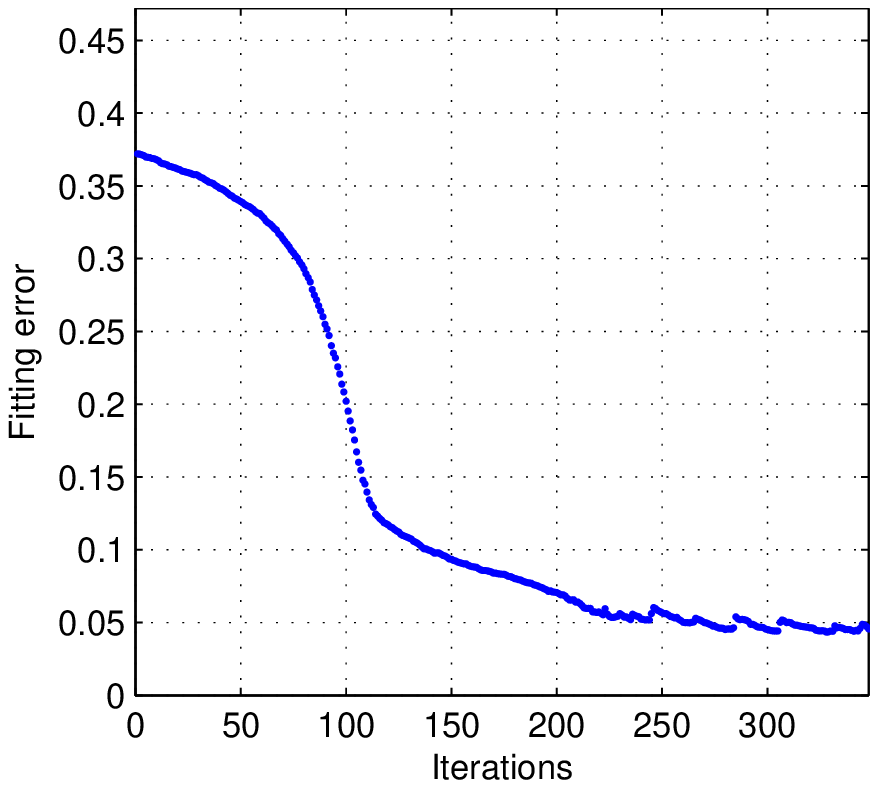}}
\subfloat[]{\includegraphics[width=0.166\textwidth]{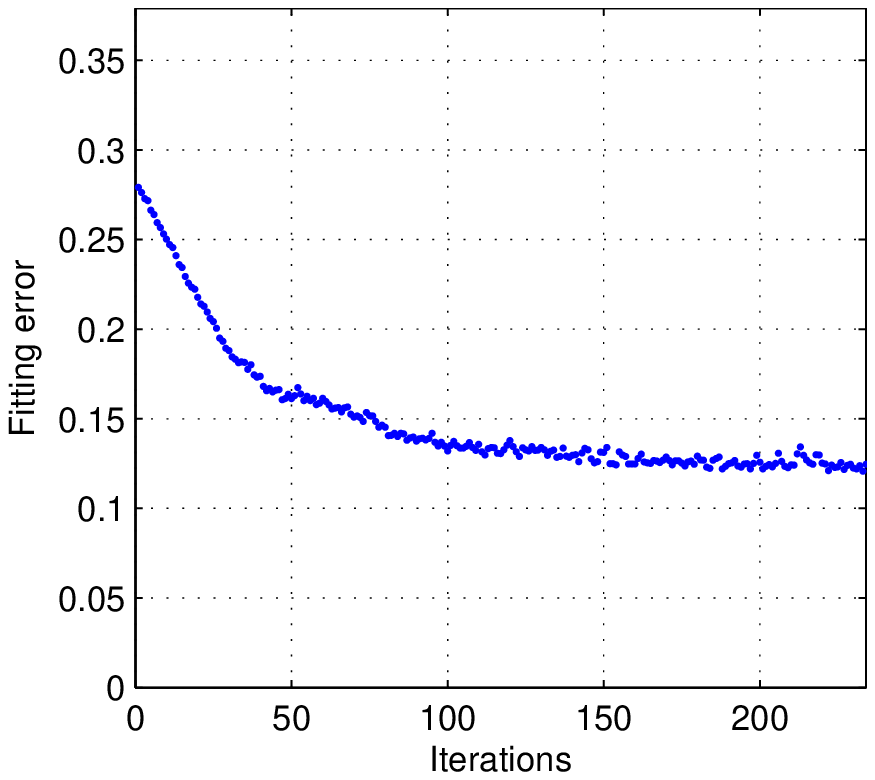}}
\caption{Minimization of the errors of shape modeling corresponding to Figs. \ref{FIG:Obj_and_Shape} (a) (b) and (c).}
\end{figure}

\begin{figure}
\centering
  \subfloat[]{\includegraphics[width=0.45\columnwidth]{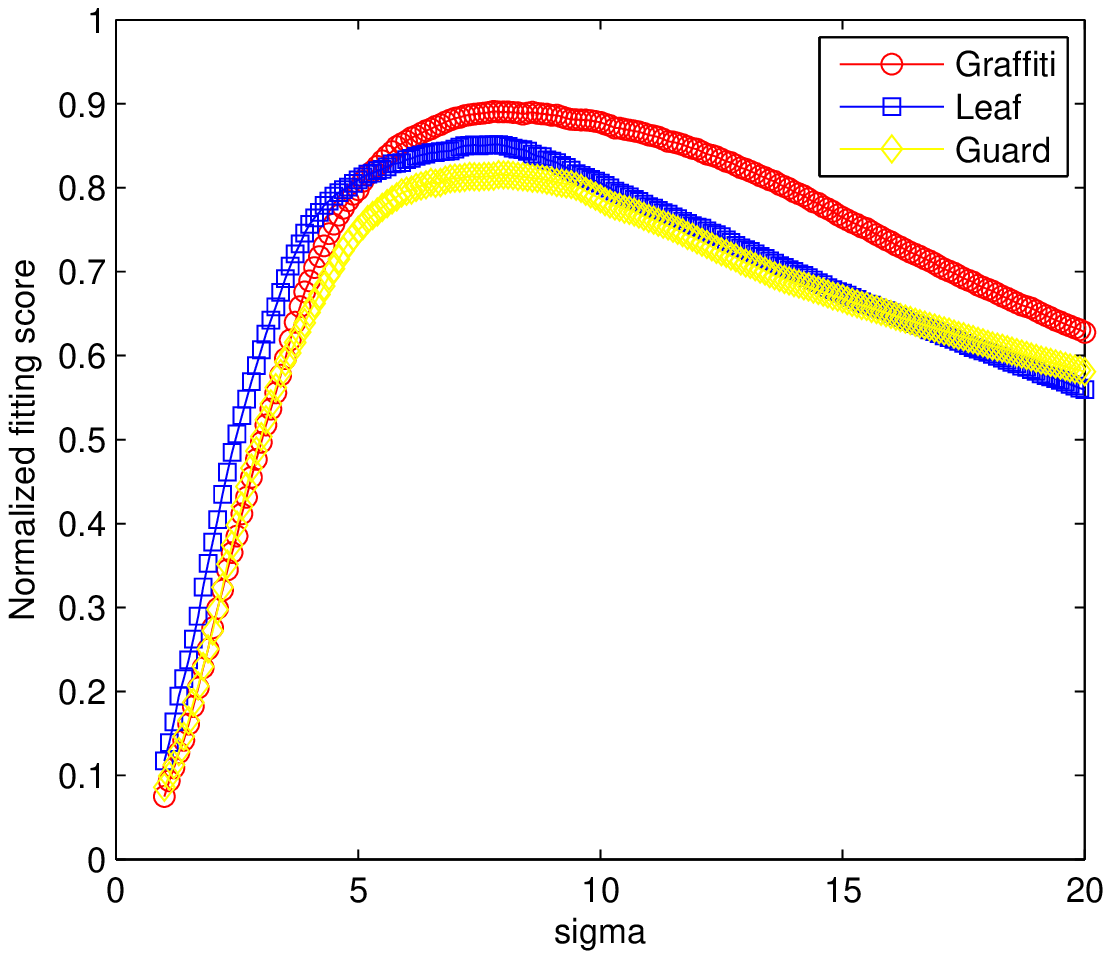}}
%\end{figure}
%
%\begin{figure}
%\centering
  \subfloat[]{\includegraphics[width=0.45\columnwidth]{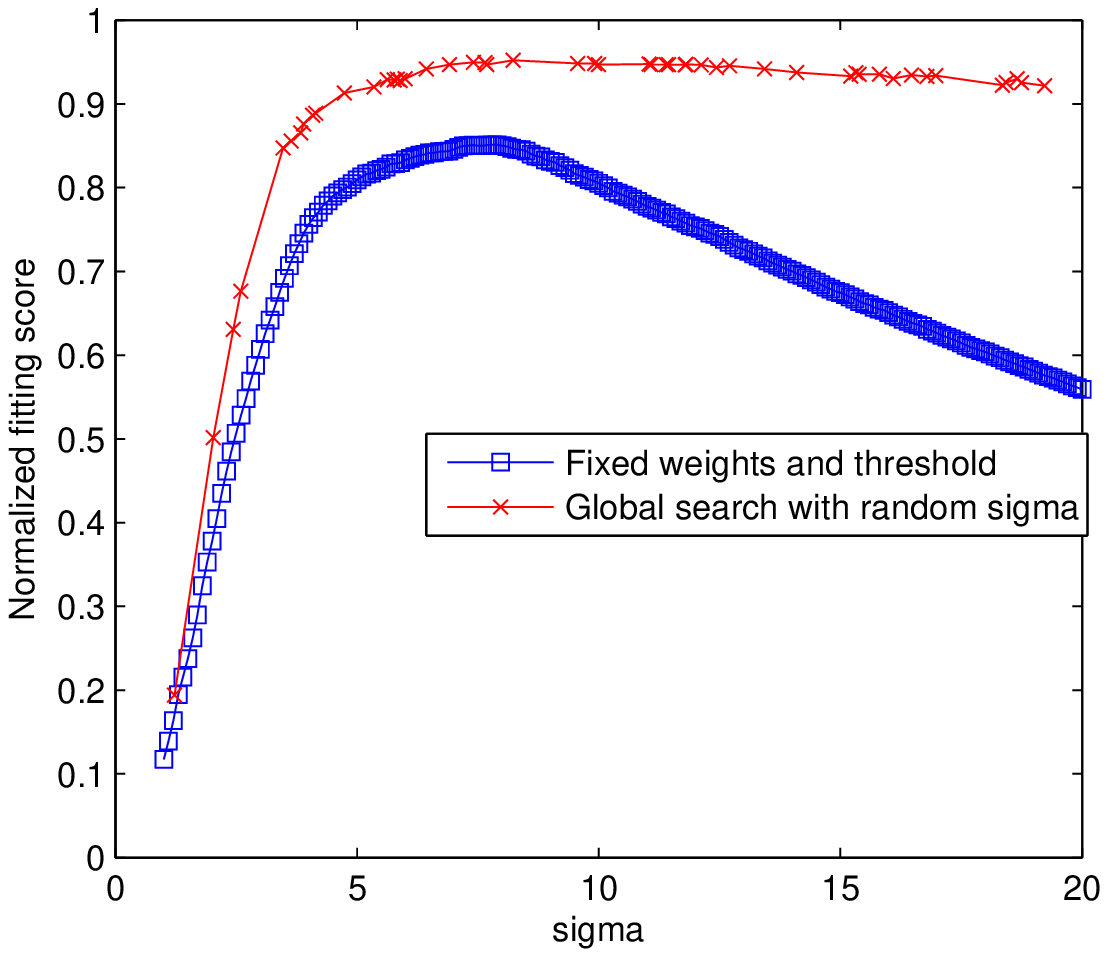}}
  \caption{Experimental results on selection of the parameter $\sigma$. (a) is the plot of fitting scores (vertical) v.s. $\sigma$ (horizontal), given the fixed initial weights and thresholds. (b) is the plot of fitting scores w.r.t. different values of $\sigma$ given the fixed initial weights and thresholds (squares with curve) comparing with the fitting scores due to the optimization of the weights and threshold given the randomly selected $\sigma$ (crosses with curve).}
\label{FIG:Select_sigma}
\end{figure}

There is a parameter $\sigma$ in the RBF based shape representation. We may use the fitting errors with fixed initial weights $\alpha_i$ and threshold $\beta$ for selecting $\sigma$. Specifically, we select $\sigma$ which corresponds to the highest fitting score ($1 - normalized~error$) from a set of $\{1,1.1,...,19.9,20\}$ containing $191$ candidates. Fig. \ref{FIG:Select_sigma}(a) shows the fitting scores w.r.t. $\sigma$ for the three shapes. We also randomly select $50$ values of $\sigma$ from $[1,20]$ and we implement the gradient descent learning to get the convergent shape models of the leaf shape corresponding to the $50$ random values of $\sigma$. The scores of the optimal fitting, in terms of the Jaccard shape similarity, given the $50$ random values of $\sigma$ are shown in Fig. \ref{FIG:Select_sigma}(b). We may observe that the peaks of the two curves are quite close, which means that the criteria for selecting $\sigma$ with fixed weights and threshold is effective. The major benefit of the selection of $\sigma$ before the shape model fitting is the computational efficiency.

\begin{figure*}[htb]
\centering
\subfloat[]{\includegraphics[scale=0.45]{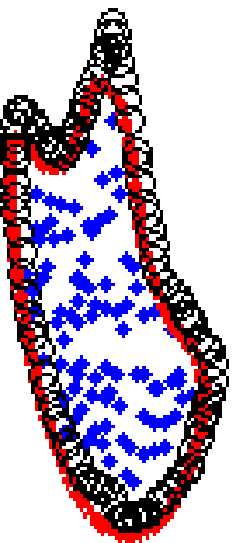}}
\subfloat[]{\includegraphics[scale=0.45]{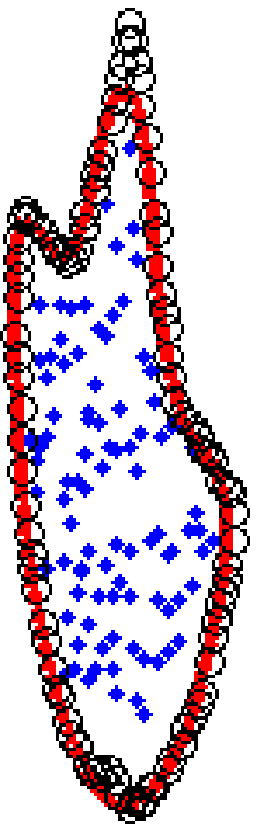}}
\subfloat[]{\includegraphics[scale=0.45]{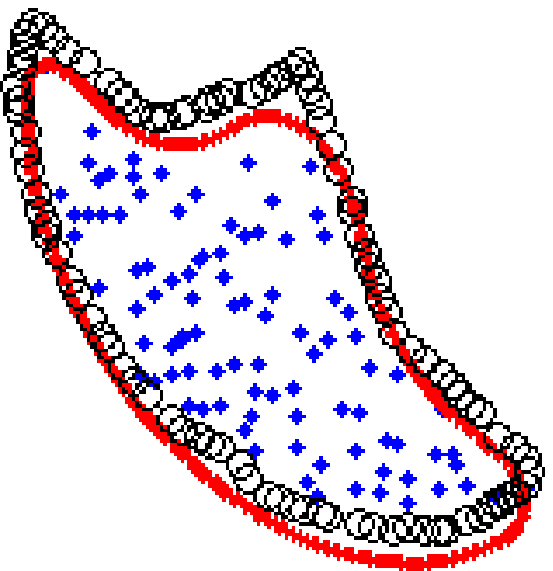}}
\subfloat[]{\includegraphics[scale=0.45]{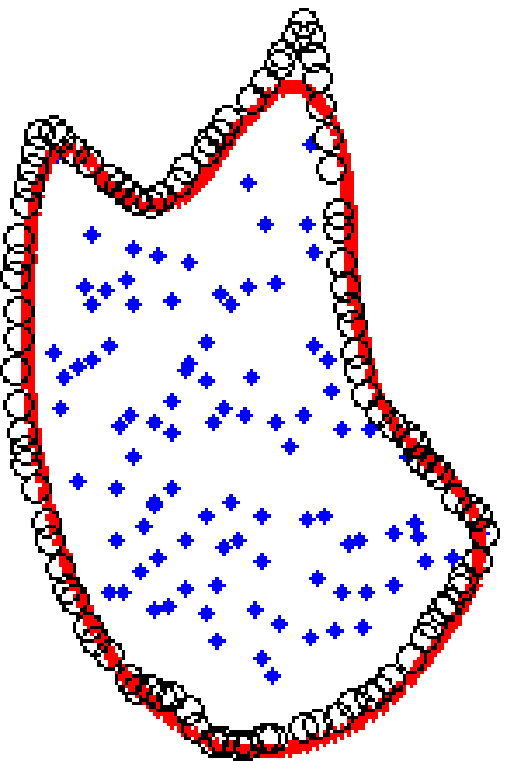}}
\subfloat[]{\includegraphics[scale=0.45]{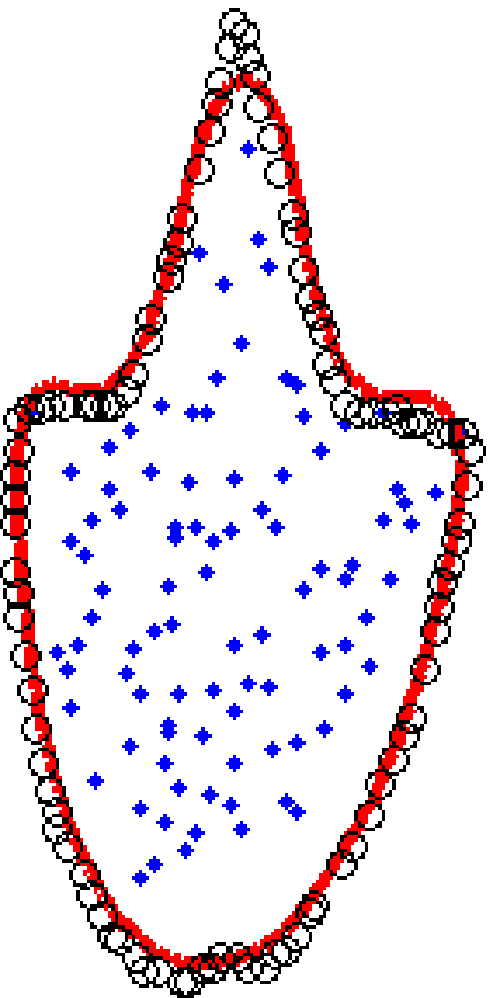}}
\subfloat[]{\includegraphics[scale=0.45]{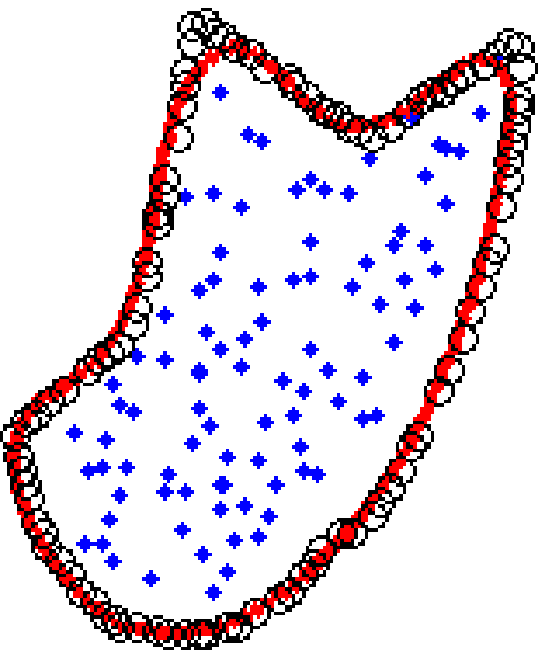}}
\subfloat[]{\includegraphics[scale=0.45]{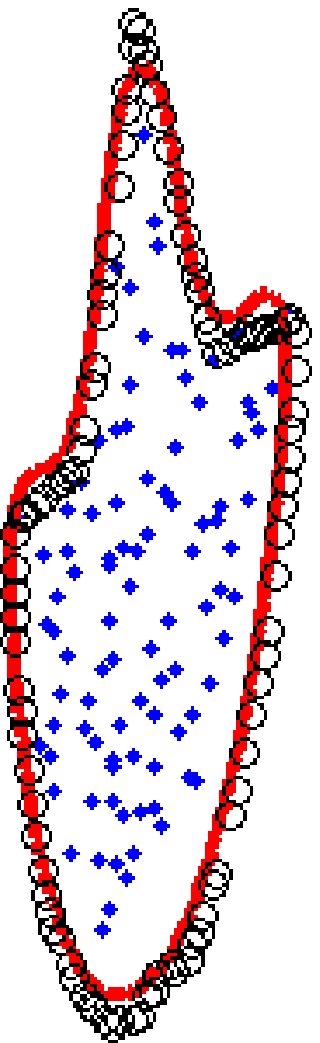}}
\subfloat[]{\includegraphics[scale=0.45]{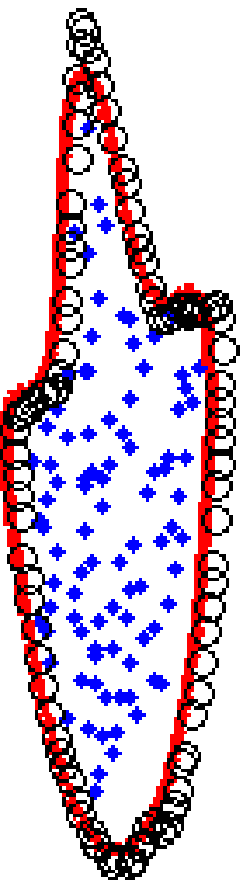}}
\subfloat[]{\includegraphics[scale=0.45]{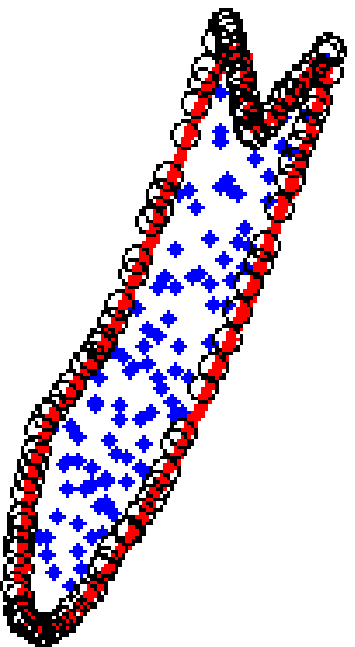}}\\
\subfloat[]{\includegraphics[scale=0.45]{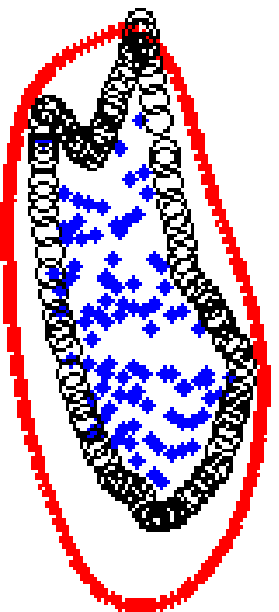}}
\subfloat[]{\includegraphics[scale=0.45]{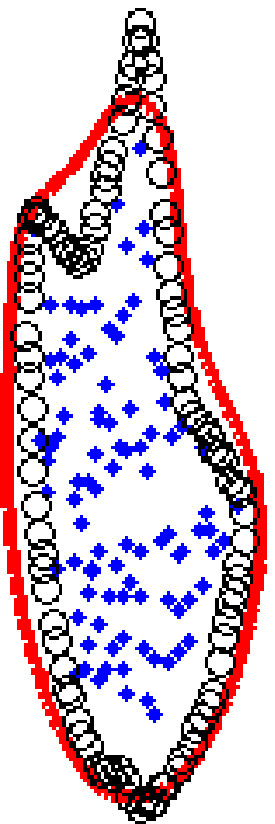}}
\subfloat[]{\includegraphics[scale=0.45]{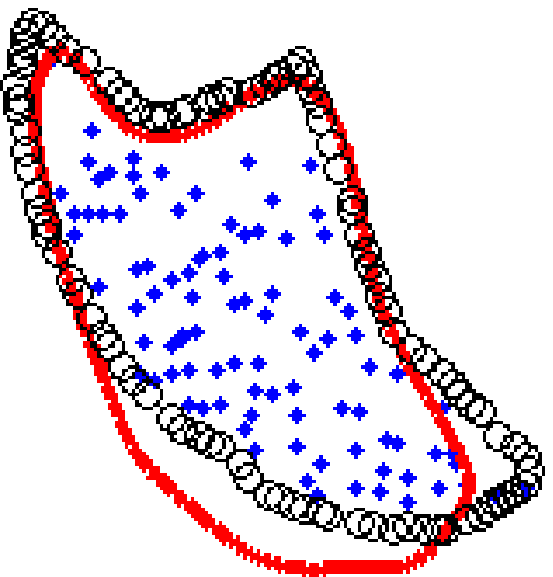}}
\subfloat[]{\includegraphics[scale=0.45]{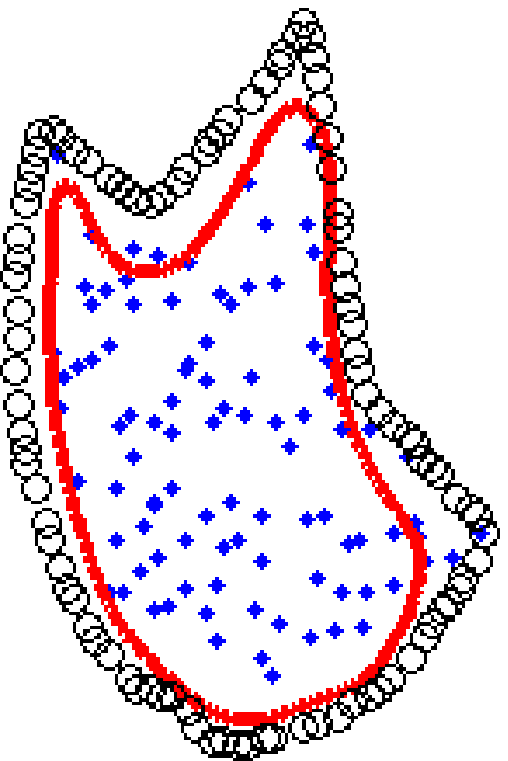}}
\subfloat[]{\includegraphics[scale=0.45]{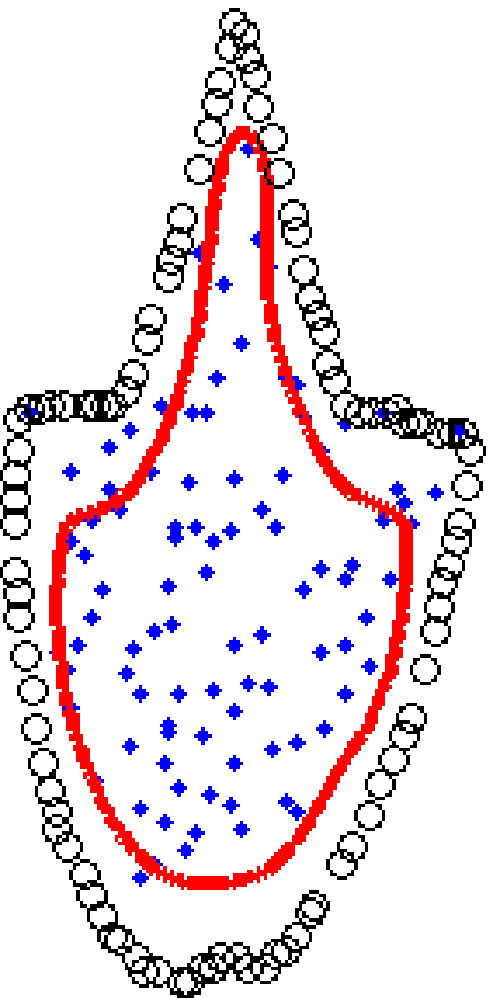}}
\subfloat[]{\includegraphics[scale=0.45]{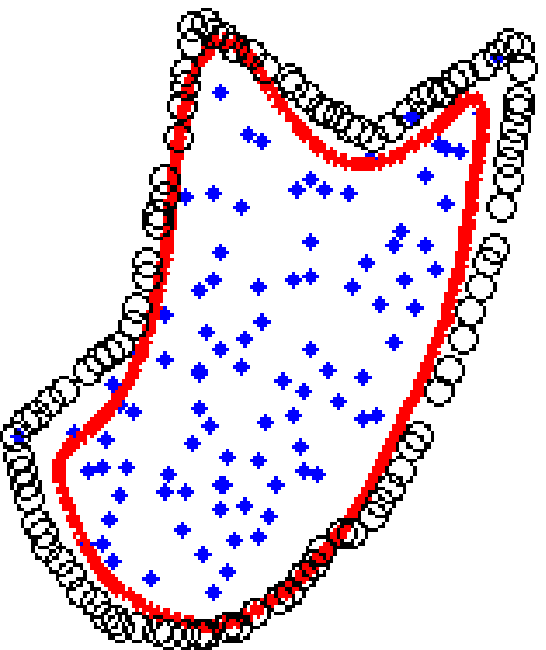}}
\subfloat[]{\includegraphics[scale=0.45]{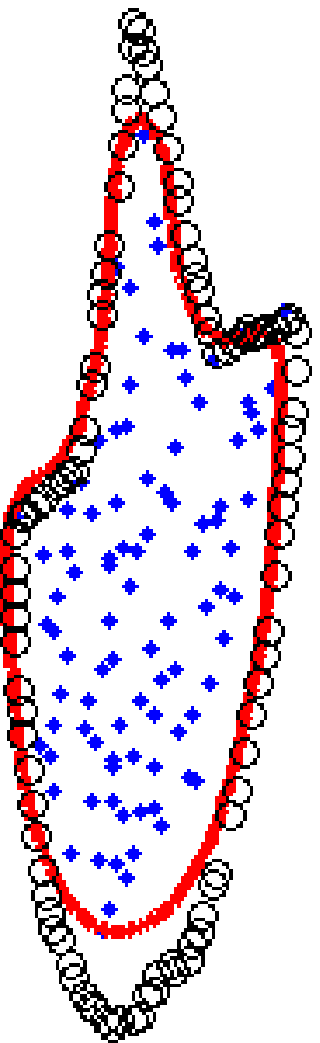}}
\subfloat[]{\includegraphics[scale=0.45]{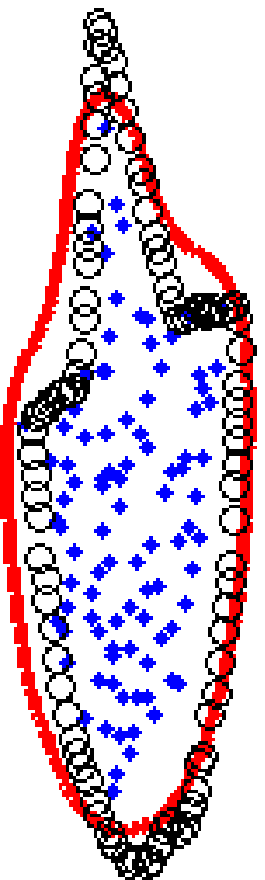}}
\subfloat[]{\includegraphics[scale=0.45]{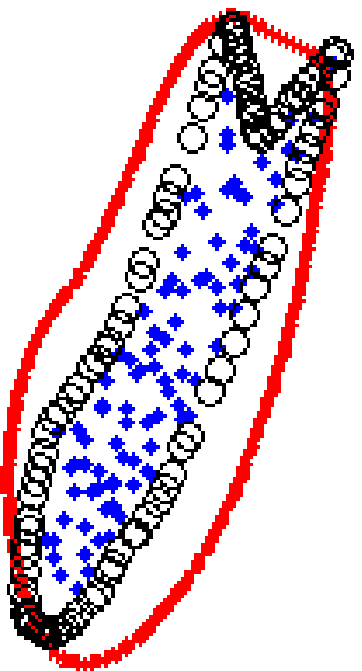}}\\
\caption{The significance of the affine invariance of the interior-points-to-shape relation. The top row shows shapes generated by the affine invariant model , i.e. Eq. (\ref{EQ:phi_S}). The bottom row shows the shapes generated by the model without affine invariance, i.e. Eq. (\ref{EQ:Shape_Dec_Trans}). The dots are the interior feature points, the curves are the implicit contours, and the circles are the explicitly transformed contours. }\label{FIG:Exp_Aff_Invar}
\end{figure*}

\begin{figure}[htb]
\centering
\includegraphics[width=0.45\textwidth]{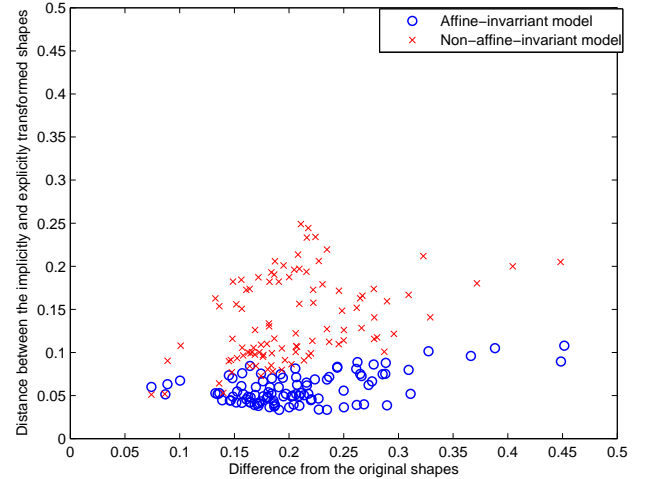}\\
  \caption{The normalized Hausdorff distances between the shapes implicitly defined by the transformed interior points and the original shapes (horizontal) v.s. the distance between the explicitly transformed contour shapes and the original shape (vertical).}\label{FIG:H_dist_all}
\end{figure}

Upon the accomplishment of the shape model training, we then validate the claimed affine invariance of the shape model defined in Eq. (\ref{EQ:phi_S}). We take the trained leaf shape model for evaluation. The principle is to compare the contour shape implicitly defined by the transformed interior points with the explicitly transformed contour shape, which could be considered as the ground truth shapes. We choose the normalized Hausdorff distance, i.e. the Hausdorff distance normalized by the maximum distance between the two point sets, as the shape distance. We randomly generate the $100$ sets of parameters for the affine transformation without translation, leading up to $100$ randomly generated leaves. Some examples of the implicit contours overlaying the explicitly transformed contour points are shown in Figs. \ref{FIG:Exp_Aff_Invar}(a)-(i). We can observe that the proposed interior-points-to-shape relation defined in Eq. (\ref{EQ:phi_S}) is affine invariant. In Figs. \ref{FIG:Exp_Aff_Invar}(j)-(r). We also present the implicit contours due to the shape model without affine invariance, i.e. Eq. (\ref{EQ:Shape_Dec_Trans}). The shape distances corresponding to Fig. \ref{FIG:Exp_Aff_Invar} are summarized in Table \ref{TB:H_dists_exps}. In Fig. \ref{FIG:H_dist_all}, we plot the shape distances corresponding to the two shape models for all the $100$ examples. We observe that even if the shape transformation is large, the distance between the explicitly and implicitly transformed contour shapes is still small for the affine-invariant shape model, whereas the shape model without the affine invariance deviates from the explicitly transformed shapes a lot. The explicitly transformed shapes can be considered as the ground-truth shapes.
\begin{table*}[htb]
\centering
\caption{Shape distances corresponding to Fig. \ref{FIG:Exp_Aff_Invar}.}\label{TB:H_dists_exps}
\begin{tabular*}{\textwidth}{@{\extracolsep{\fill}}l|ccccccccc}
\toprule
Fig. \ref{FIG:Exp_Aff_Invar}  &  (a) & (b) & (c) & (d) & (e) & (f) & (g) & (h) & (i) \\ \hline
NHD&0.108&0.086&0.046&0.054&0.053&0.033&0.049&0.056&0.036\\\hline
Fig. \ref{FIG:Exp_Aff_Invar}  &  (j) & (k) & (l) & (m) & (n) & (o) & (p) & (q) & (r)\\ \hline
NHD&0.180&0.116&0.113&0.130&0.154&0.080&0.100&0.114&0.126\\
\bottomrule
\end{tabular*}
{\begin{flushleft}
\scriptsize{NHD = Normalized Hausdorff distance}
\end{flushleft}}
\end{table*}

%\subsection{The importance of good initialization to active contours}
%In this subsection, we elucidate the necessity for a good initialization in our proposed affine active contour for segmentation.
%
%In Fig. \ref{FIG:demo_init_AffAC}, we show that the result by using the affine active contour is bad given an initial contour deviating from the object, whereas the result is good given the initial contour shape well-aligned with the object in the image. These all elucidate that a good initialization is crucial to the conventional active contours which are based on gradient descent curve evolution.
%\begin{figure}[htb]
%\centering
%\subfloat[]{\includegraphics[width=0.25\columnwidth]{Imgs_Leaves_ASP_demos_bad_init_MCAC_Iter_1.eps}}
%\subfloat[]{\includegraphics[width=0.25\columnwidth]{Imgs/Leaves_ASP/demos_bad_init_MCAC_Iter_39.eps}}
%\subfloat[]{\includegraphics[width=0.25\columnwidth]{Imgs_Leaves_ASP_demos_manual_init_MCAC_Iter_1.eps}}
%\subfloat[]{\includegraphics[width=0.25\columnwidth]{Imgs_Leaves_ASP_demos_manual_init_MCAC_Iter_200.eps}}
%\caption{The object segmentation by affine active contour driven by affine motion of interior points. (a) is a bad initialization which leads to a bad result shown in (b). (c) is a good initialization which leads to a good result shown in (d).}\label{FIG:demo_init_AffAC}
%\end{figure}

\subsection{The consequence of globally minimizing improper active contour energies}
There could be various global optimization strategies for active contours such as those reported in \cite{Cremers08GlobalSP} \cite{Schoenemann10}. In this subsection, we show that the global optimal solution to improper active contour energies may not correspond to the target object. This claim should not depend on the choice of the optimization method. Hence, we adopt the exhaustive search for the optimization. To examine the global optimality of object shape in image for a given active contour energy, we ensure that the object shape in the image has been included in the search space.

\begin{figure}[htb]
\centering
\subfloat[]{\includegraphics[width=0.4\columnwidth, height=1.3in]{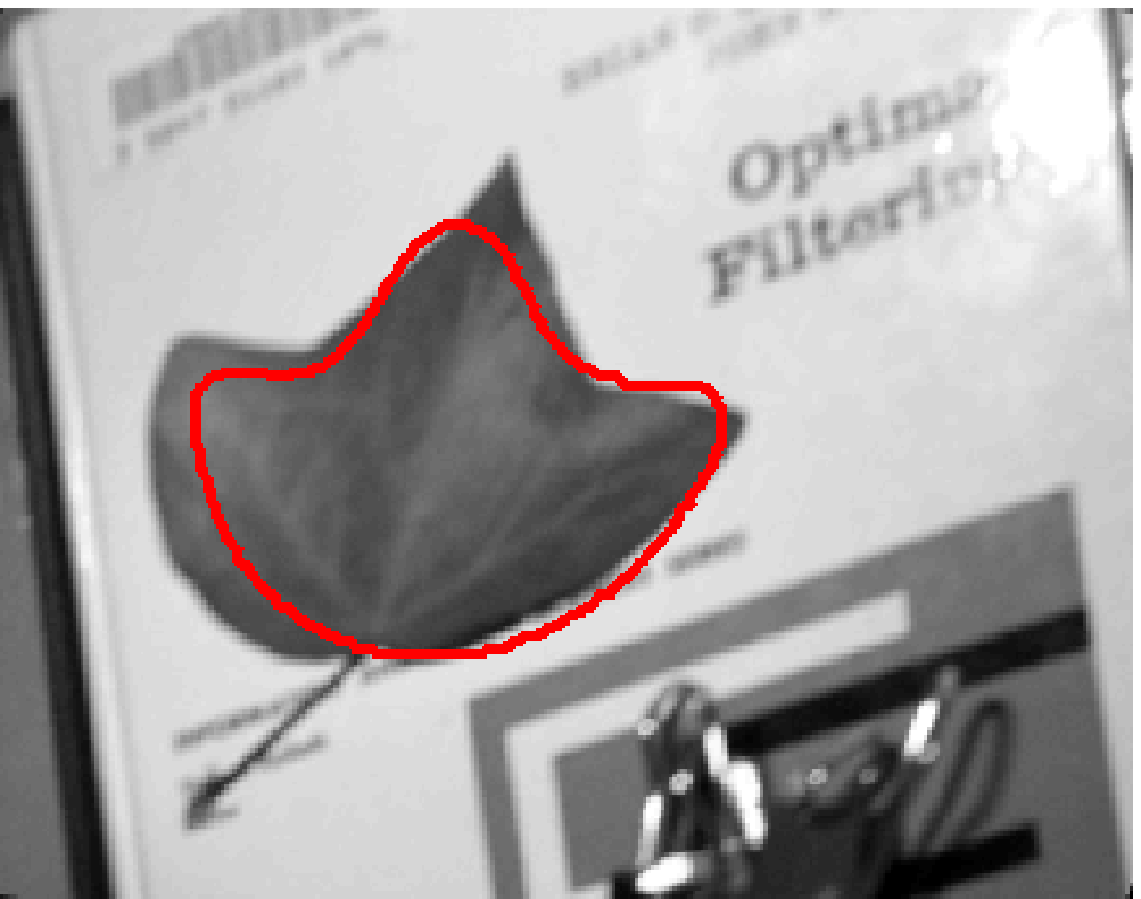}}
\subfloat[]{\includegraphics[width=0.4\columnwidth,height=1.3in]{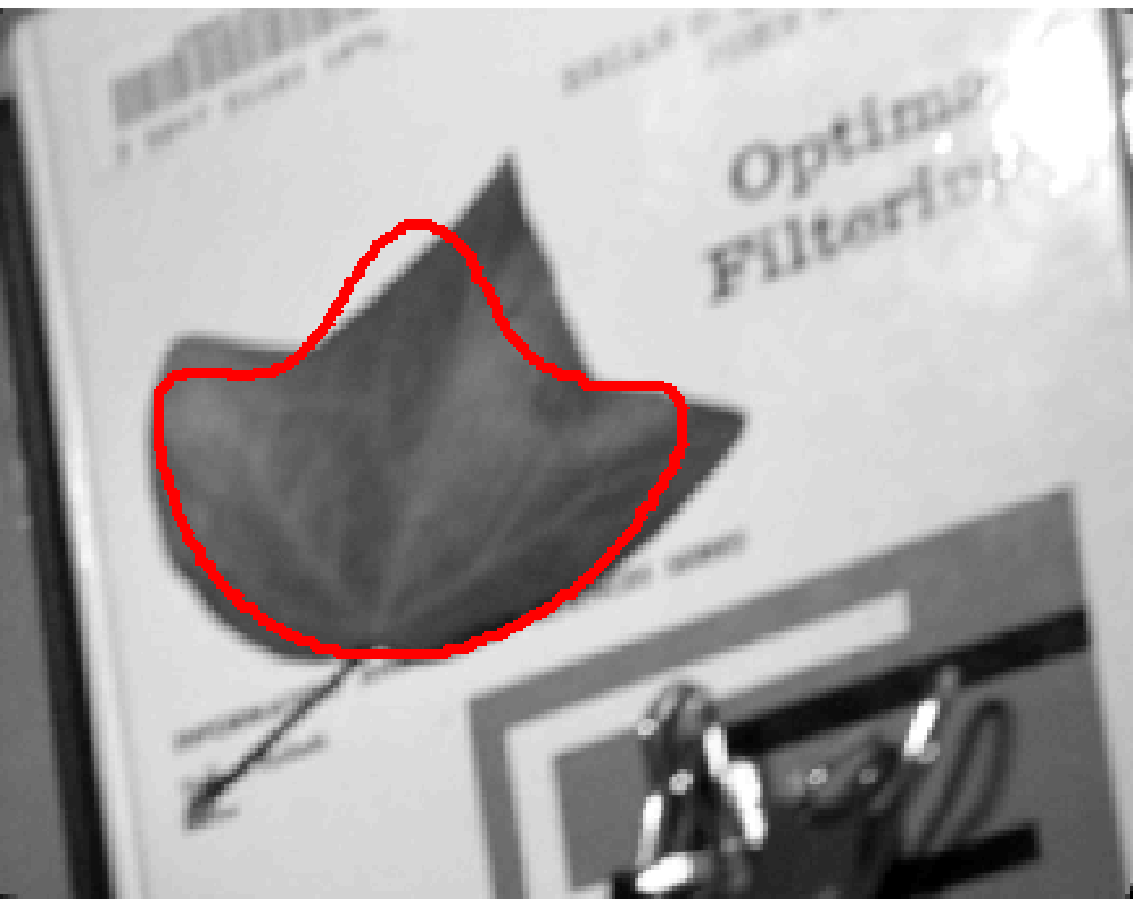}}
\caption{The object segmentation by exhaustive search in simple image. (a) is the result by global search with Chan-Vese model (b) is the result by global search with GAC model }\label{FIG:global_good}
\end{figure}

In this experiment, we perform exhaustive search for both the Chan-Vese model and the GAC model with hard but correct shape priors. In the implementation, we search over 8 orientations of the given shape, i.e. $\{0,{\pi\over4},{\pi\over2},{3\pi\over4},\pi,{5\pi\over4},{3\pi\over2},{7\pi\over4}\}$, at all possible locations within the image domain. We make sure that this relatively small shape space covers roughly the correct object shape. A result is shown in Fig. \ref{FIG:global_good}. In this experiment, we have fixed the size of the shapes.

\begin{figure}[htb]
\centering
\subfloat[]{\includegraphics[width=0.4\columnwidth]{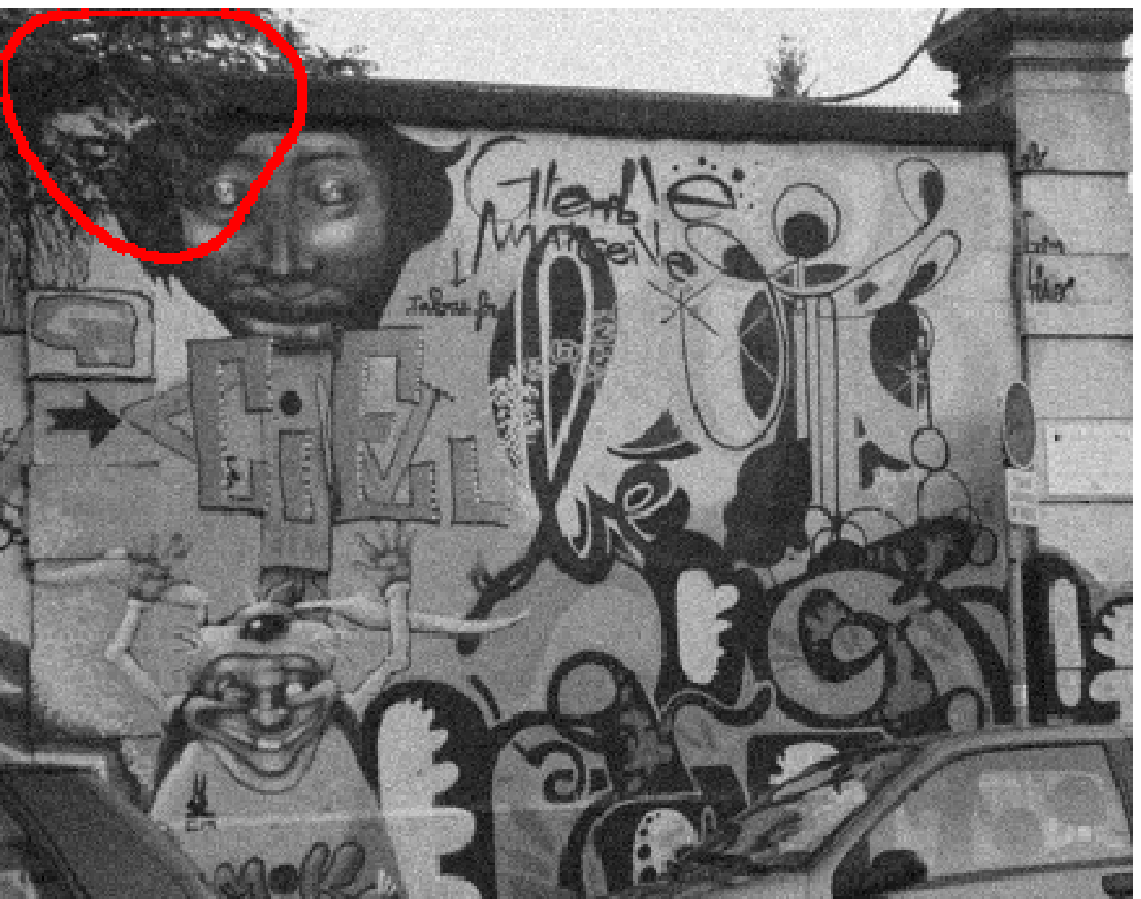}}
\subfloat[]{\includegraphics[width=0.4\columnwidth]{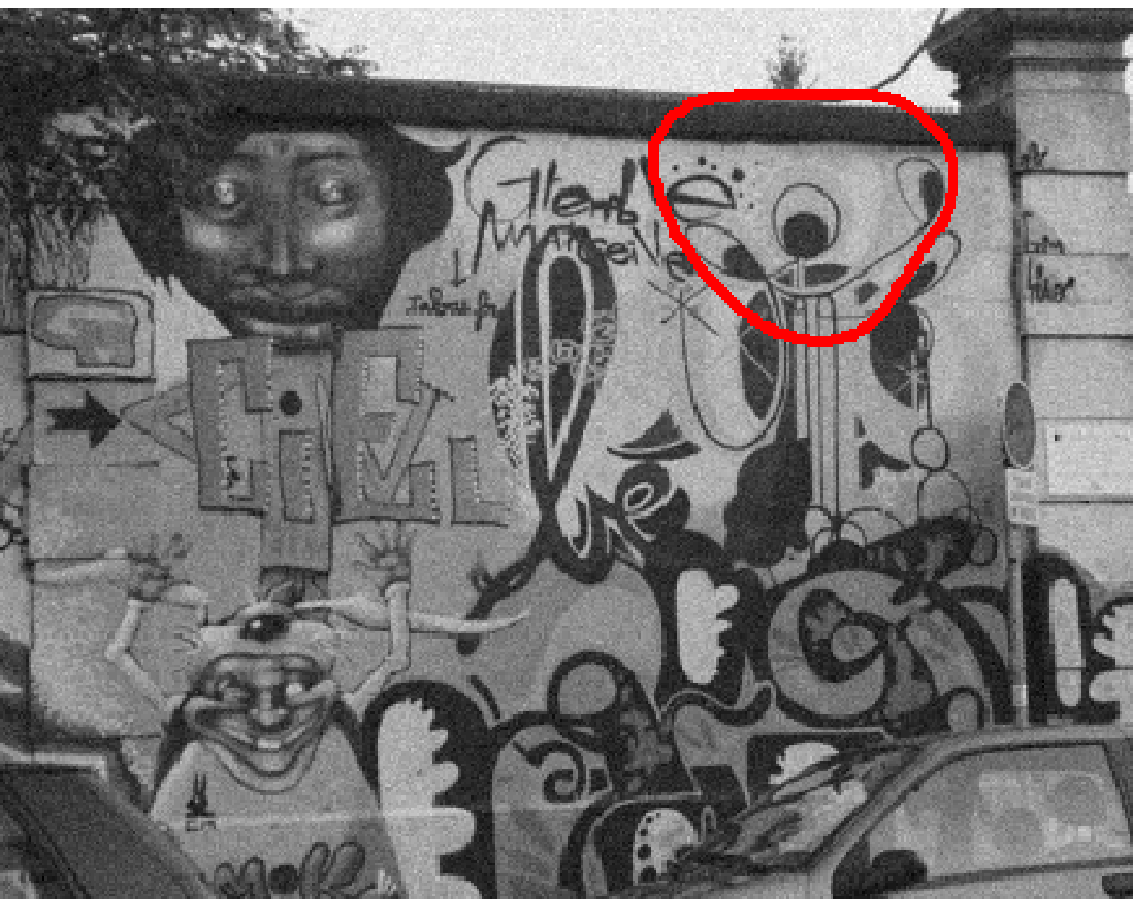}}
\caption{The object segmentation by exhaustive search in a cluttered image. (a) is the result by global search with Chan-Vese model (b) is the result by global search with GAC model }\label{FIG:global_bad}
\end{figure}

We also show that given the correct fixed size, the global search also may not provide a reliable result of object segmentation in Fig. \ref{FIG:global_bad}. We can observe that the global optimal solution to the Chan-Vese model (at the top-left corner) provides the region in which the image values contrast the outer region most, and the GAC model locates the group of the strongest edges that fit the shape prior best while not necessarily being the target object. Neither of the results are satisfactory. This is because we are not able to ensure the object of interest to correspond to the globally optimal solution of the formulated energy minimization problem.

\begin{figure}[htb]
\centering
\subfloat[]{\includegraphics[width=0.4\columnwidth]{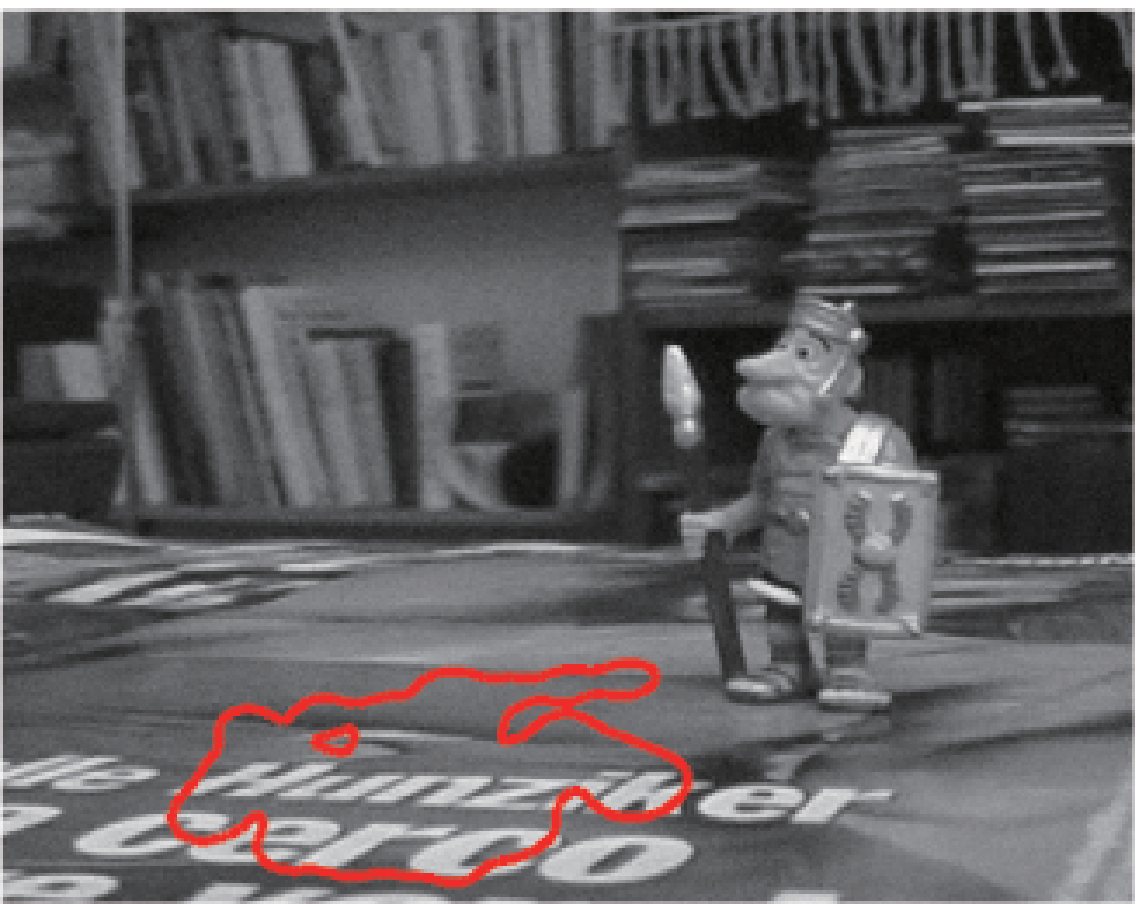}}
\subfloat[]{\includegraphics[width=0.4\columnwidth]{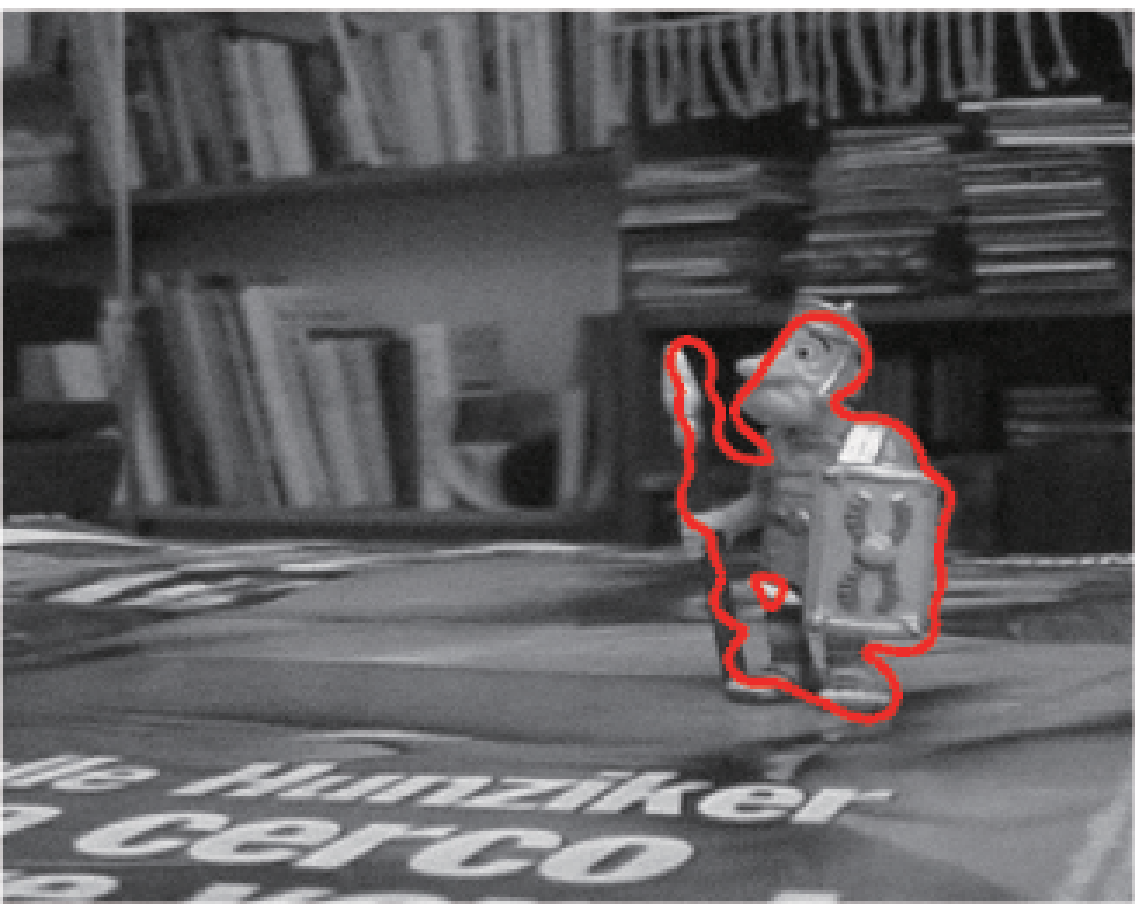}}\\

\caption{The object segmentation by exhaustive search in a cluttered image containing single object, with the correct object size. (a) is the result by global search with Chan-Vese model (b) is the result by global search with GAC model }\label{FIG:global_ok}
\end{figure}

\begin{figure}[htb]
\centering
\subfloat[]{\includegraphics[width=0.4\columnwidth]{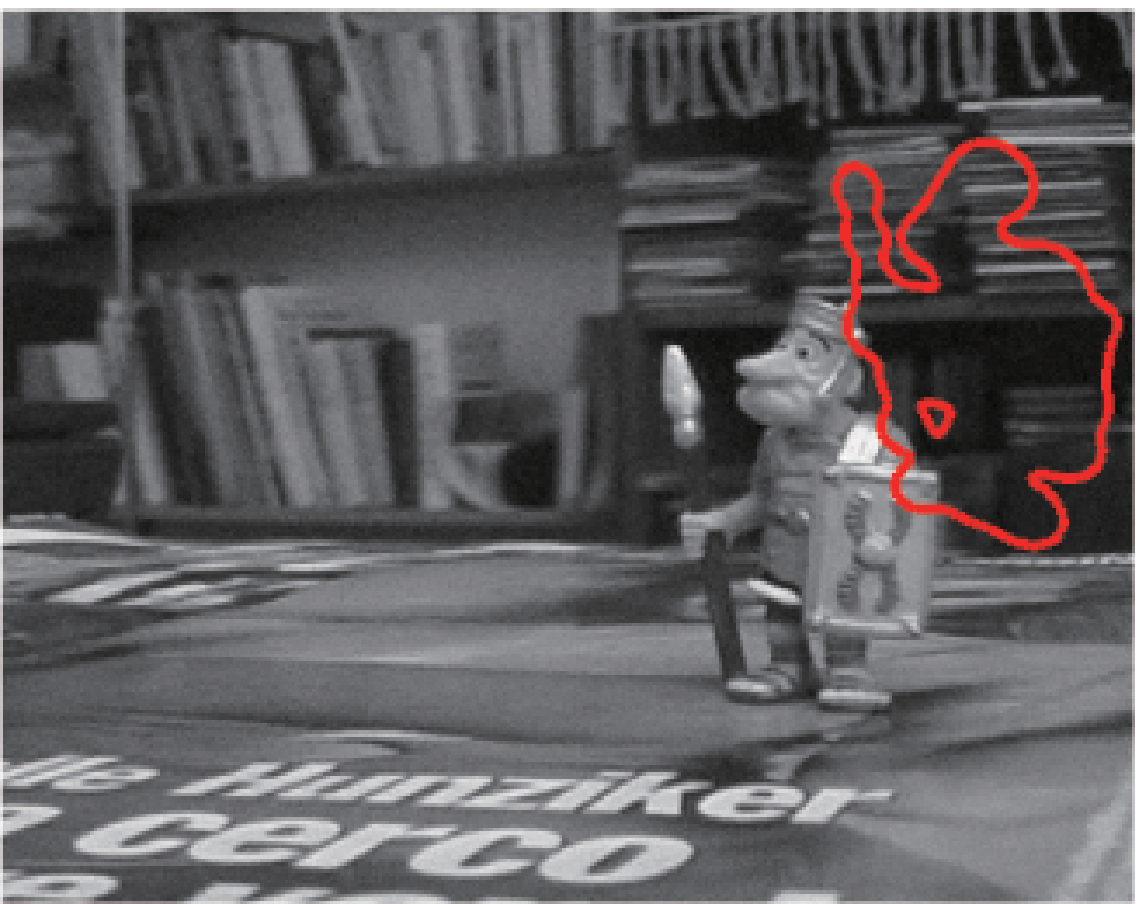}}
\subfloat[]{\includegraphics[width=0.4\columnwidth]{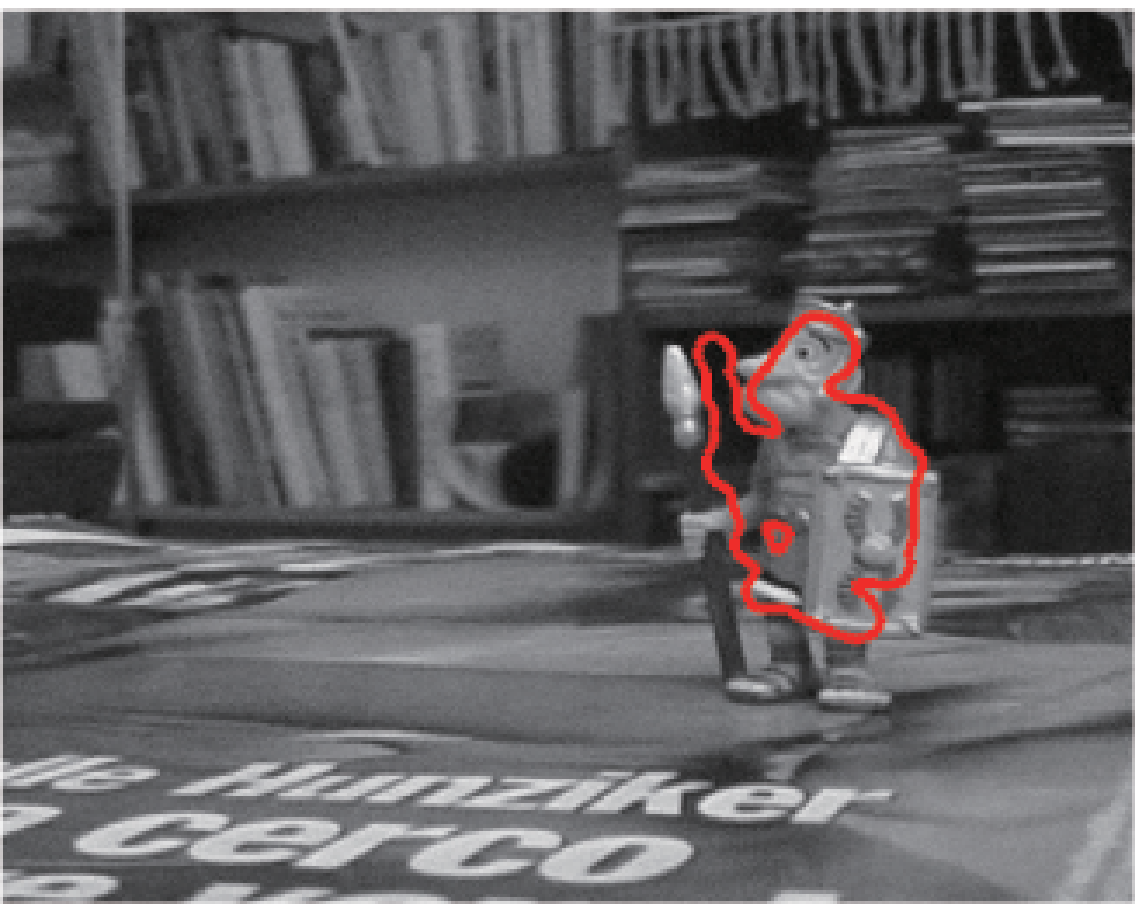}}\\

\caption{The object segmentation by exhaustive search in a cluttered image containing single object, without knowing the size. (a) is the result by global search with Chan-Vese model (b) is the result by global search with GAC model }\label{FIG:global_nosize}
\end{figure}

There definitely exist cases that the global search can output satisfactory segmentation in relatively complex images given the correct size of the object. An example is shown in Fig. \ref{FIG:global_ok} (b), which is a result of the global optimization for the GAC model. However, the global optimization of the Chan-Vese model is unsatisfactory, since the object of interest does not have significant contrast against the background. If we allow the size to vary in the search space, the results are often undesirable, such as in Fig. \ref{FIG:global_nosize}. In this experiment, we include the correct size with additionally one smaller and one larger sizes in the search space. The boundary of the template deviates a bit from the object boundary in the result of global optimization for GAC in Fig. \ref{FIG:global_nosize}(b), since the gradients on the object boundary are small while the gradients inside the object are significant. The result of segmentation by Chan-Vese model is in Fig. \ref{FIG:global_nosize}(a).

The experimental results in this subsection show that if the energy measure is improper, the global optimal solution of the active contour does not necessarily correspond to the desired object boundary.

\subsection{Automatic object segmentation by matching-constrained active contour}
In this subsection, we evaluate the matching-constrained active contour. The centroids of the template objects are set to be the origin $(0,0)$. This position can be anywhere, and it does not affect the feature point matching since the formulation of feature point matching does not involve the absolute position.

\begin{figure}[htb]
\centering
  \subfloat{\includegraphics[width=0.33\columnwidth,height=0.75in]{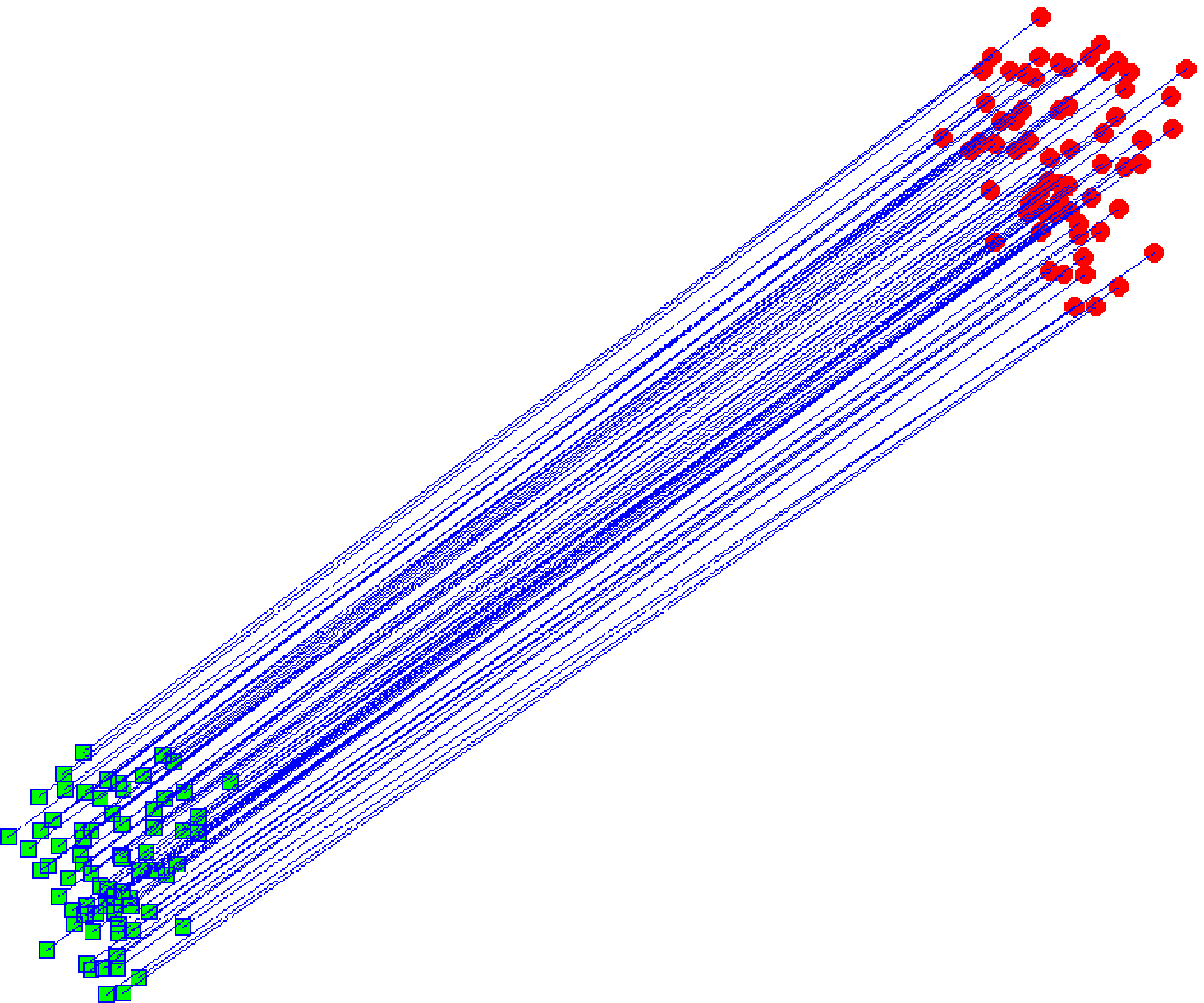}
            \includegraphics[width=0.33\columnwidth]{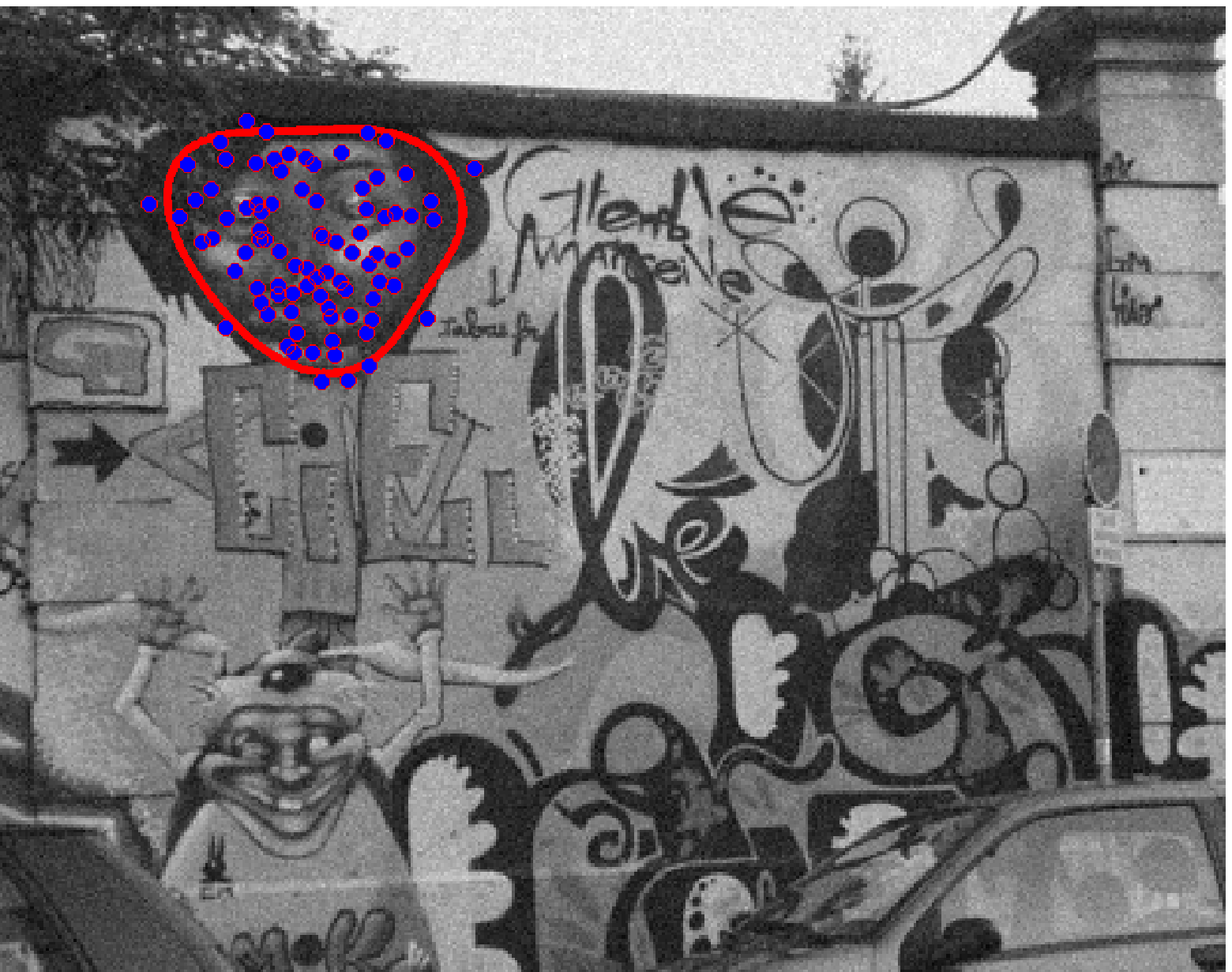}
            \includegraphics[width=0.33\columnwidth]{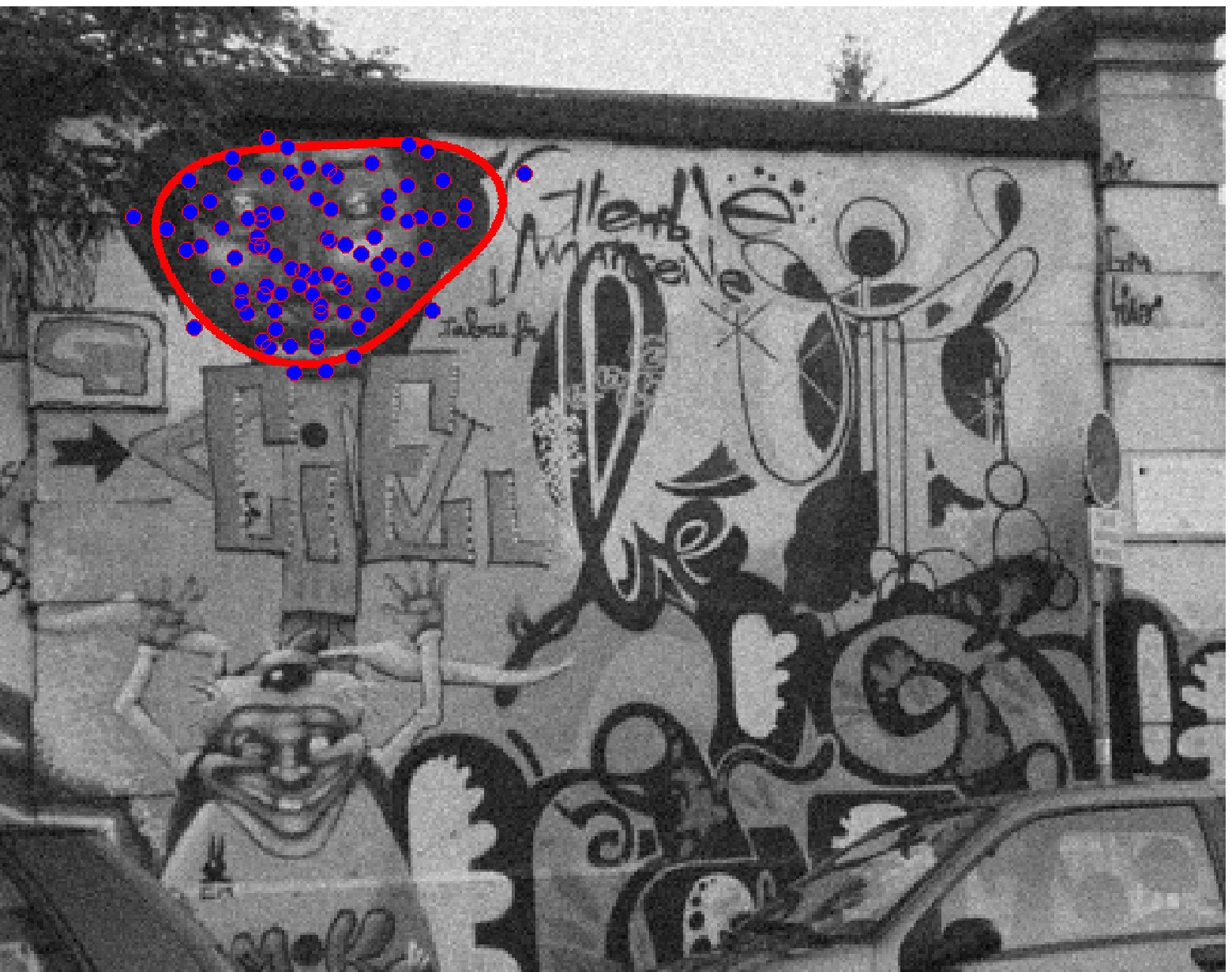}}\\
  \subfloat{\includegraphics[width=0.33\columnwidth,height=0.65in]{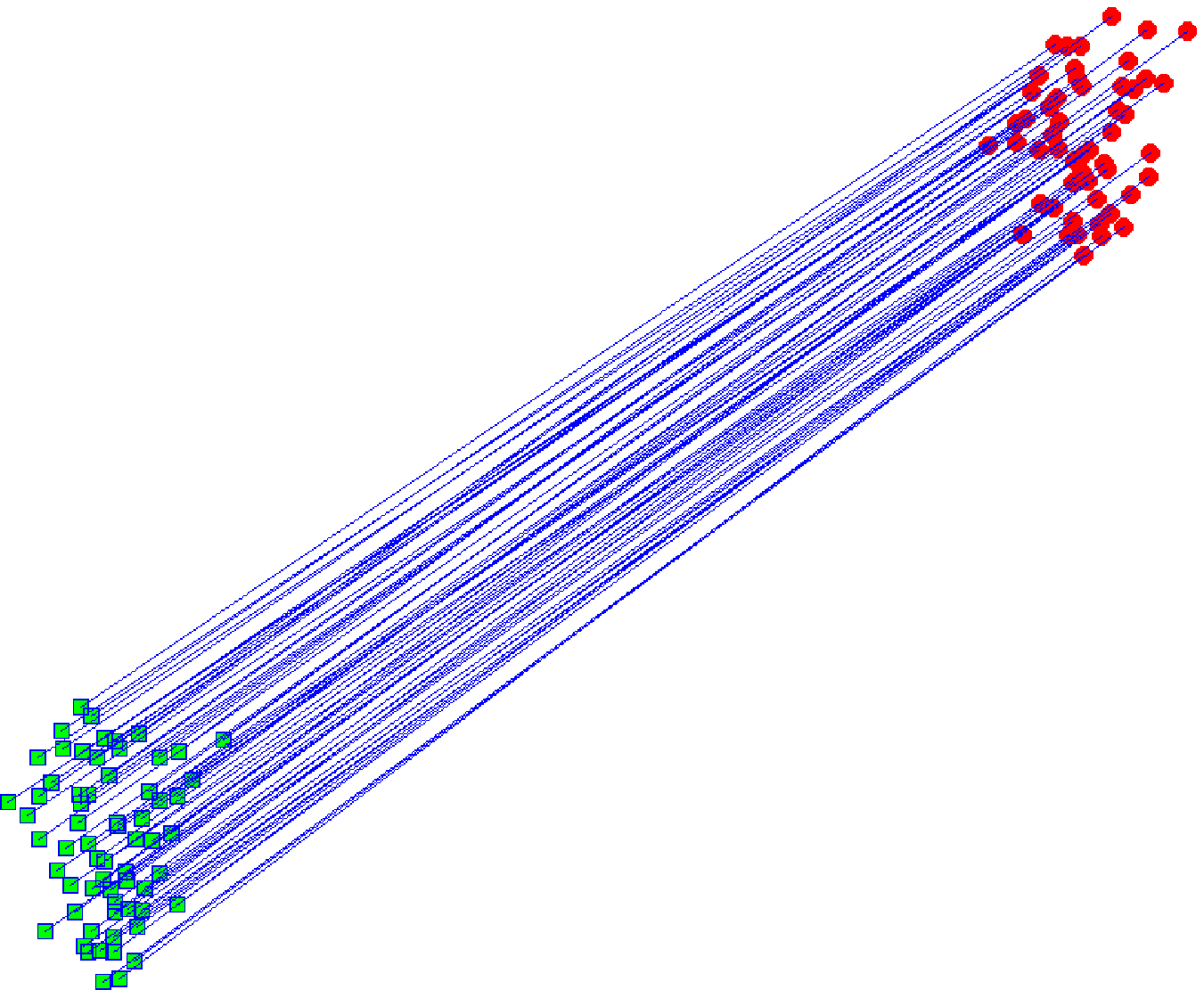}
            \includegraphics[width=0.33\columnwidth]{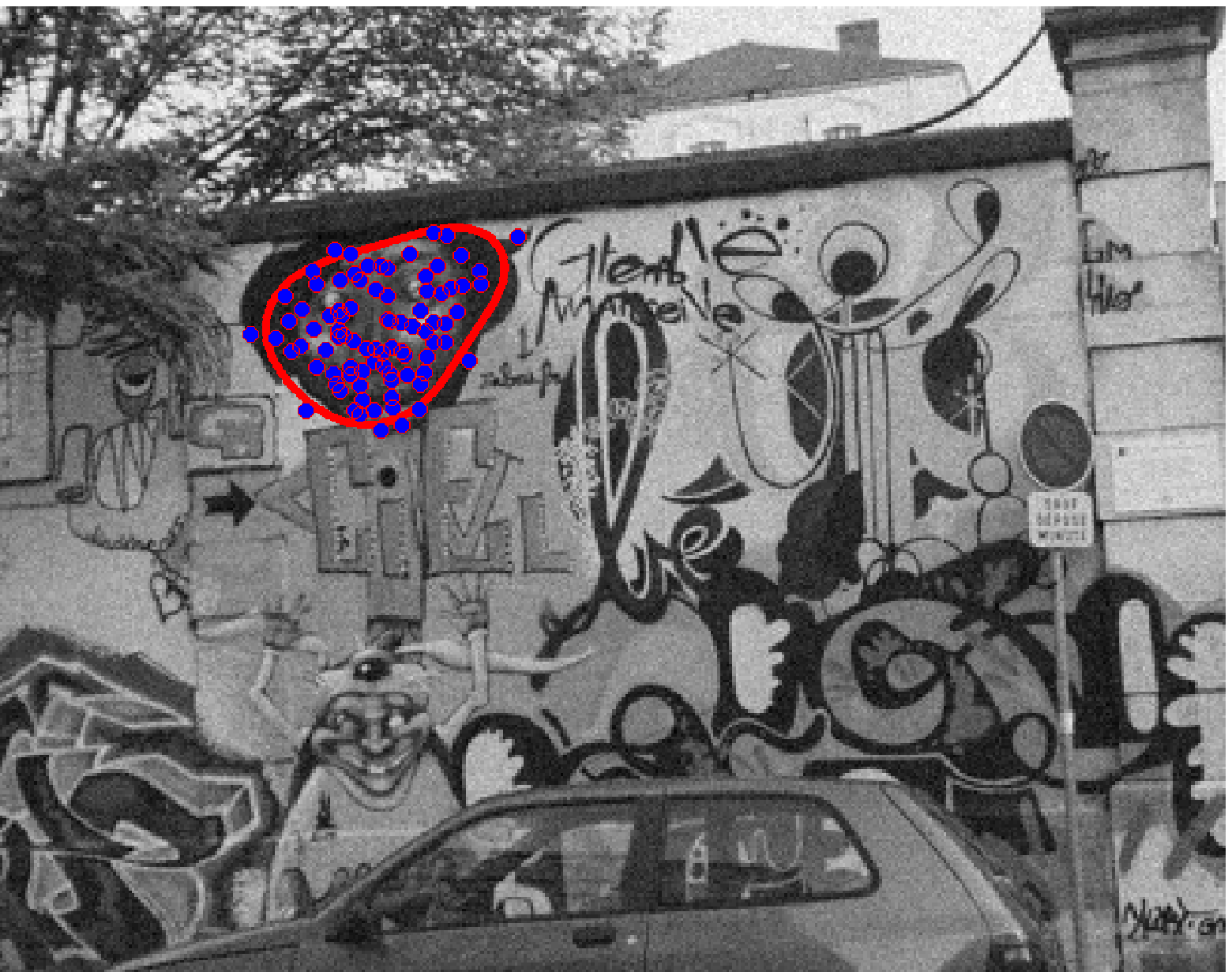}
            \includegraphics[width=0.33\columnwidth]{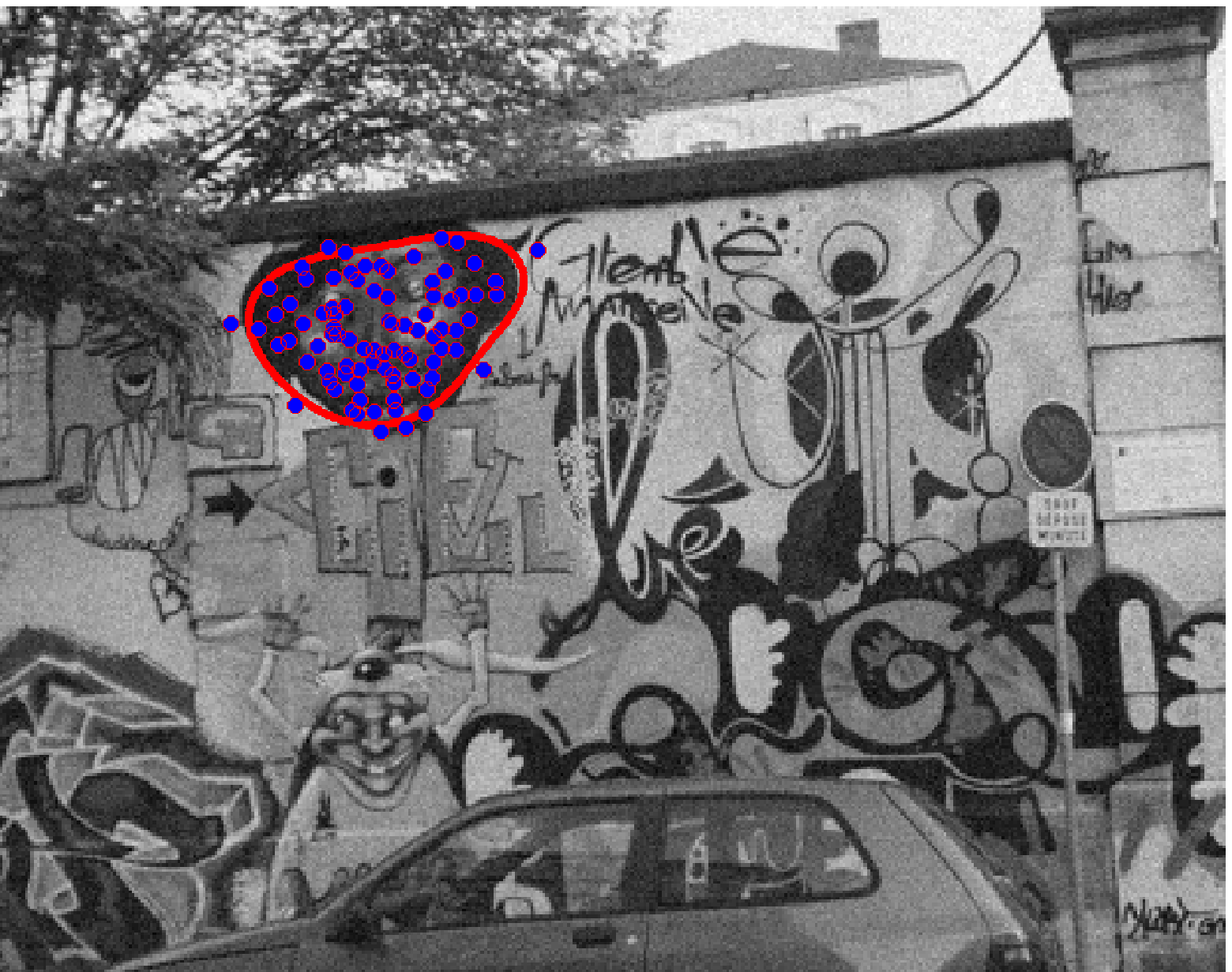}}\\
%  \subfloat[]{\includegraphics[width=0.33\columnwidth,height=0.7in]{Imgs_show_result_demo_result_N5_match.eps}
%            \includegraphics[width=0.33\columnwidth]{Imgs_show_result_demo_result_N5_initial.eps}
%            \includegraphics[width=0.33\columnwidth]{Imgs_show_result_demo_result_N5_final.eps}}\\
  \subfloat{\includegraphics[width=0.33\columnwidth,height=0.6in]{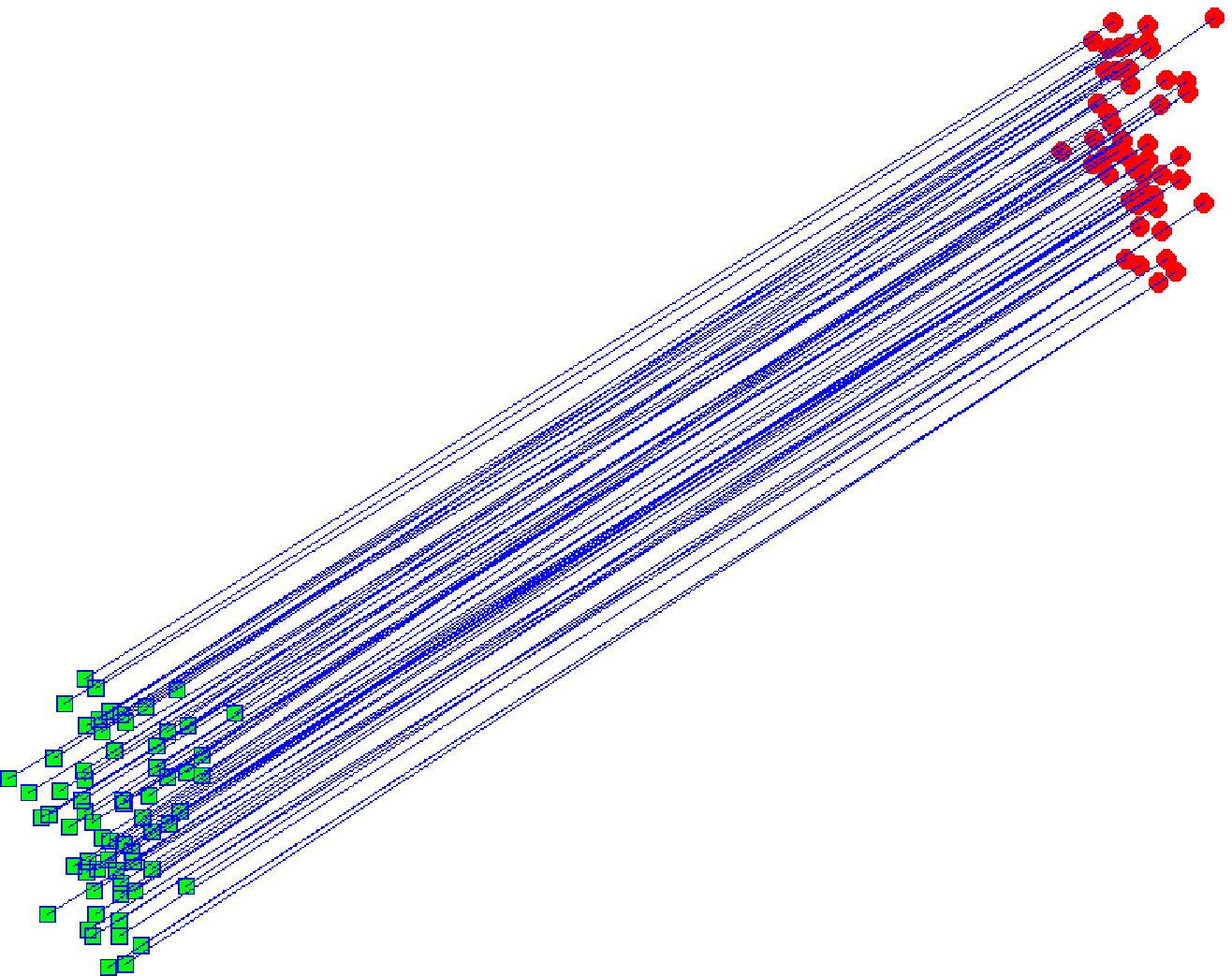}
            \includegraphics[width=0.33\columnwidth]{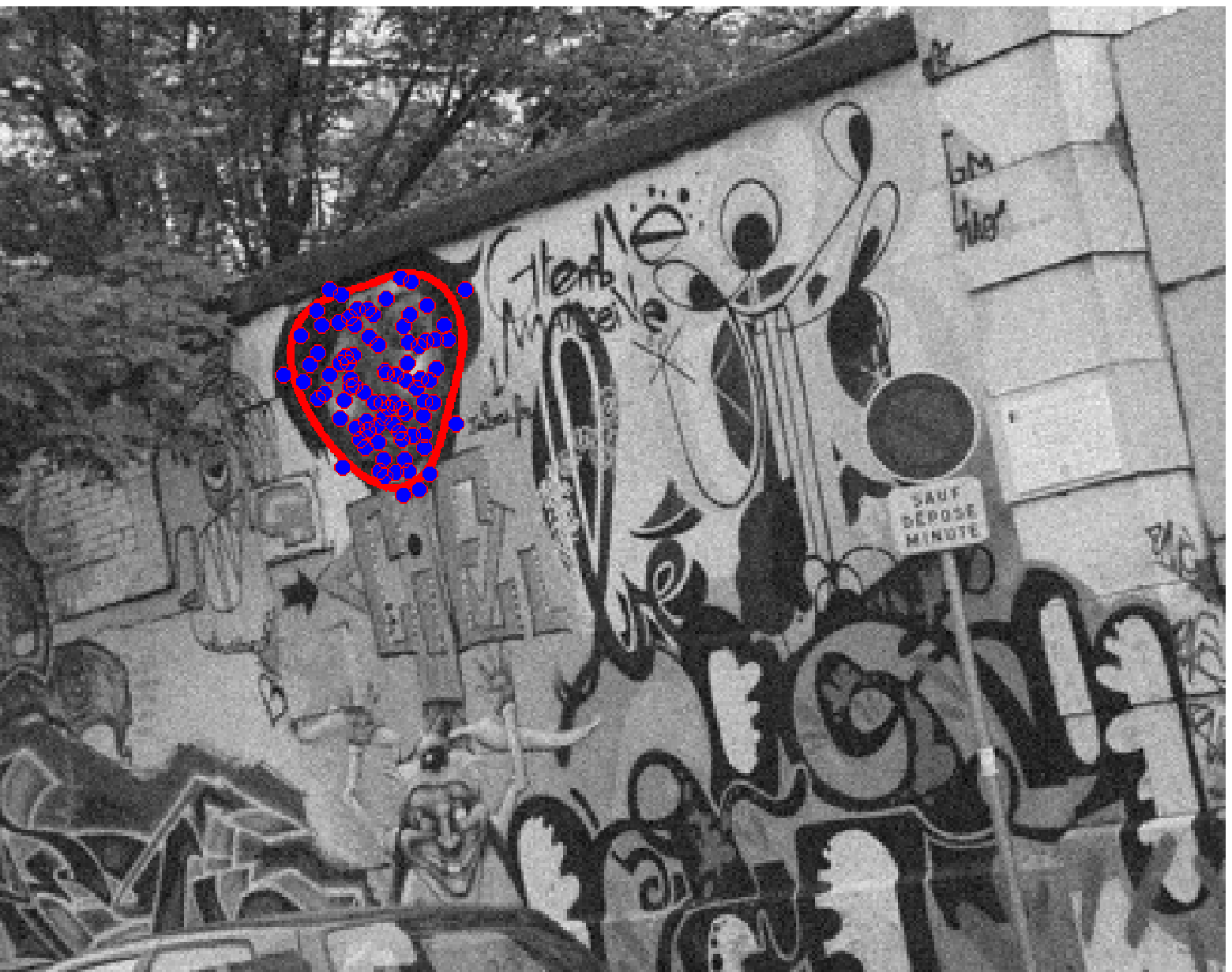}
            \includegraphics[width=0.33\columnwidth]{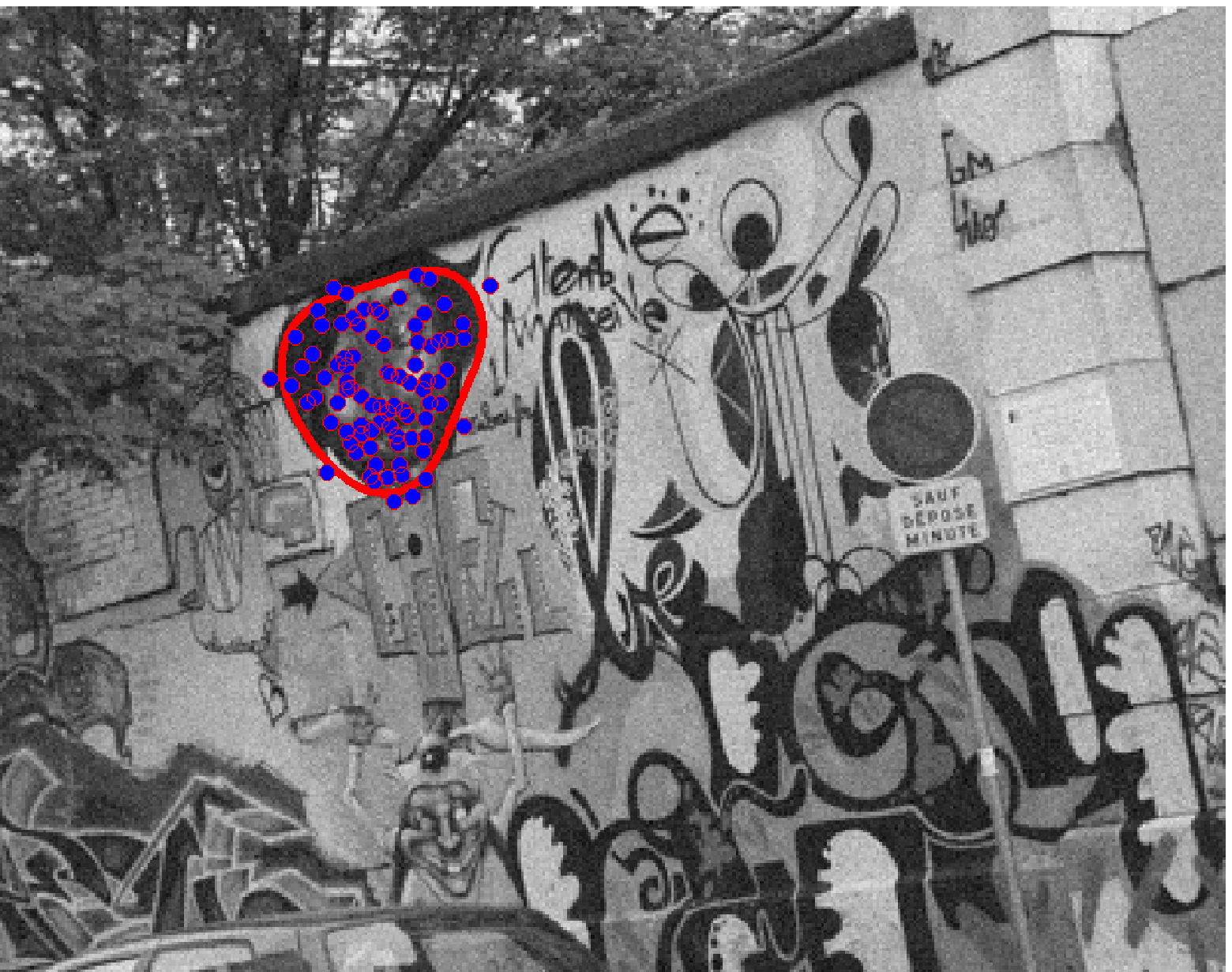}}\\
  \subfloat{\includegraphics[width=0.33\columnwidth,height=0.8in]{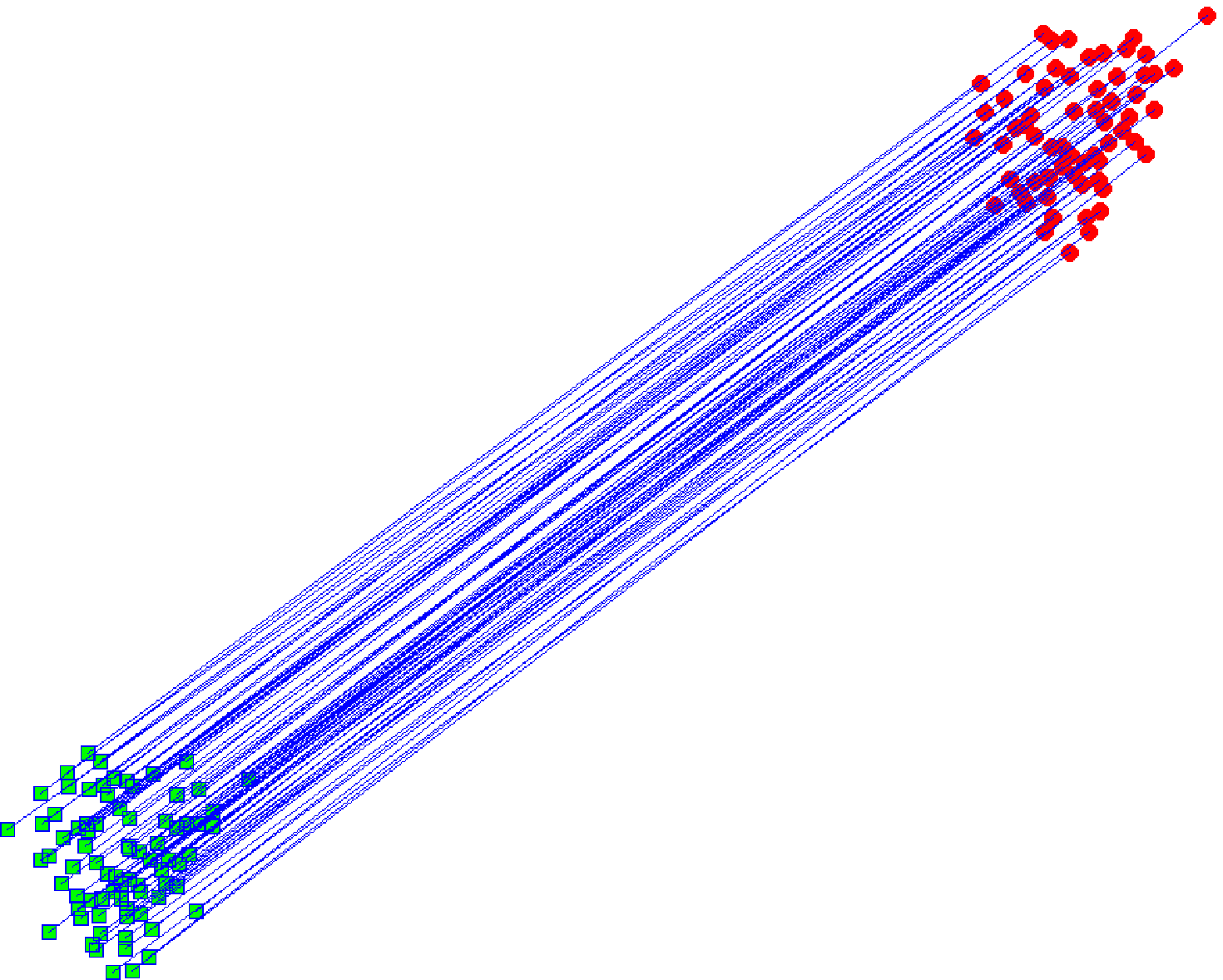}
            \includegraphics[width=0.33\columnwidth]{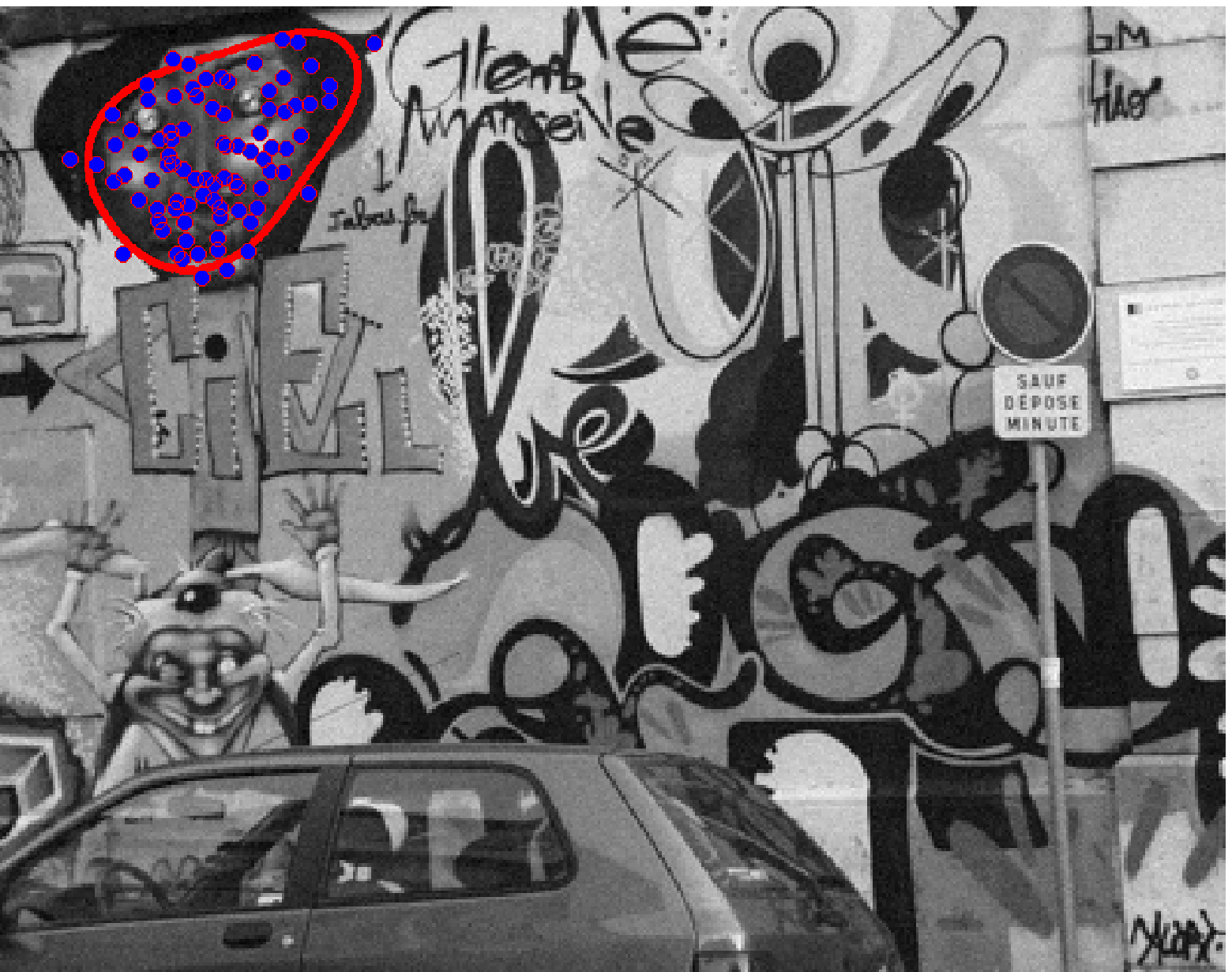}
            \includegraphics[width=0.33\columnwidth]{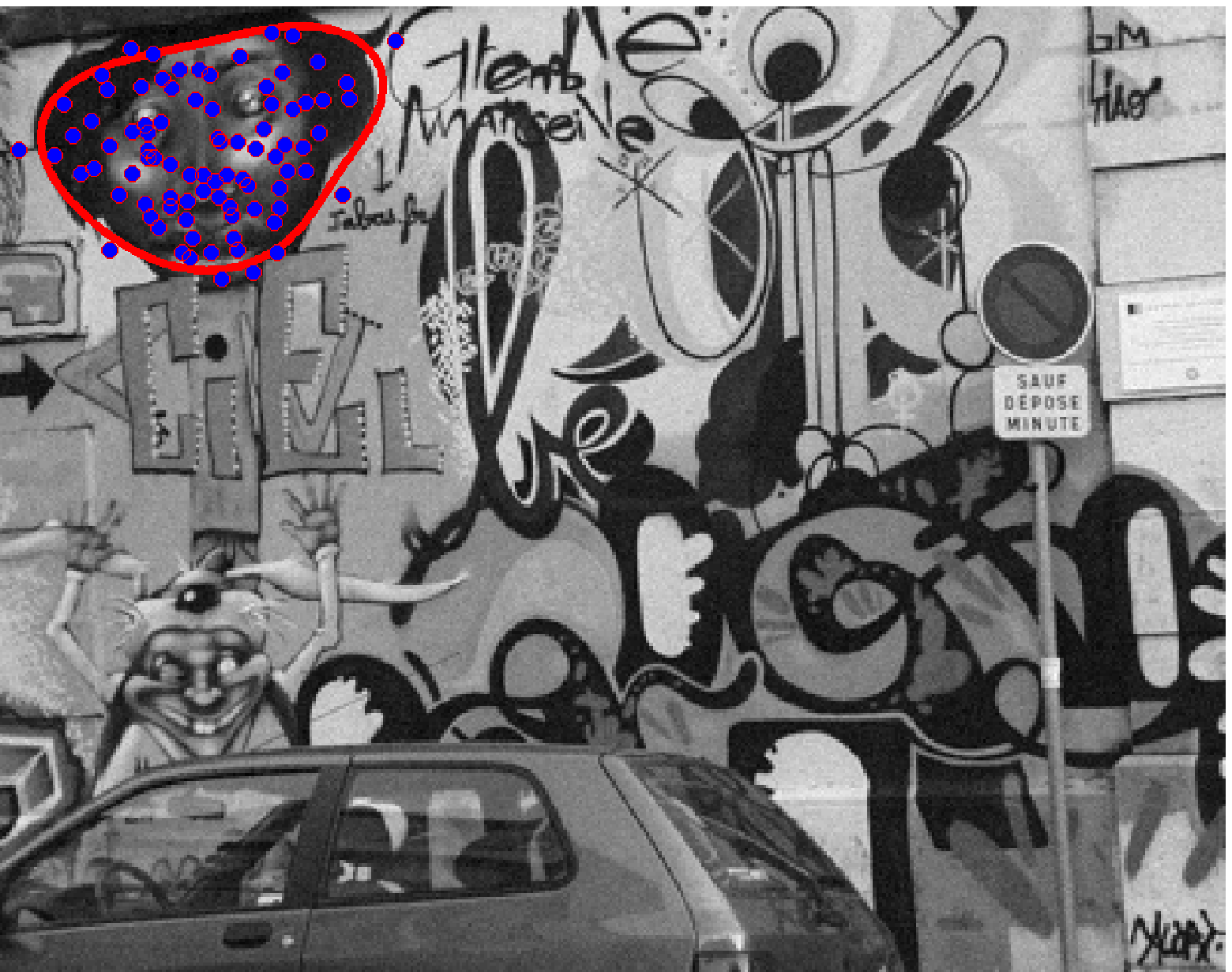}}
  \caption{Segmentation of an object of different poses in noisy images in the presence of occlusion by leaves, clutter background and indefinite boundary on the top. We label each triple of the figures from top to bottom as (a) (b) (c) and (d).}\label{FIG:Face}
\end{figure}

\begin{figure}[htb]
\centering
  \subfloat{\includegraphics[width=0.33\columnwidth,height=0.6in]{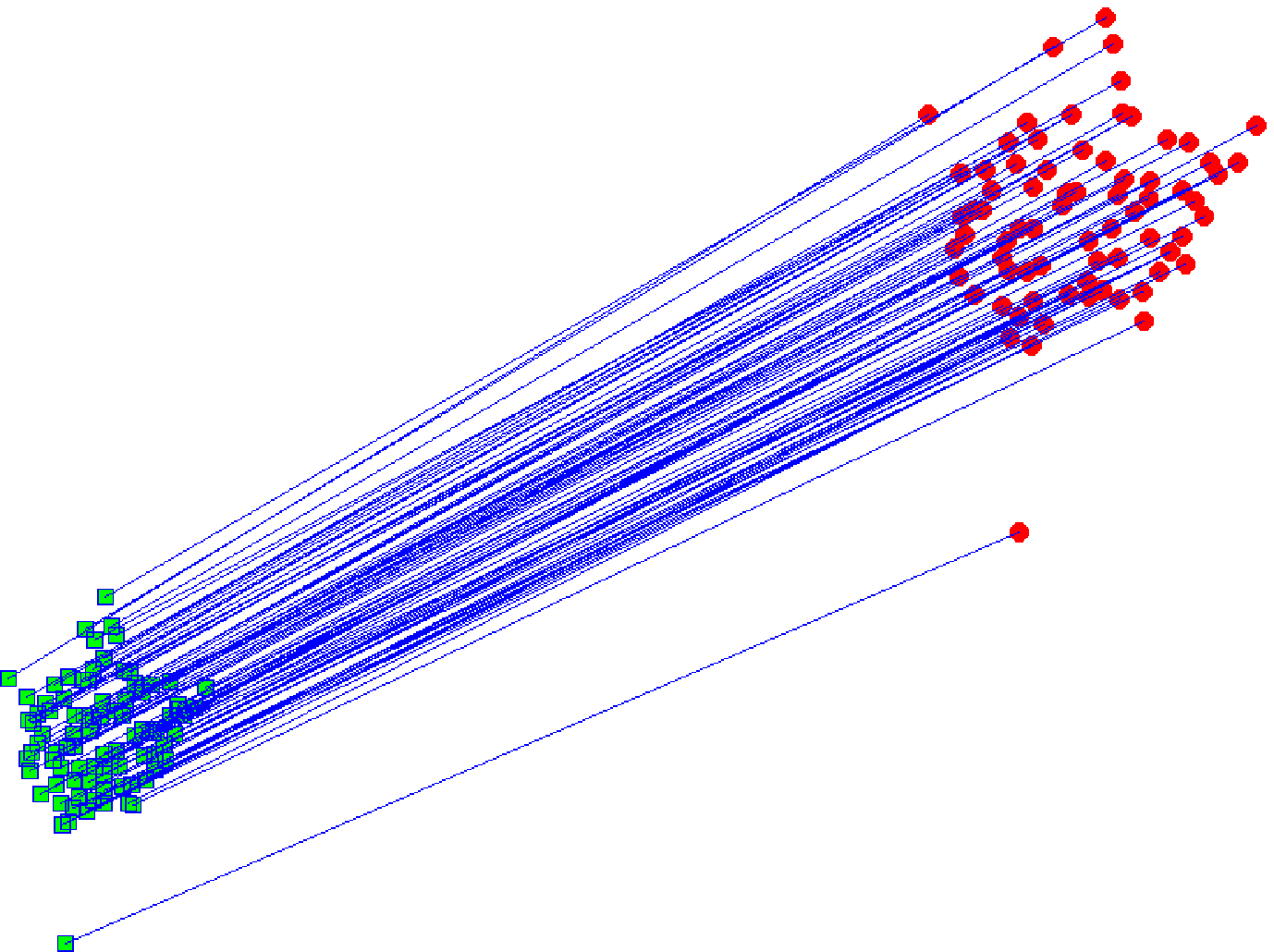}
            \includegraphics[width=0.33\columnwidth]{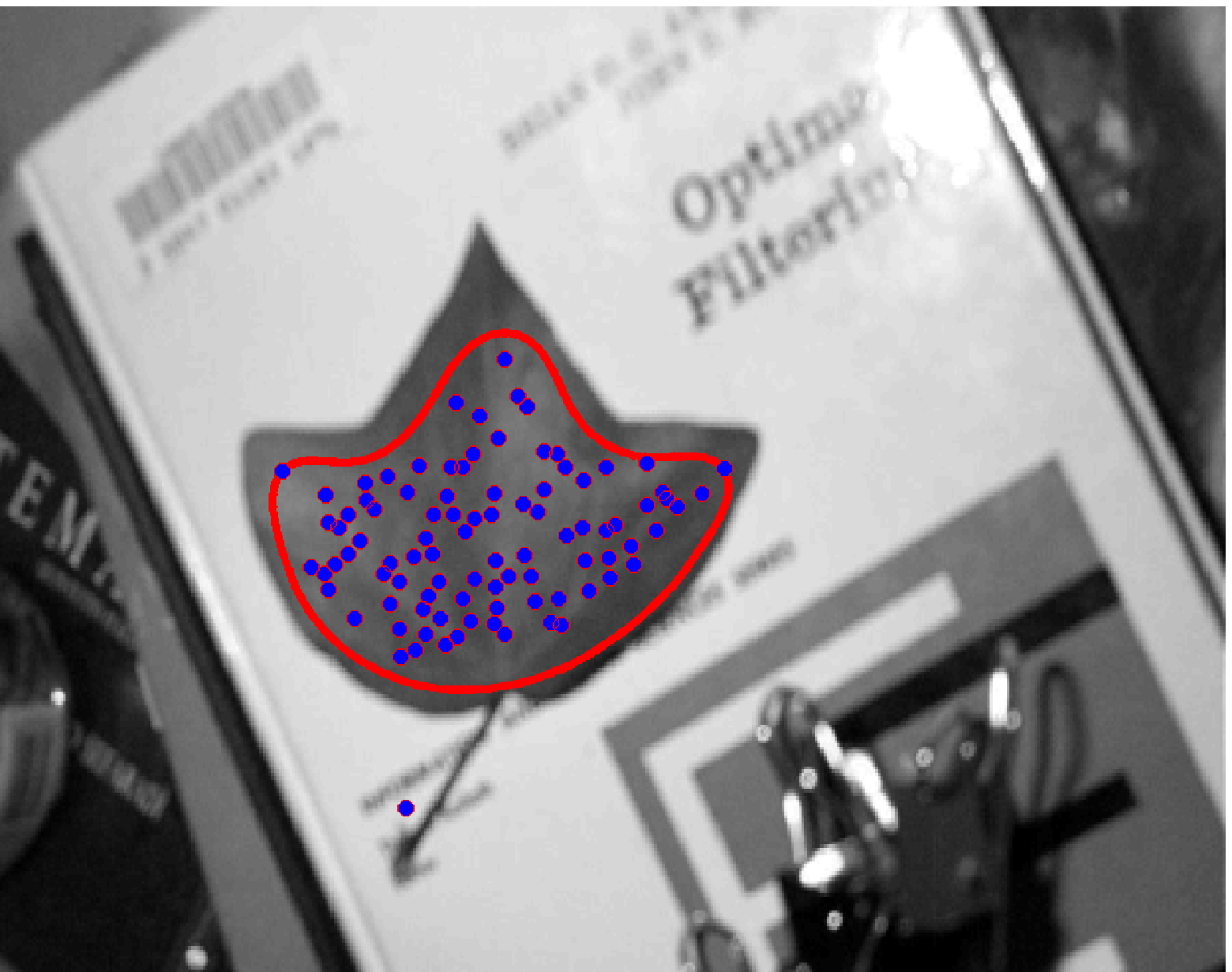}
            \includegraphics[width=0.33\columnwidth]{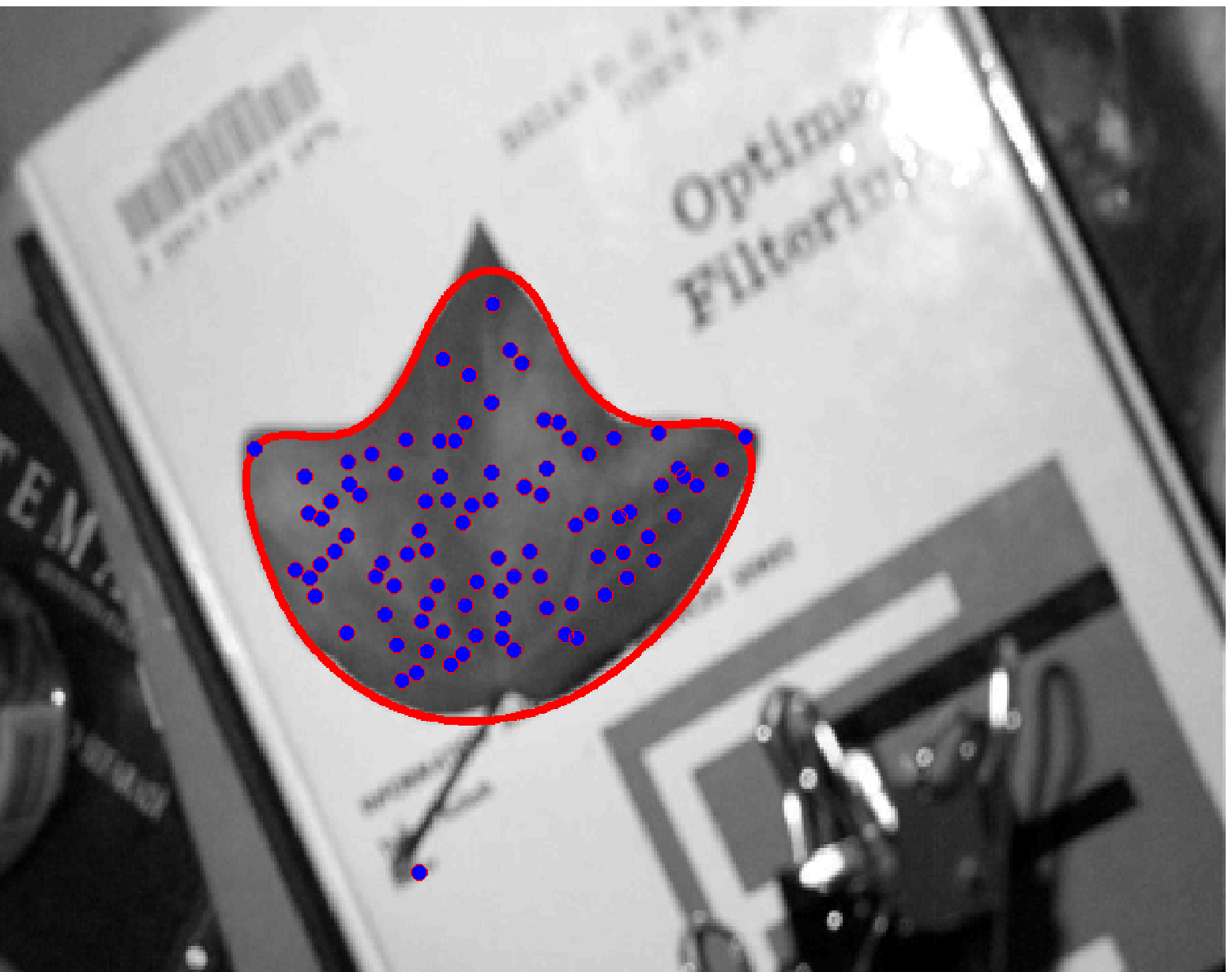}}\\
  \subfloat{\includegraphics[width=0.33\columnwidth,height=0.7in]{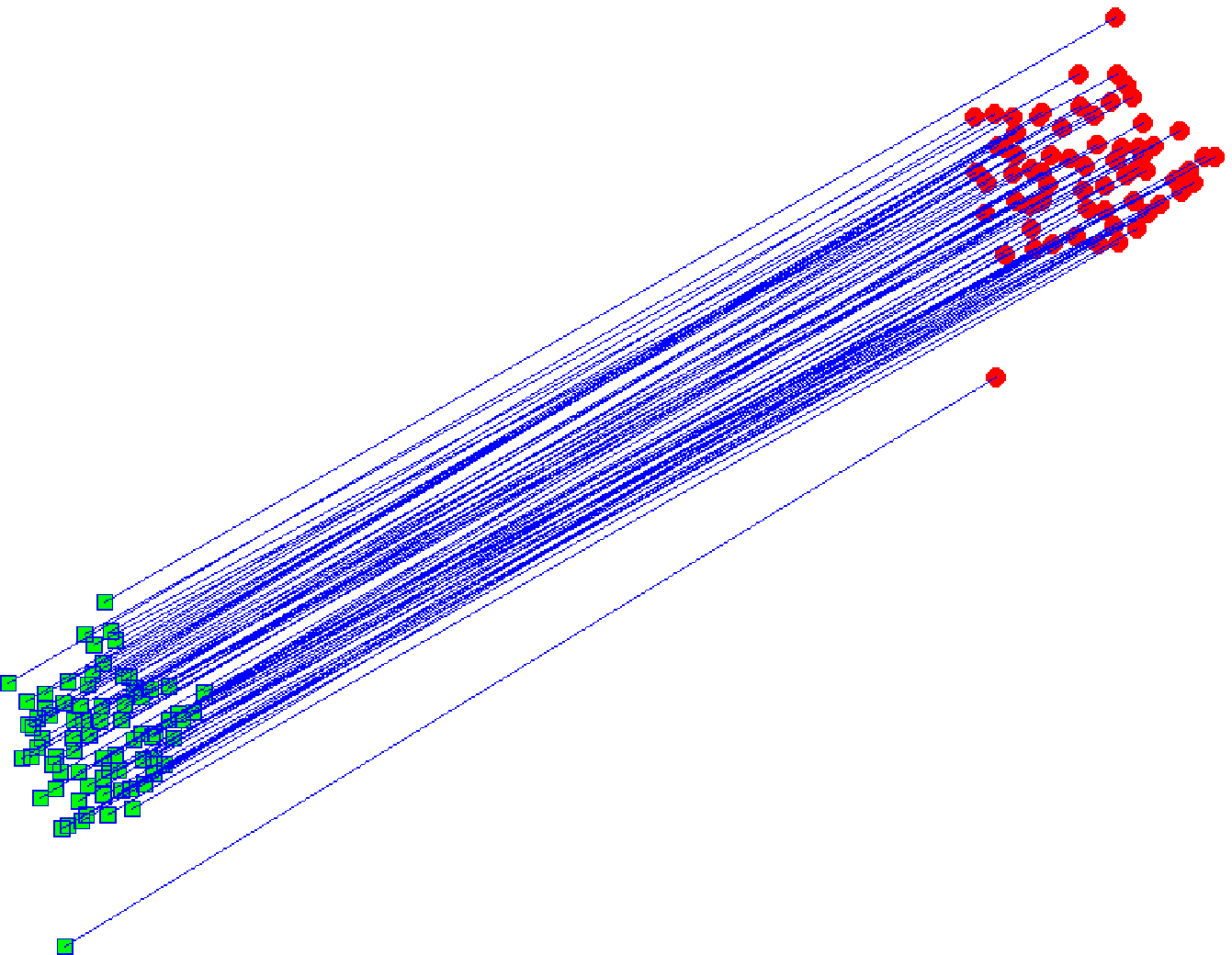}
            \includegraphics[width=0.33\columnwidth]{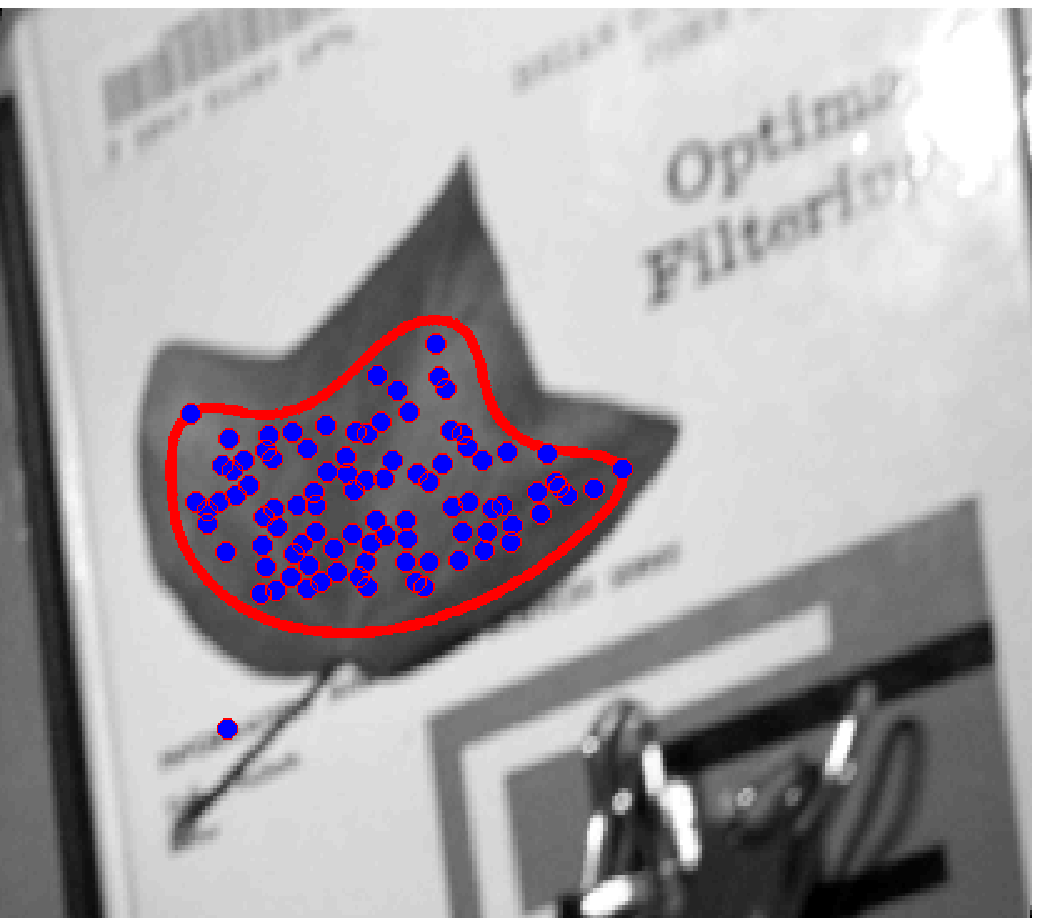}
            \includegraphics[width=0.33\columnwidth]{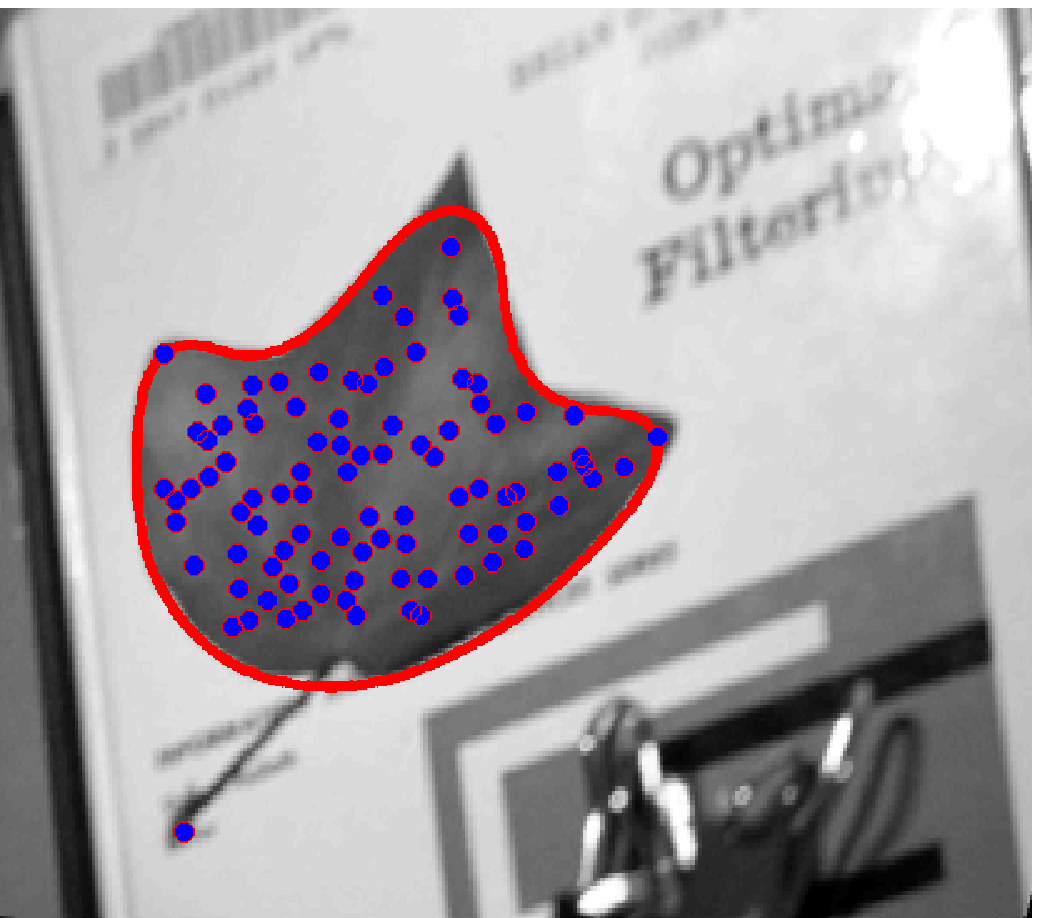}}
  \caption{Segmentation of an object of relatively complex shape under different poses and scales with non-ideal initial matchings. The top and bottom triple of figures are labeled as (a) and (b).}\label{FIG:Leaf}
\end{figure}

\begin{figure}[htb]
\centering
\subfloat{\includegraphics[width=0.33\columnwidth,height=0.5in]{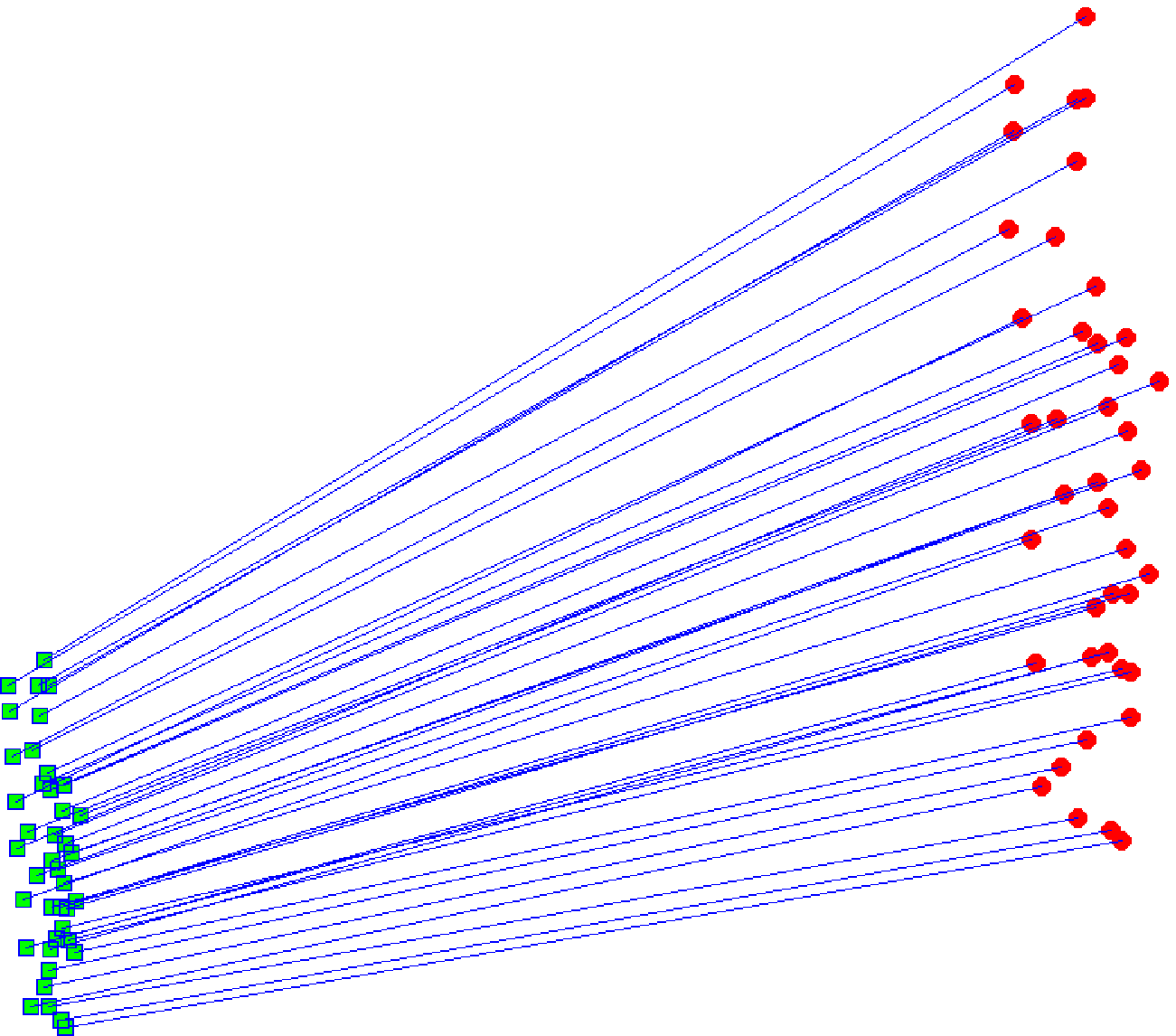}
\includegraphics[width=0.33\columnwidth]{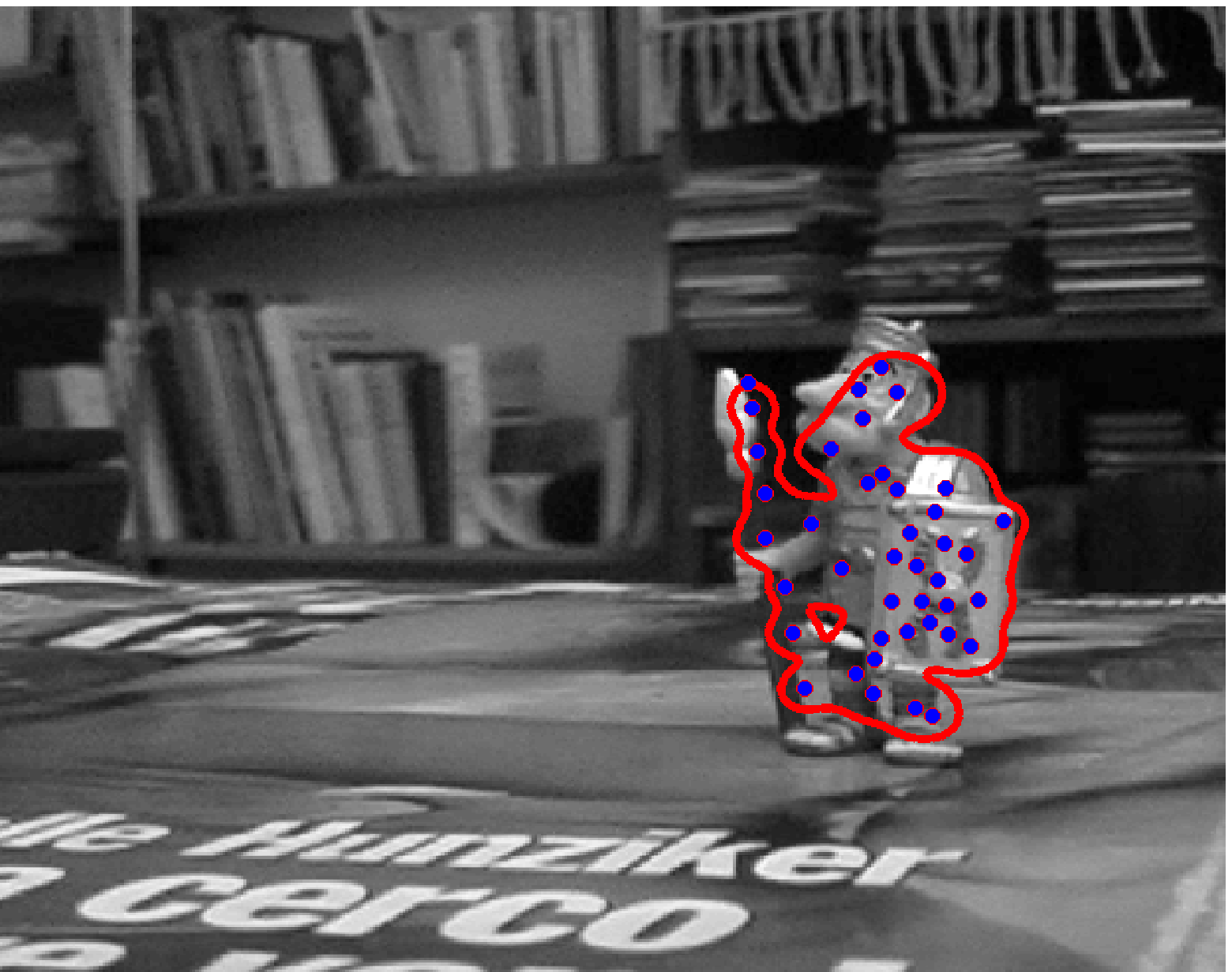}
\includegraphics[width=0.33\columnwidth]{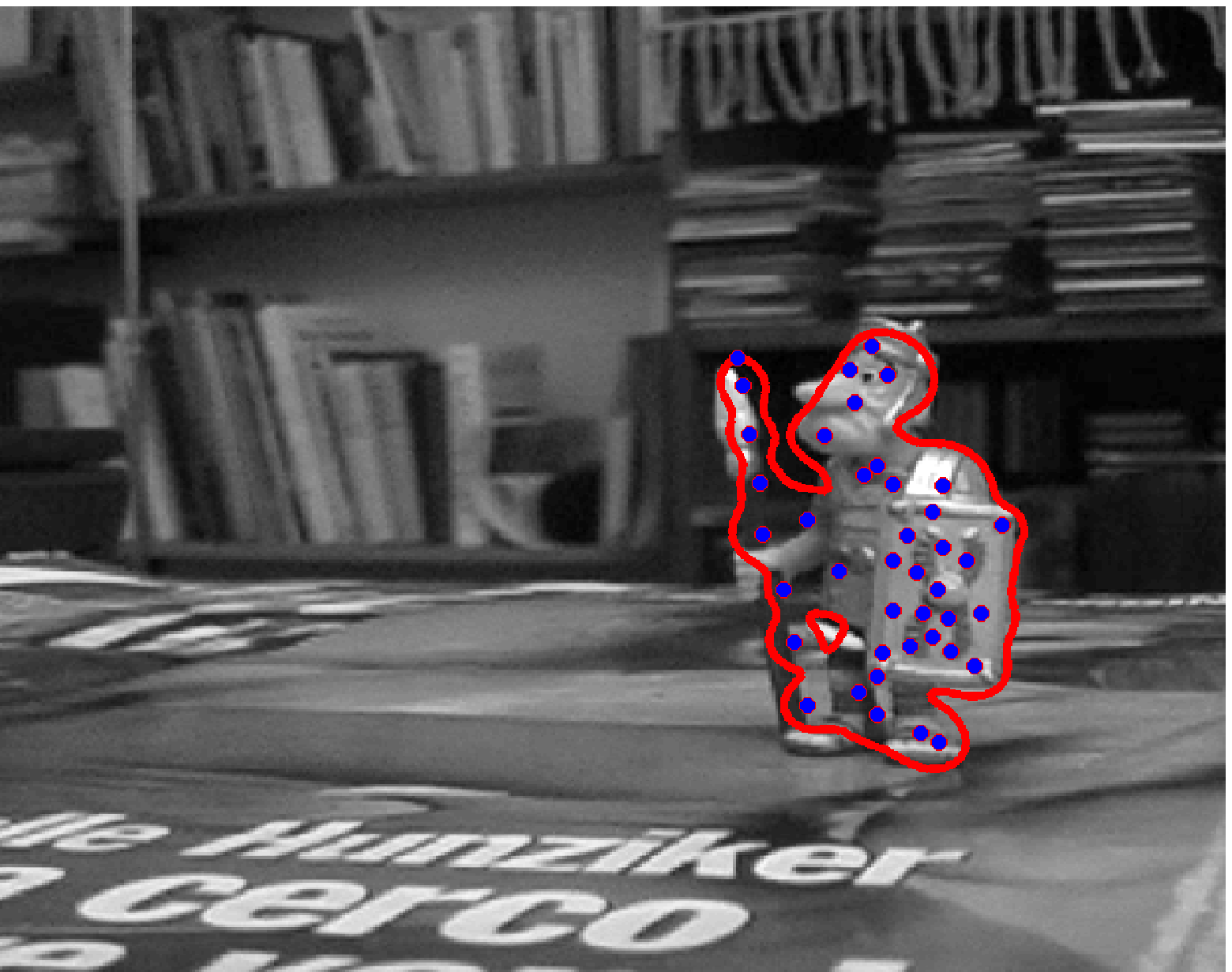}}
\caption{Segmentation of an object of complex shape under different scale, clutter background and indefinite boundary.}\label{FIG:ToyGuard}
\end{figure}

From the point correspondences (obtained from SIFT matching) as shown in the left-most columns of Figs. \ref{FIG:Face}, \ref{FIG:Leaf} and \ref{FIG:ToyGuard}, we obtain the initial contours shown in the middle columns of the figures and we finally obtain the segmentation results shown in the right-most columns of the figures. We can observe that the initial contours produced by our interior-points-to-shape relation are close to the object boundaries, and the active contour further improves the boundary location. Our method is capable of achieving the desirable segmentation results. We also compare the initial and final segmentation results with manually labeled ground truth shapes by Jaccard similarity measure to validate our visual observations quantitatively. We summarize the results in Table \ref{TAB:JC}. We can see the improvement of the initial contour and the region enclosed by the final contour overlaps well with the ground truth. We also present the computational cost of the entire framework on the images in Table \ref{TAB:CC} for reference (Note that the entire process is run in MATLAB on a PC with Intel\circledR Core$^{\textrm{TM}}$ i5-450M Processor and 4GB memory).
\begin{table}[htb]
\centering
{\caption{Jaccard similarity against the ground truth.}\label{TAB:JC}
\begin{tabular}{@{\extracolsep{\fill}}l|cccc|cc|c}
\toprule
              &\multicolumn{4}{c|}{Fig. \ref{FIG:Face}}& \multicolumn{2}{c|}{Fig. \ref{FIG:Leaf}}& Fig. \ref{FIG:ToyGuard}\\
              & (a) & (b) & (c) & (d) & (a) & (b) & \\ \hline
Initial       & 0.73  & 0.75  & 0.74 & 0.63 & 0.53 & 0.68  & 0.65 \\
Final   & 0.75 & 0.84 & 0.86 & 0.86 & 0.90 & 0.93 & 0.75 \\
\bottomrule
\end{tabular}}
\end{table}
\begin{table*}[htb]
{\footnotesize\caption{Computational costs (seconds) }\label{TAB:CC}
\begin{tabular*}{\textwidth}{@{\extracolsep{\fill}}lcccc|cc|c}
\toprule

Figures            & Fig. \ref{FIG:Face}(a) & Fig. \ref{FIG:Face}(b) & Fig. \ref{FIG:Face}(c) & Fig. \ref{FIG:Face}(d) & Fig. \ref{FIG:Leaf}(a) & Fig. \ref{FIG:Leaf}(b) & Fig. \ref{FIG:ToyGuard}\\ \hline
Size(pixel)       & 320$\times$400 & 320$\times$400 & 320$\times$400 & 320$\times$400 & 296$\times$448 & 282$\times$448 & 411$\times$408\\
M\&R       & 17.83  & 18.3  & 18.53 & 17.9 & 18.44 & 16.27  & 10.05 \\
GC & 4.26  & 4.21  & 4.20 & 4.32 & 7.14 & 7.10  & 2.29 \\
AC   & 98.14 & 40.93 & 82.03 & 650.56 & 504.97 & 401.95 & 82.95 \\
Total & 120.23 &63.46 &104.75&672.78&530.56&425.32&95.29\\
\bottomrule
\end{tabular*}}
\scriptsize{M\&R=matching and registration, GC= generation of initial contour, AC = Active contour before convergence, Total = Total running time}
\end{table*}

\begin{figure}
\centering
  \includegraphics[width=0.45\textwidth]{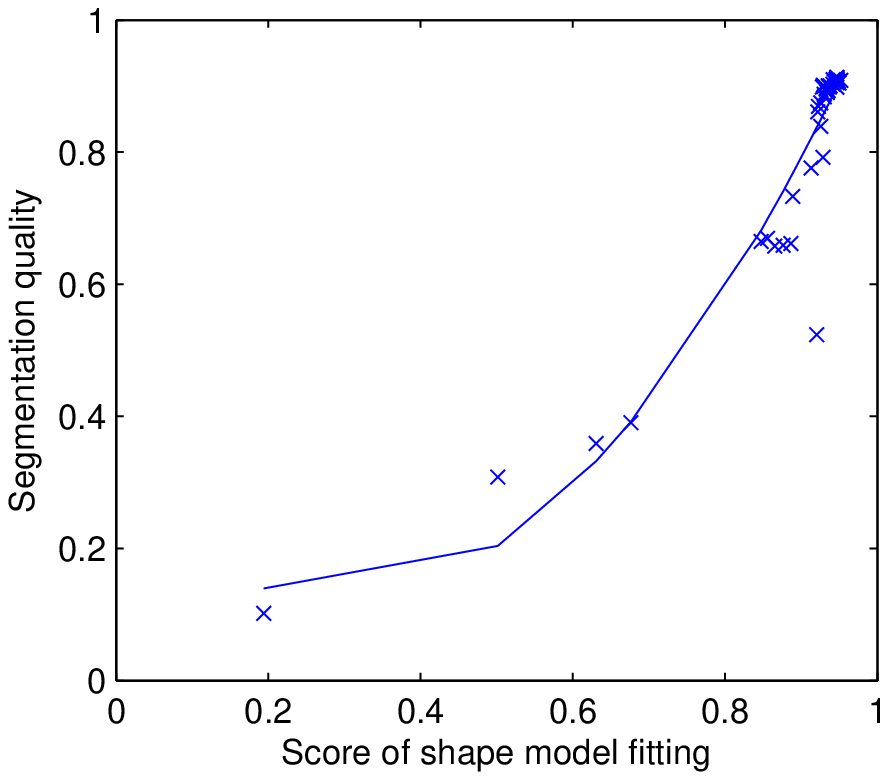}\\
  \caption{Quality of segmentation (vertical) v.s. score of model fitting (horizontal)}\label{FIG:model_vs_seg}
\end{figure}

We have implicitly assumed that a better shape model leads to a better result of segmentation throughout the paper. We ascertain this by experiments. We use the $50$ shape models due to the $50$ random $\sigma$ generated previously for extraction of the leaf in the top image in Fig \ref{FIG:Leaf}. In Fig. \ref{FIG:model_vs_seg}, we show the strong correlation between the quality of shape modeling, in terms of fitting score, and the quality of segmentation, in terms of Jaccard similarity between the result and the ground truth.

\subsection{Robustness to noise}
\begin{figure}
\centering
  \subfloat[]{\includegraphics[width=0.9\columnwidth]{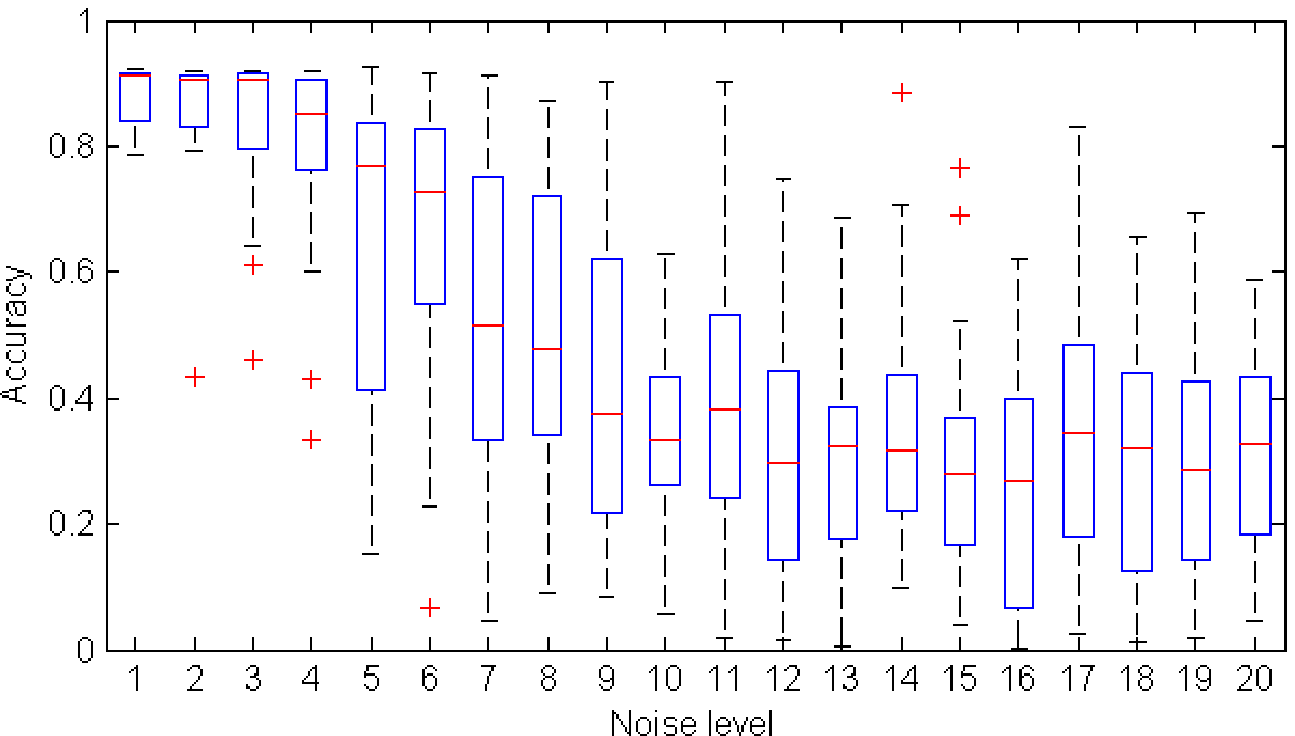}}\\
  \subfloat[]{\includegraphics[width=0.45\columnwidth]{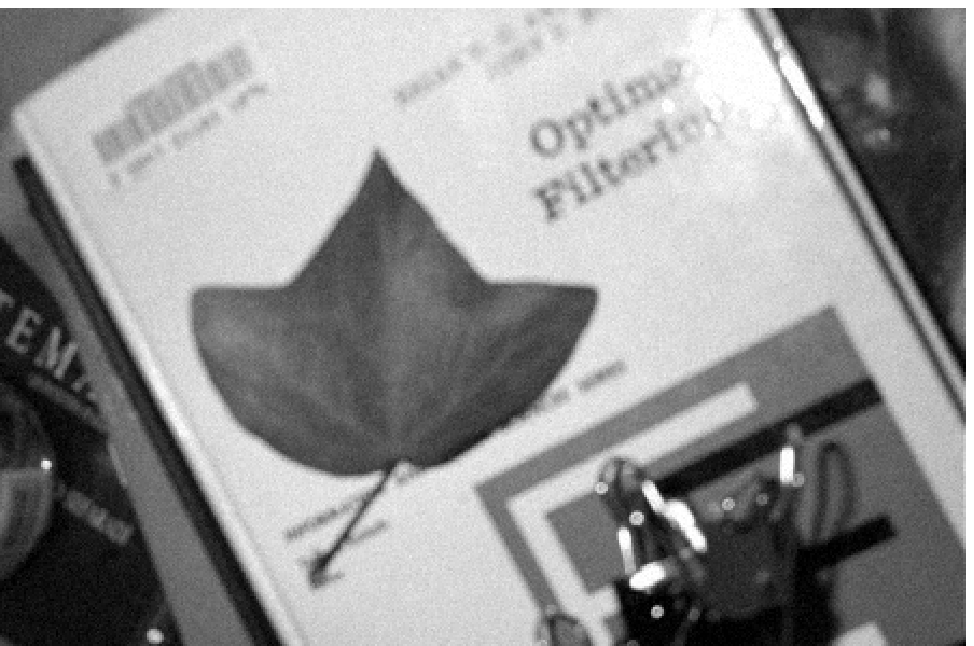}}
  \subfloat[]{\includegraphics[width=0.45\columnwidth]{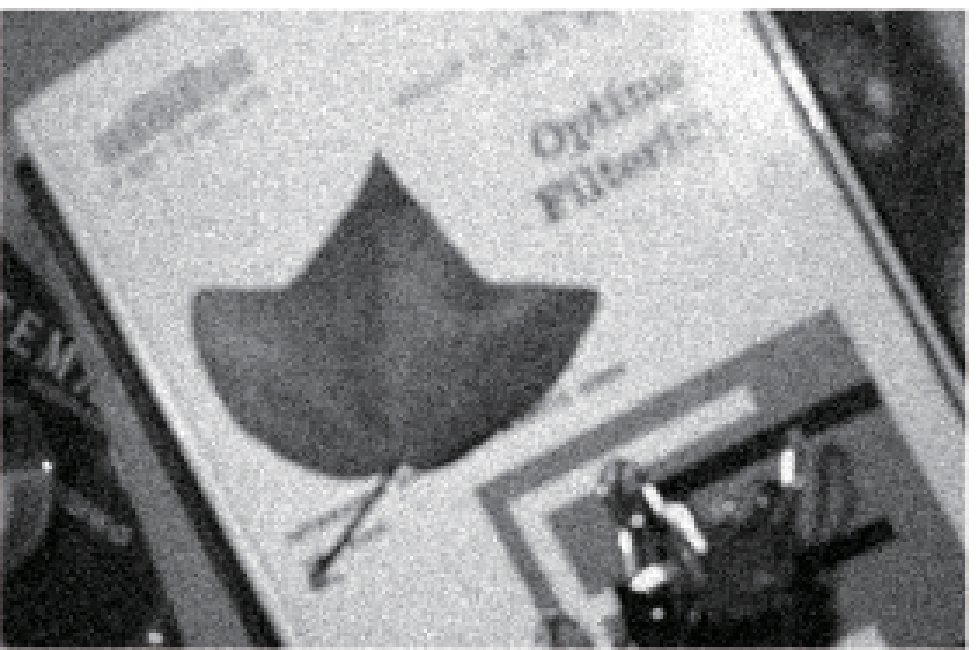}}
  \caption{Segmentation under different noise level. (a) is the boxplot of the accuracy v.s. noise level. (b) and (c) are the images with noise. The standard deviation of the noise in (b) is 5, and the other is 20. }\label{FIG:test_noise}
\end{figure}

We also evaluate our method under different noise level. We use the image in Fig. \ref{FIG:Leaf}(a). We add Gaussian noise to the image. The means of the Gaussian noise are set to be zero and the standard deviations are varied from 1 to 20. For each noise level, we create 30 images and we apply our method to the images. The boxplot of the results of segmentation are shown in Fig. \ref{FIG:test_noise}. We can observe that when the standard deviation of the noise is below $5$, our method is robust. However, the decay of accuracy is sharp when the standard deviation is larger than $5$. Hence, the proposed method may be sensitive to noise. This is because we adopt the SIFT feature for matching. This problem of sensitivity to noise in object matching have been addressed by robust feature representations such as the PCA-SIFT \cite{Ke04PCASIFT}.

\subsection{Experimental comparison}
In this subsection, we conduct experiments for comparing our method with other related methods, including global optimization of the GAC model (global search, or GS), co-segmentation based on discriminative clustering (DCCoSeg) \cite{JouBacPon12MultiClassCoSeg} and co-segmentation based on submodular optimization (SOCoSeg) \cite{GunheeKimEricXingLiFeiFeiKanade11SOCoSeg}. We use eight orientations and 3 scales in the global search. The implementations of the co-segmentation algorithms are taken from the authors' websites. All the methods in the comparison require an object example. The task is also the same, namely to outline the same or similar object in the image of interest.

\begin{figure}
\centering
  \includegraphics[width=0.45\textwidth]{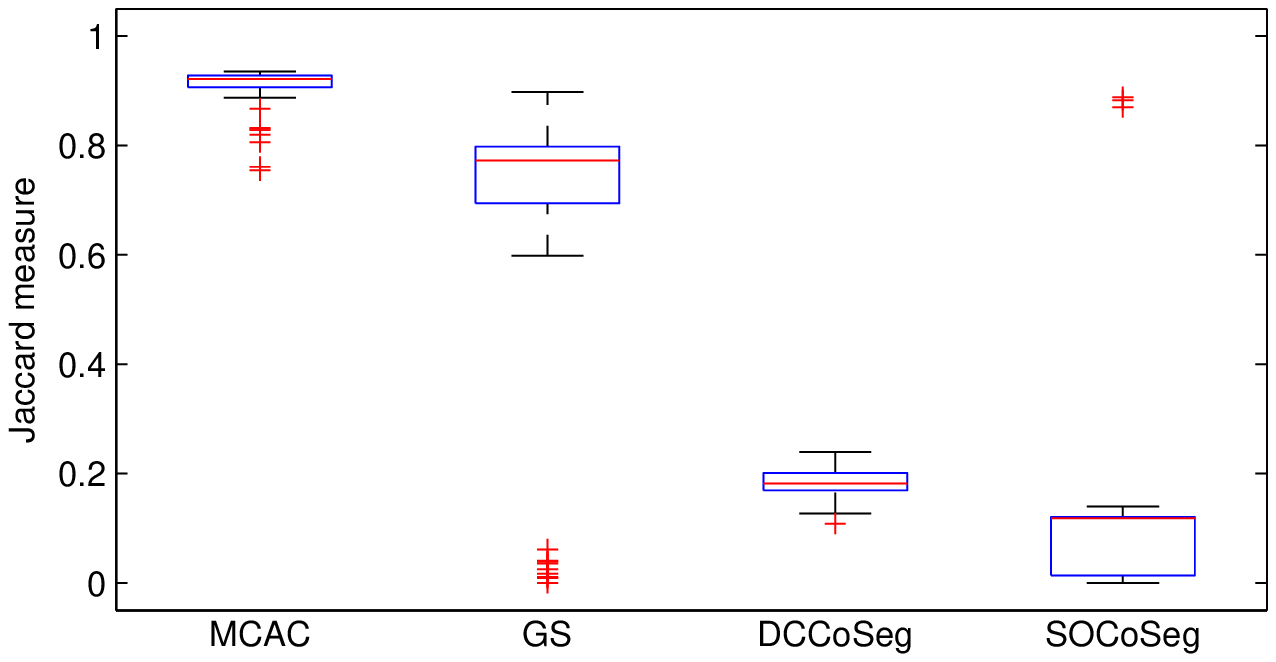}\\
  \caption{Quantitative comparison}\label{FIG:quant_cmp}
\end{figure}

\begin{figure*}
\centering
\begin{tabular}{ccccc}
  \includegraphics[height=0.8in]{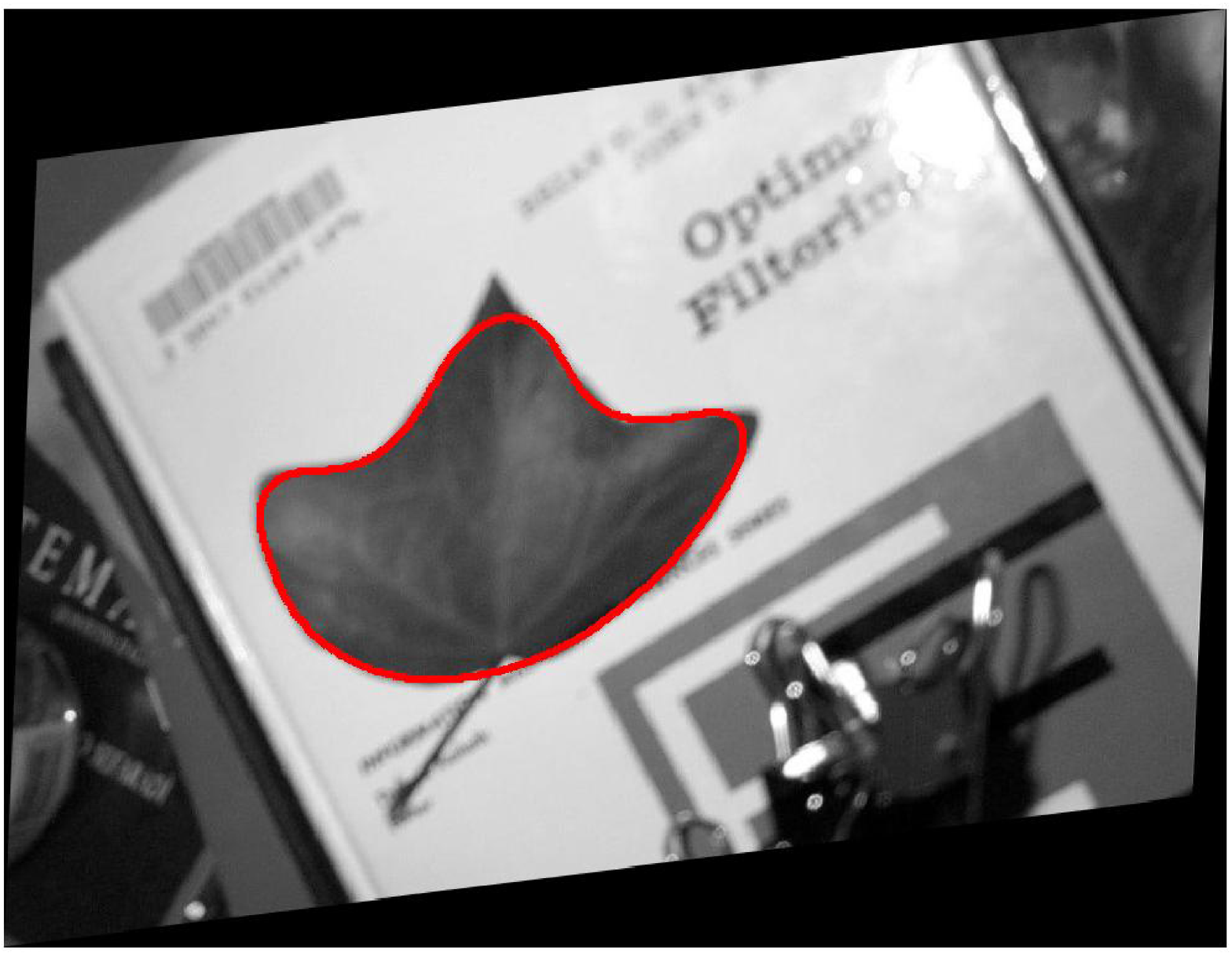}&\hspace{-10pt}
  \includegraphics[height=0.8in]{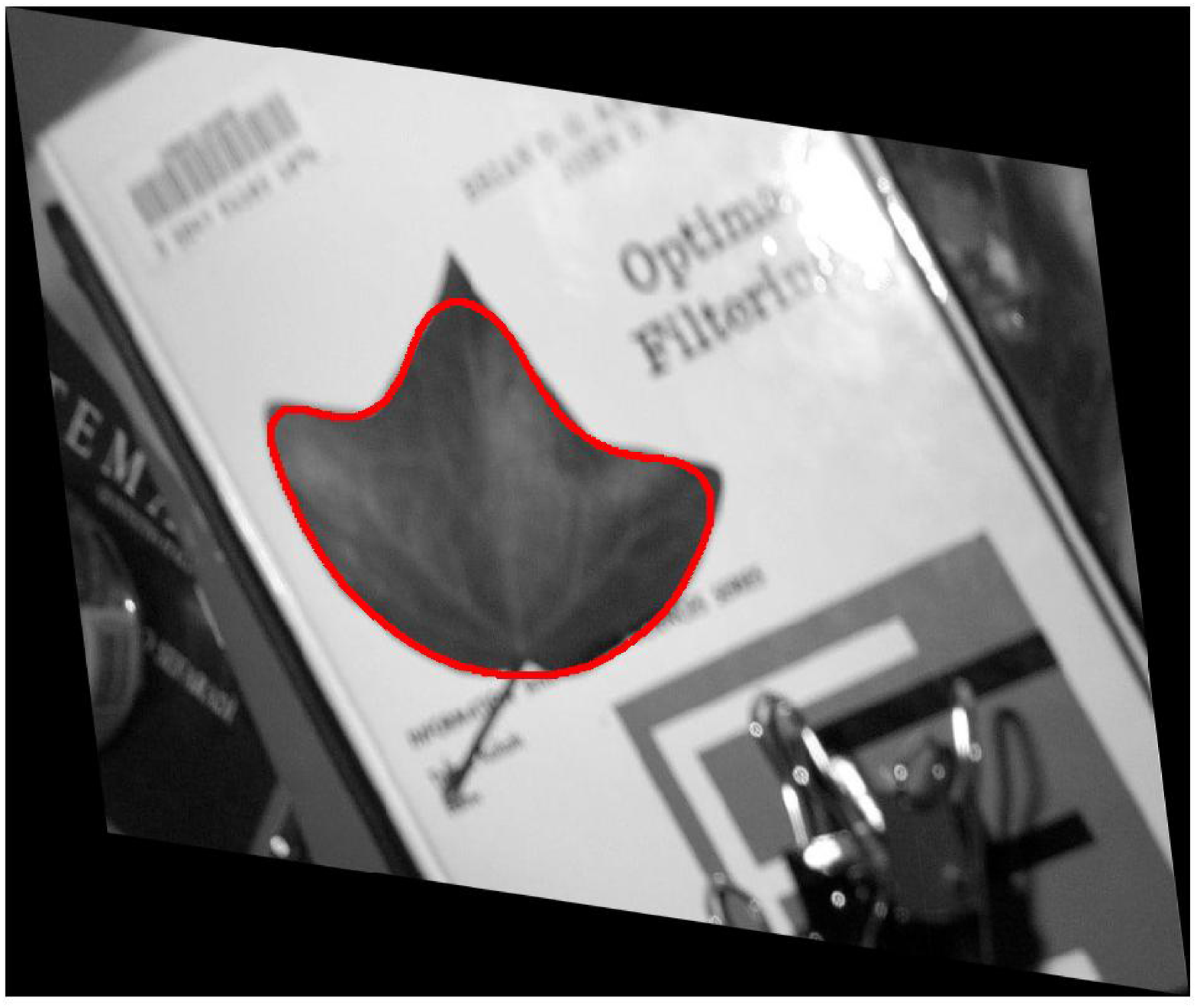}&\hspace{-10pt}
  \includegraphics[height=0.8in]{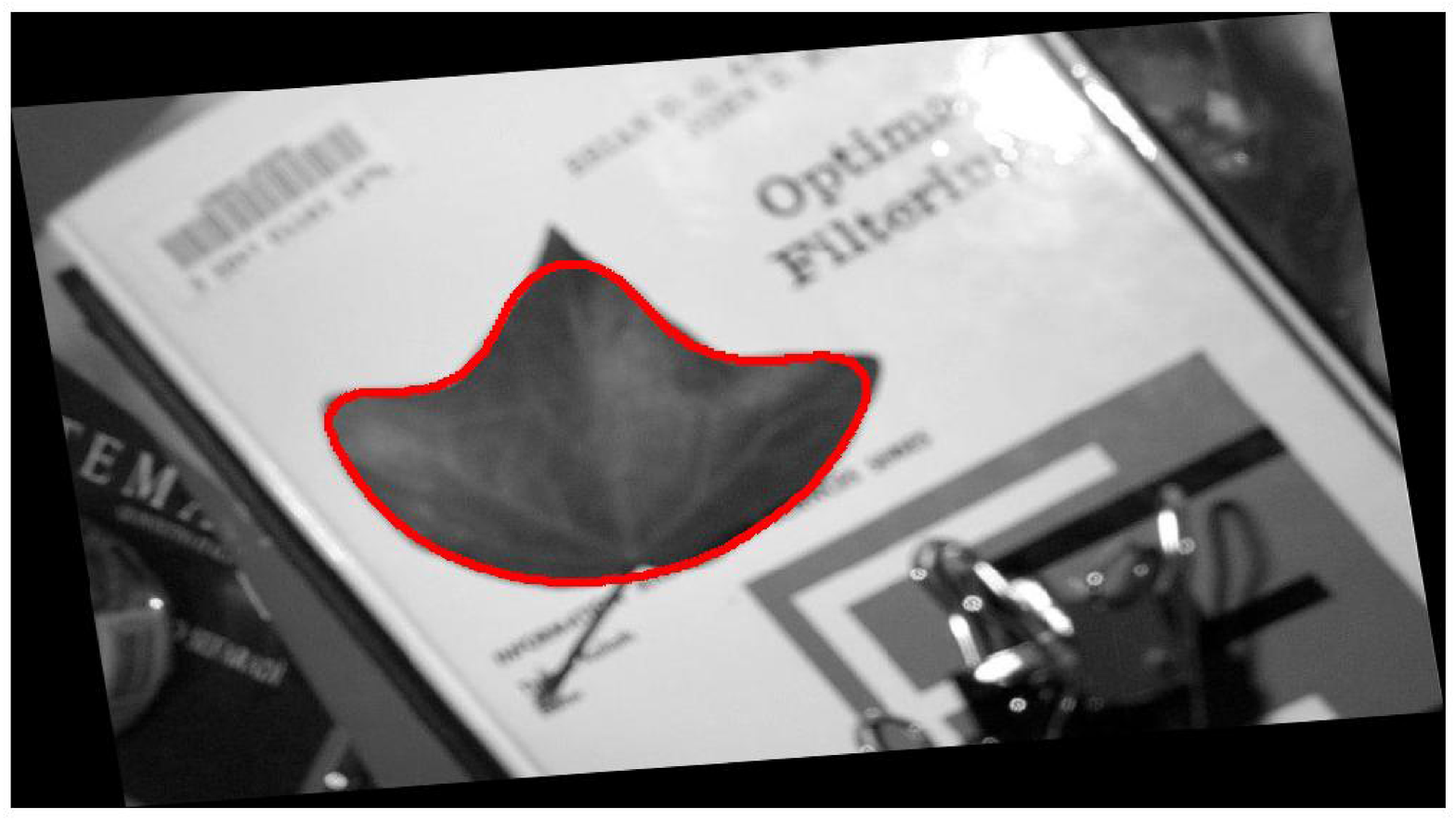}&\hspace{-10pt}
  \includegraphics[height=0.8in]{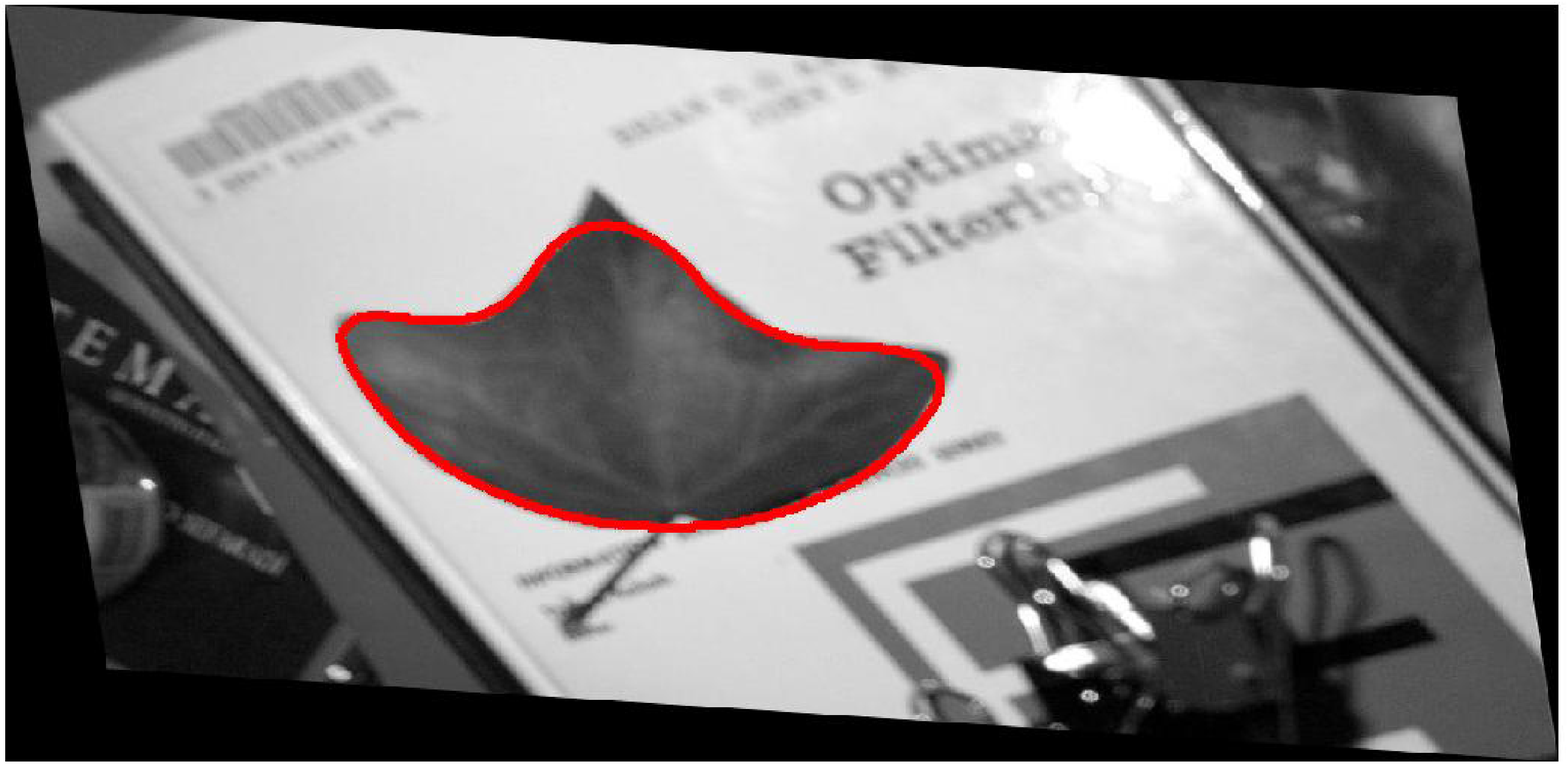}&\hspace{-10pt}
  \includegraphics[height=0.8in]{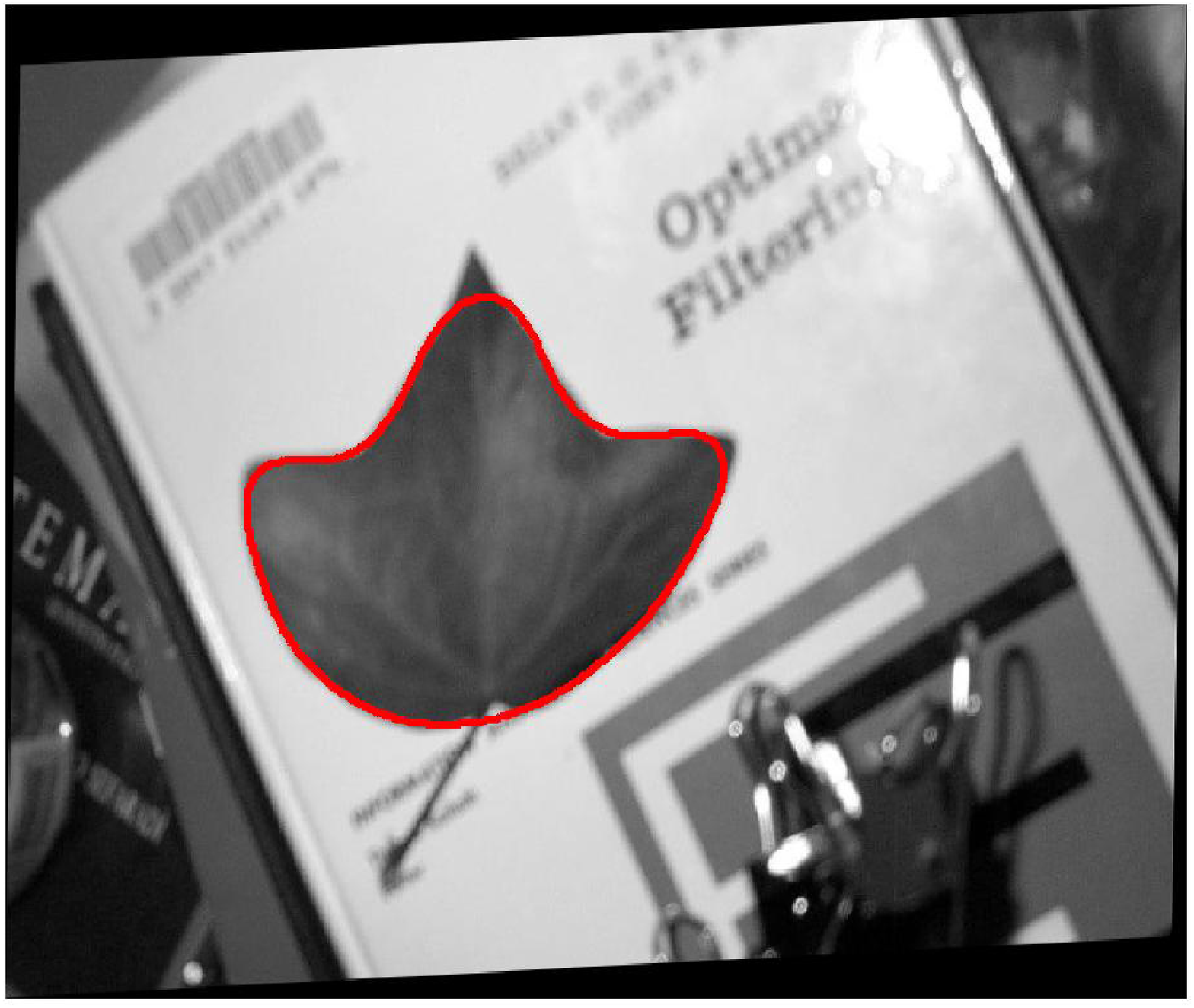}\\ \vspace{-10pt}
  \includegraphics[height=0.8in]{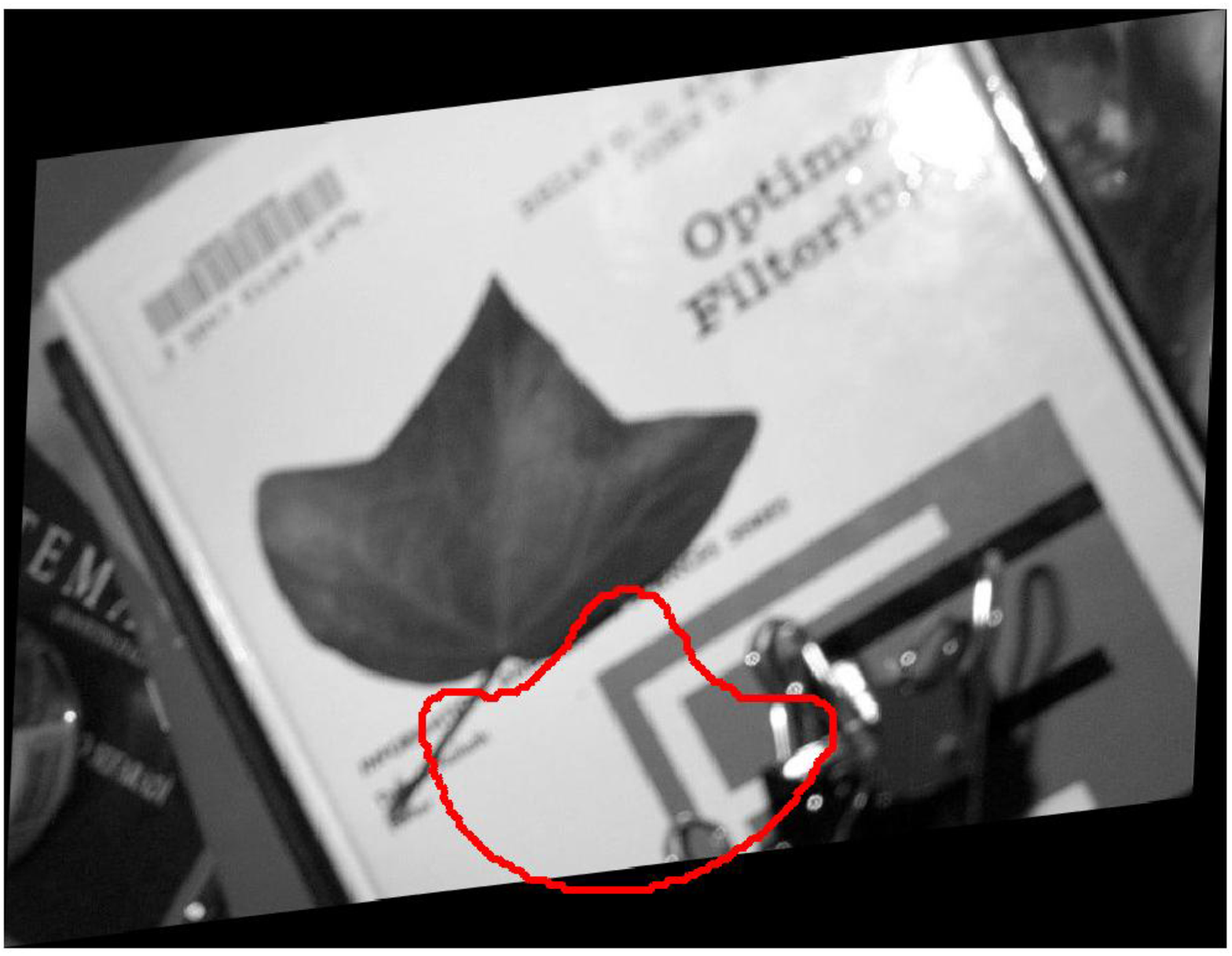}&\hspace{-10pt}
  \includegraphics[height=0.8in]{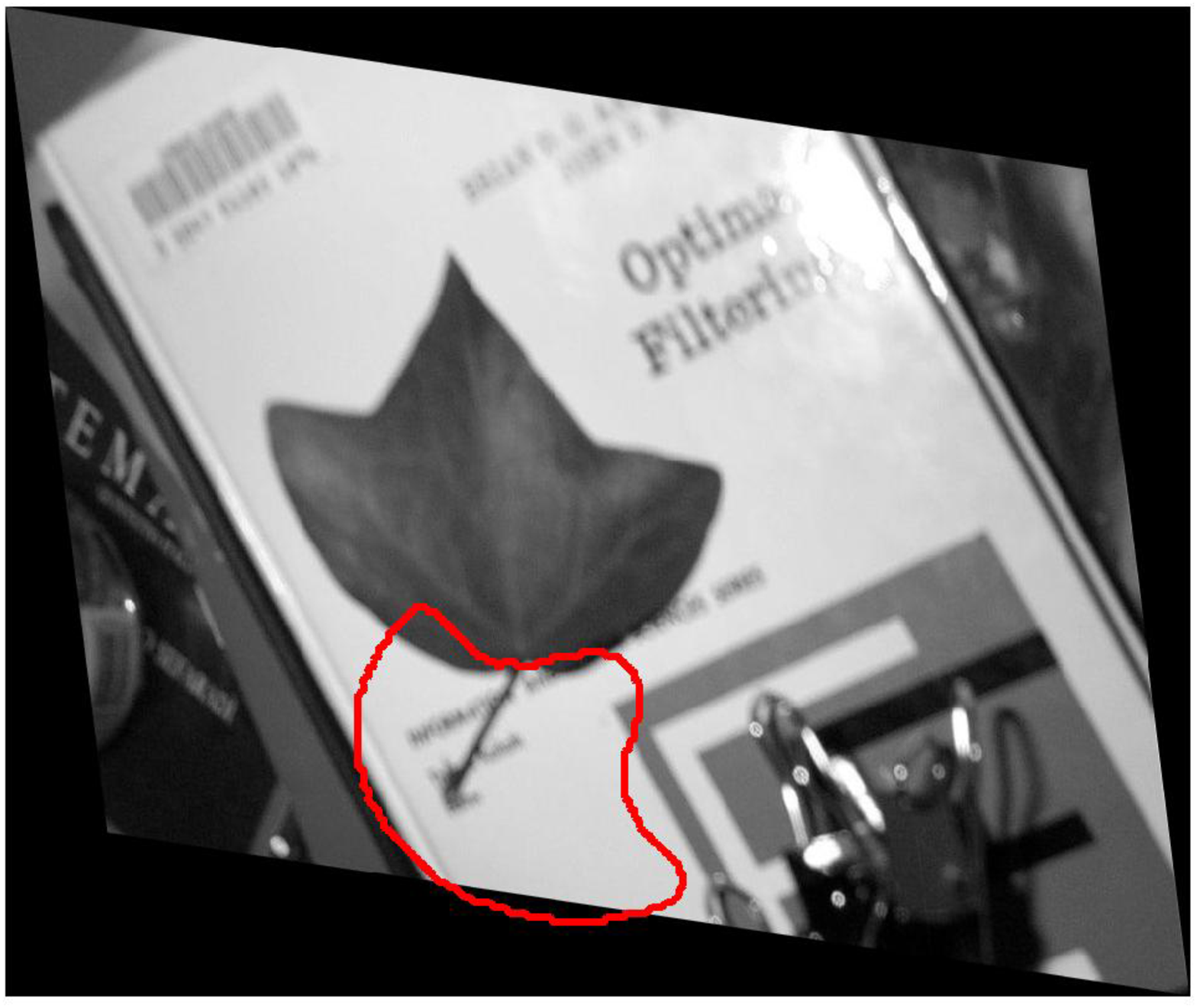}&\hspace{-10pt}
  \includegraphics[height=0.8in]{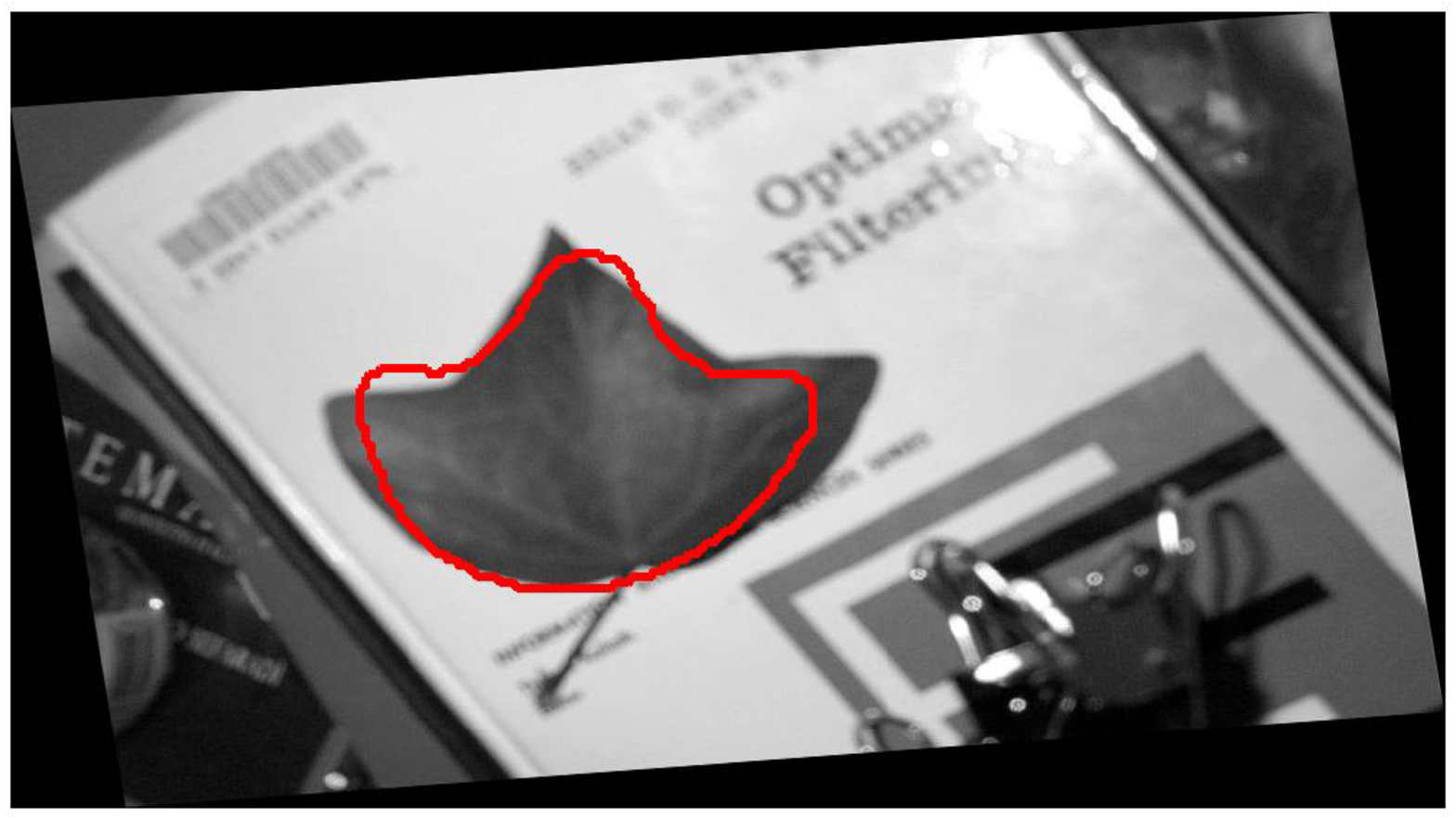}&\hspace{-10pt}
  \includegraphics[height=0.8in]{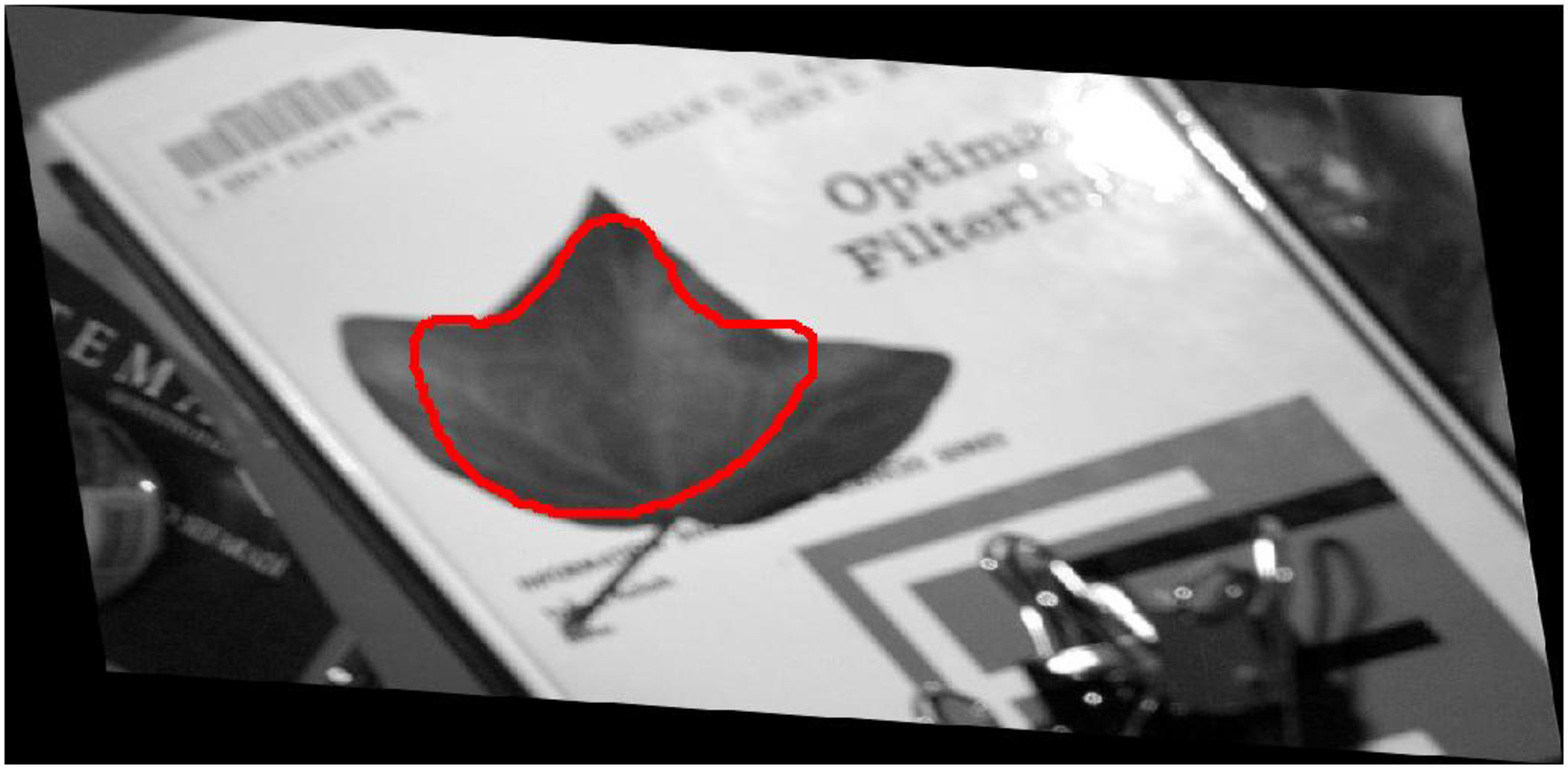}&\hspace{-10pt}
  \includegraphics[height=0.8in]{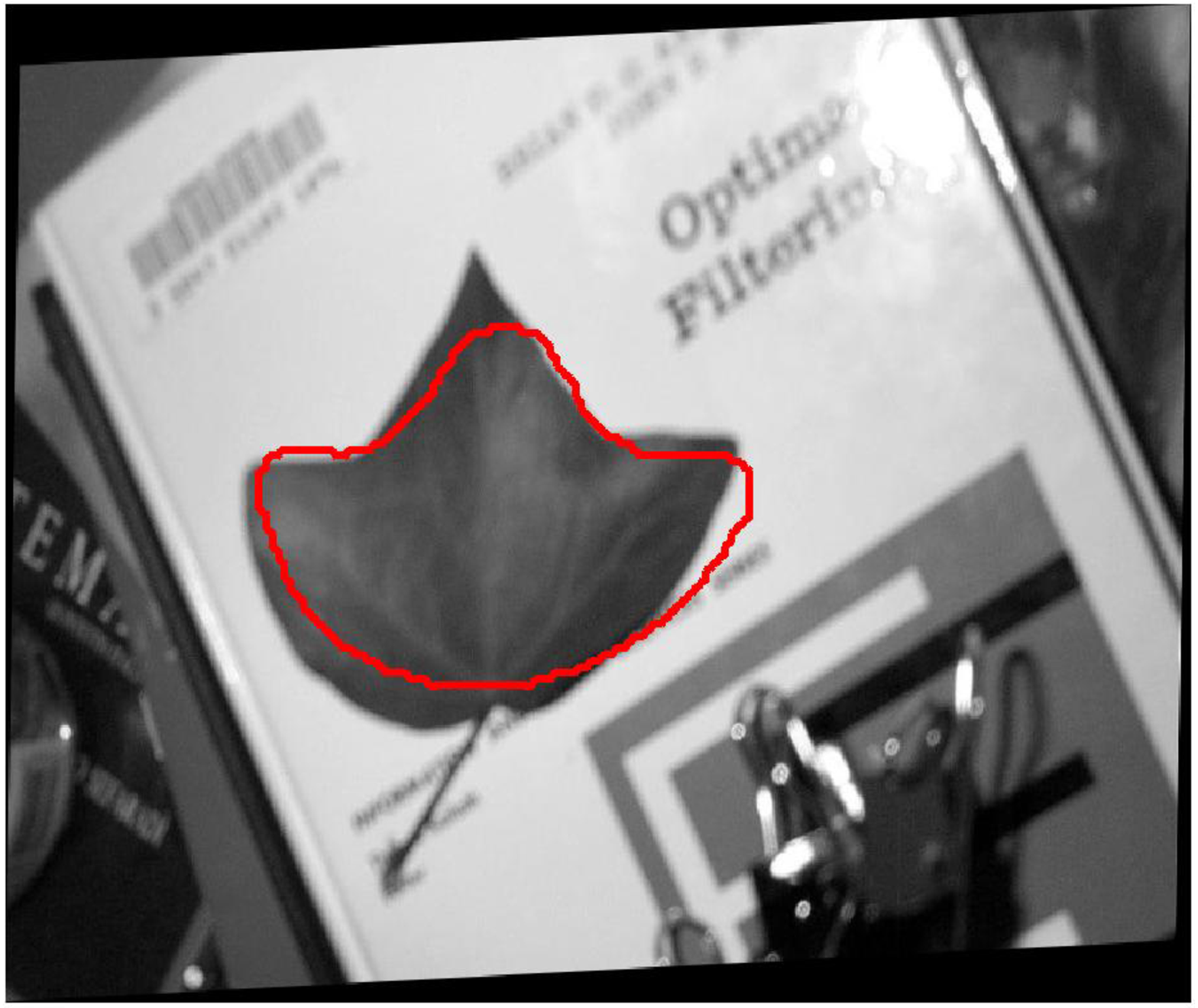}\\ \vspace{-10pt}
  \includegraphics[height=0.8in]{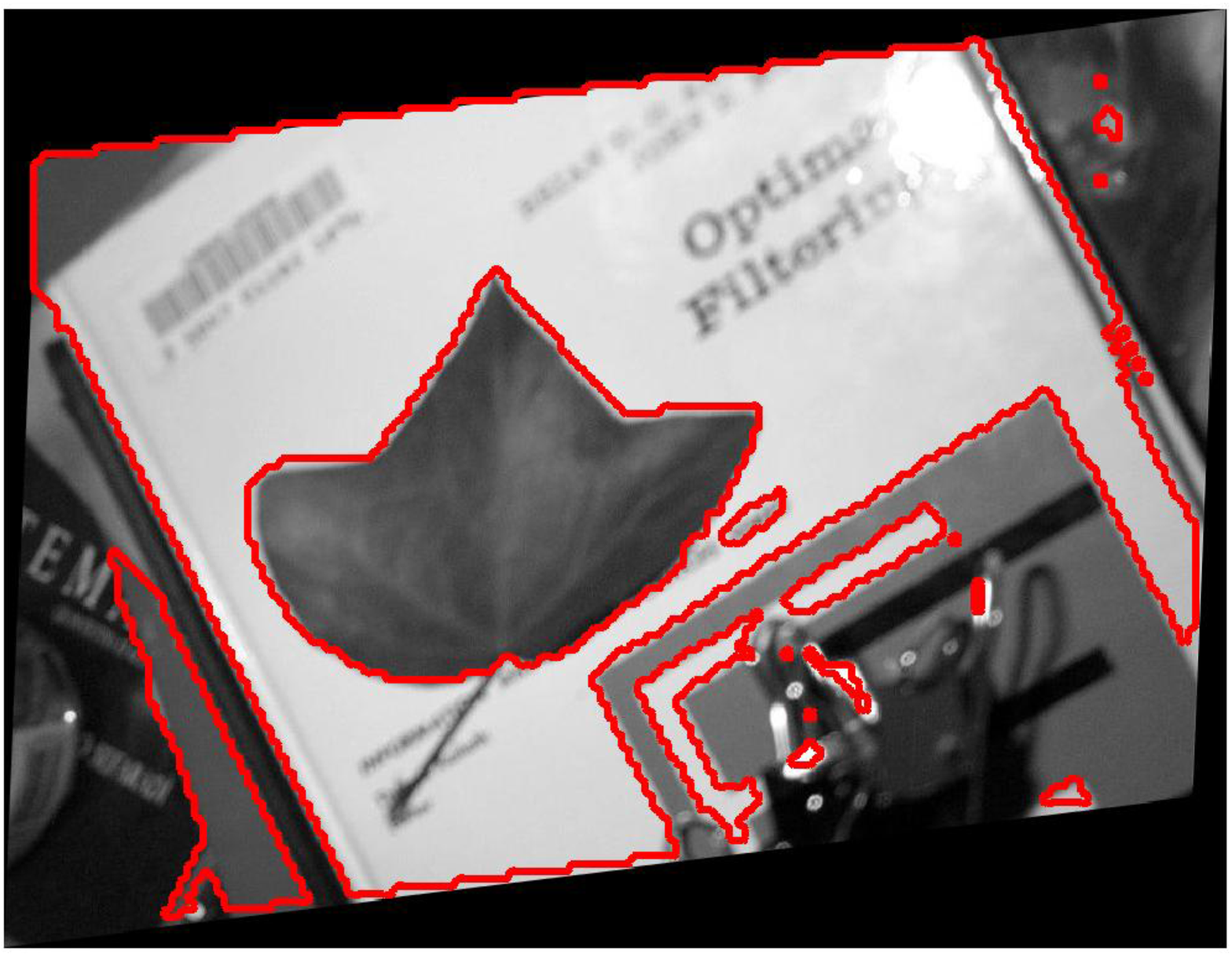}&\hspace{-10pt}
  \includegraphics[height=0.8in]{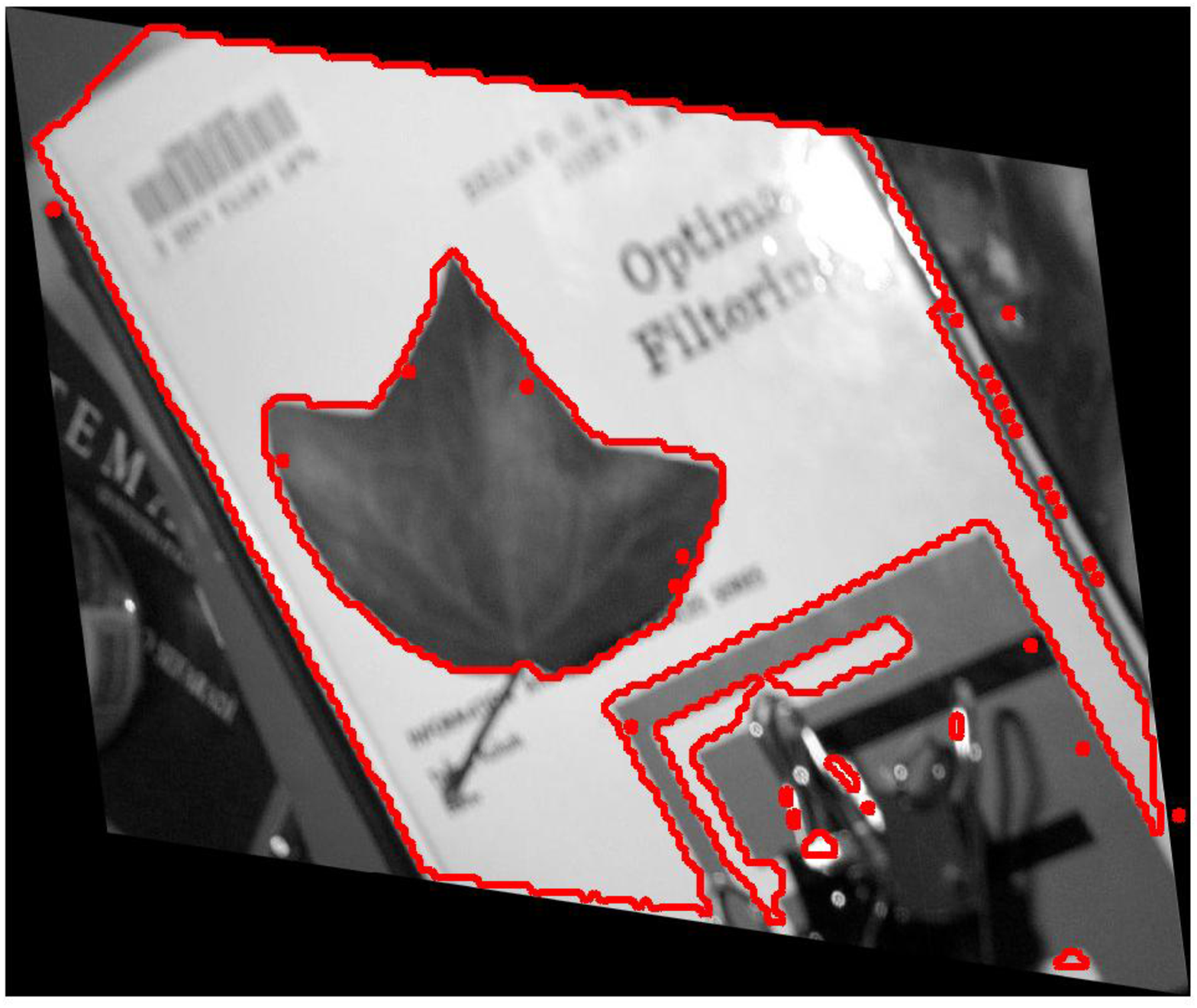}&\hspace{-10pt}
  \includegraphics[height=0.8in]{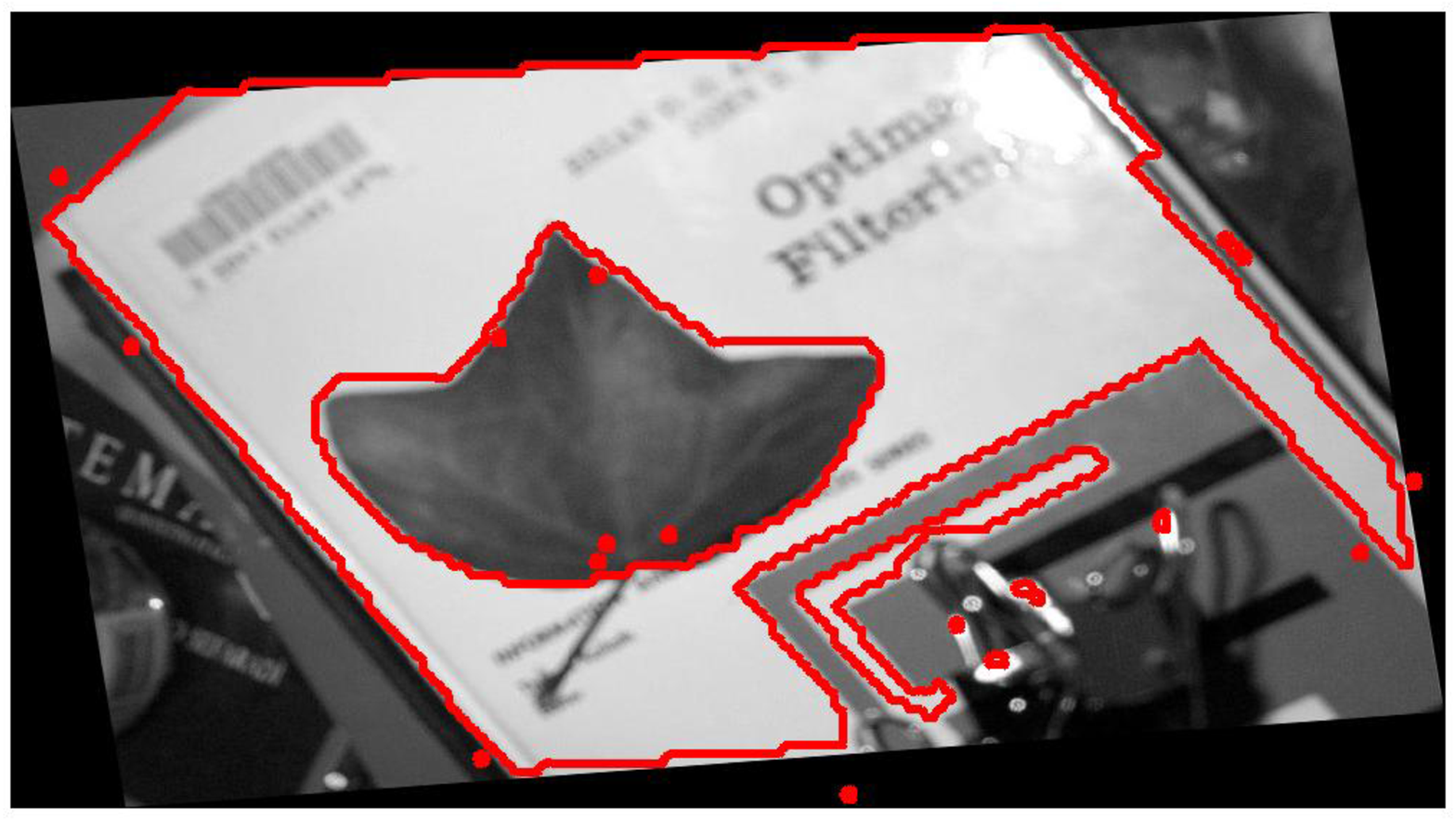}&\hspace{-10pt}
  \includegraphics[height=0.8in]{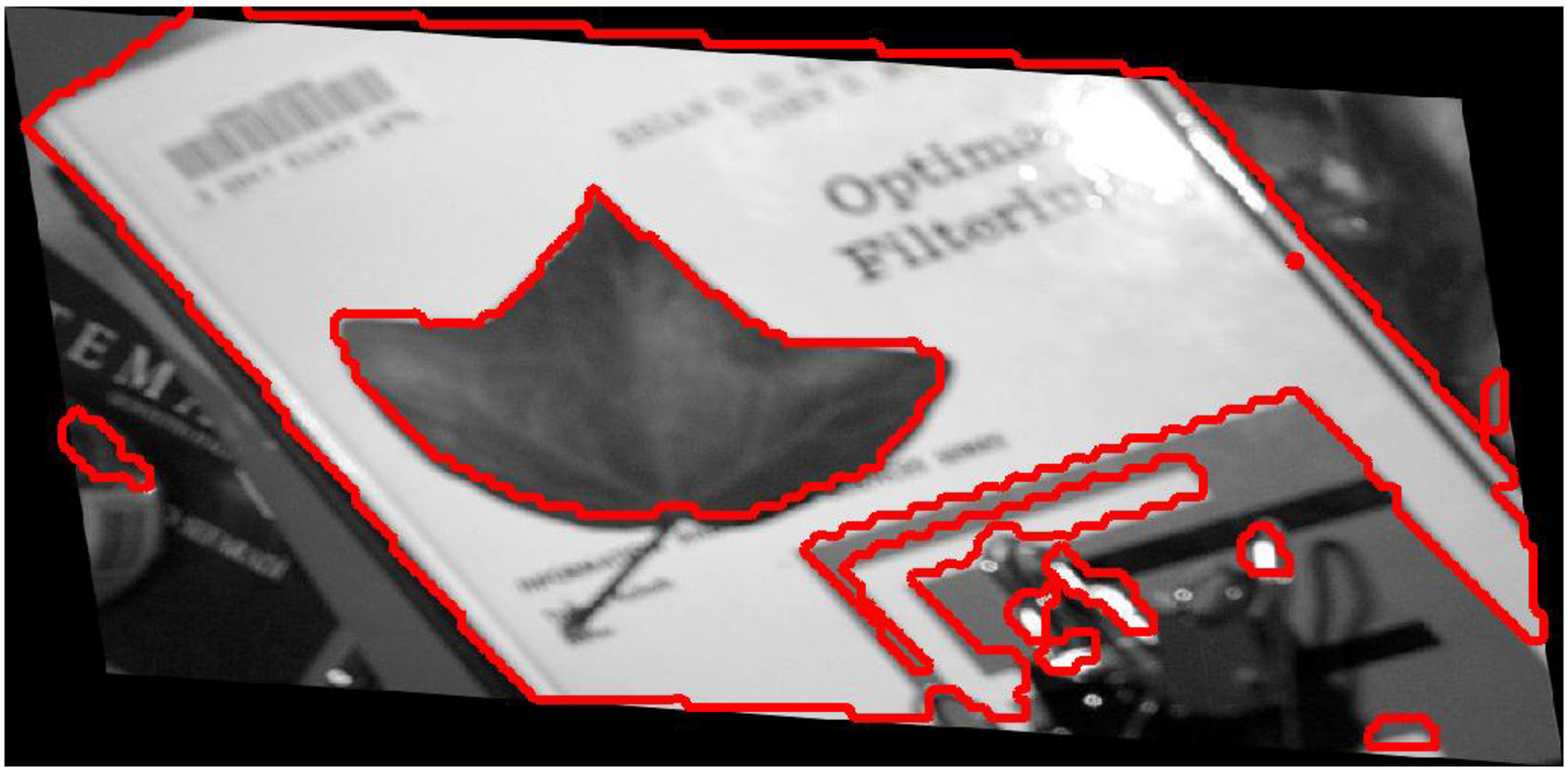}&\hspace{-10pt}
  \includegraphics[height=0.8in]{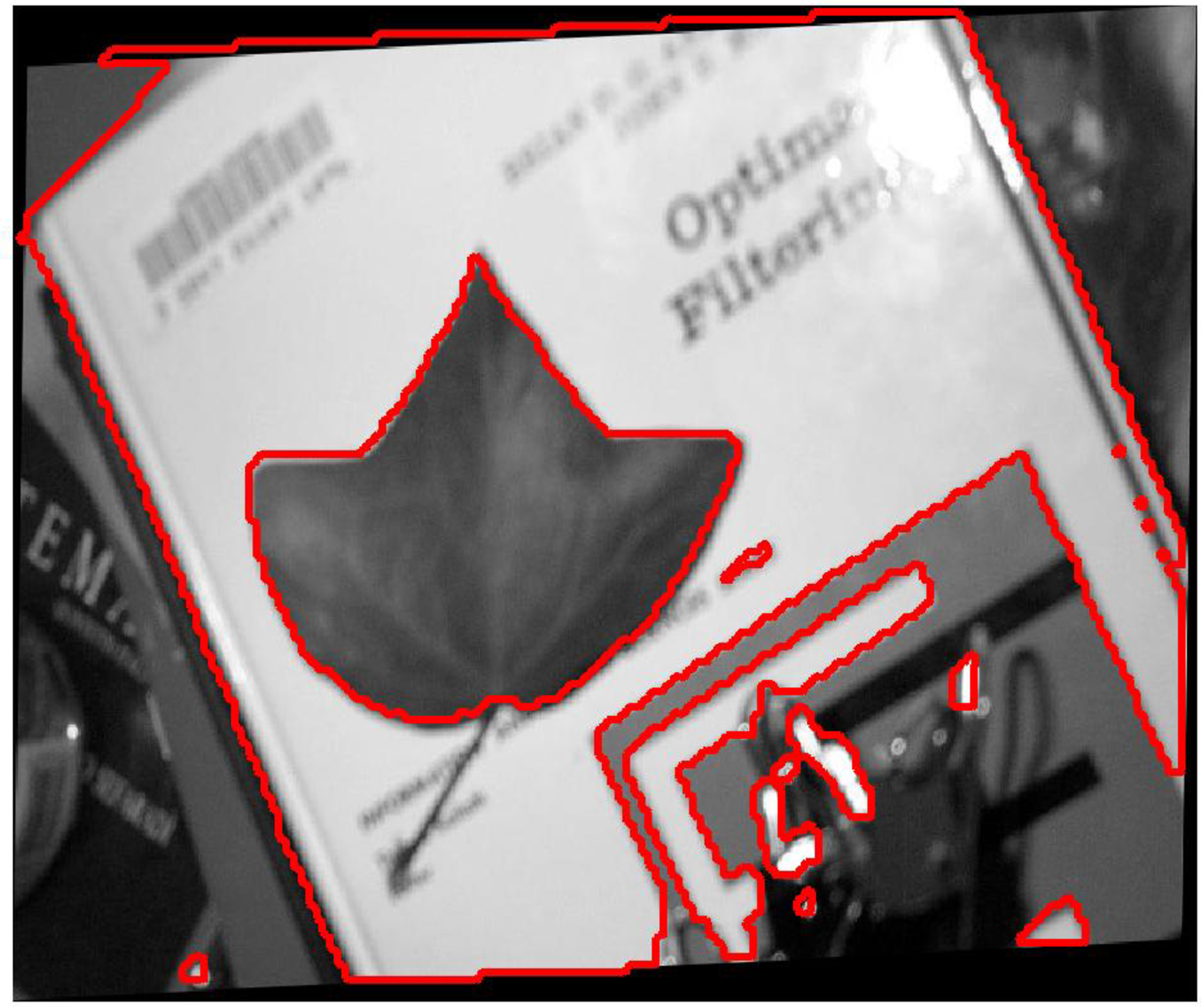}\\ \vspace{-10pt}
  \includegraphics[height=0.8in]{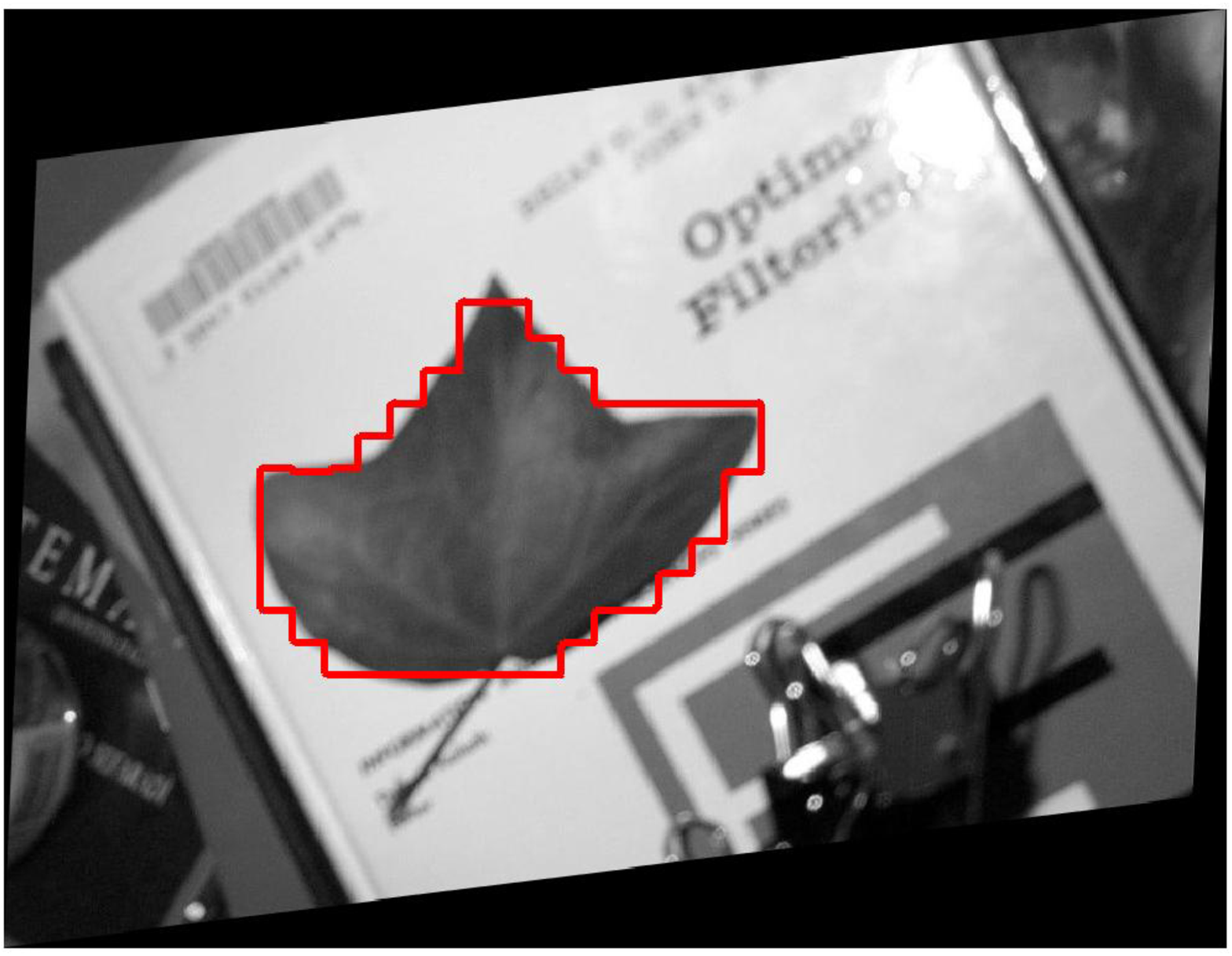}&\hspace{-10pt}
  \includegraphics[height=0.8in]{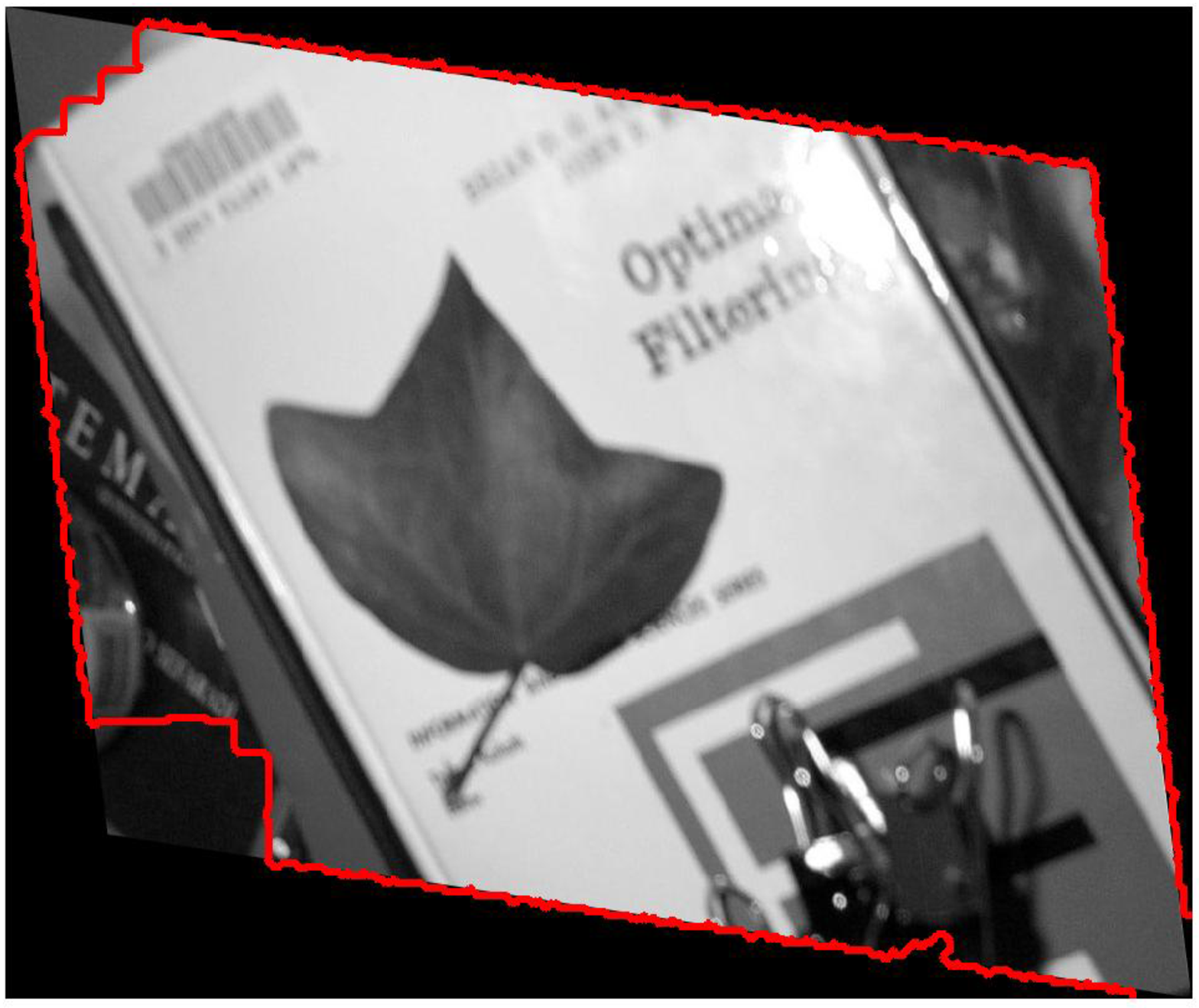}&\hspace{-10pt}
  \includegraphics[height=0.8in]{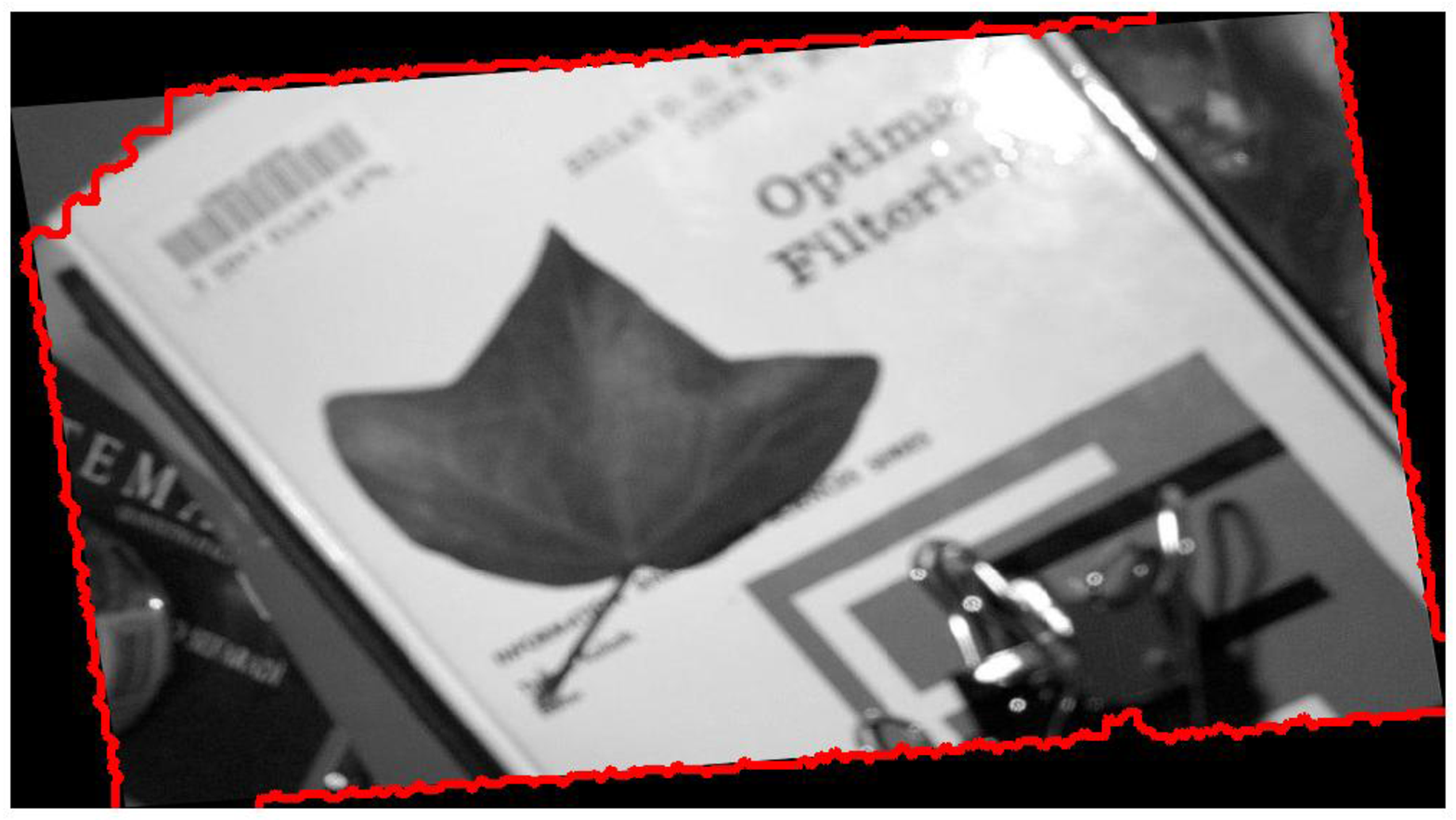}&\hspace{-10pt}
  \includegraphics[height=0.8in]{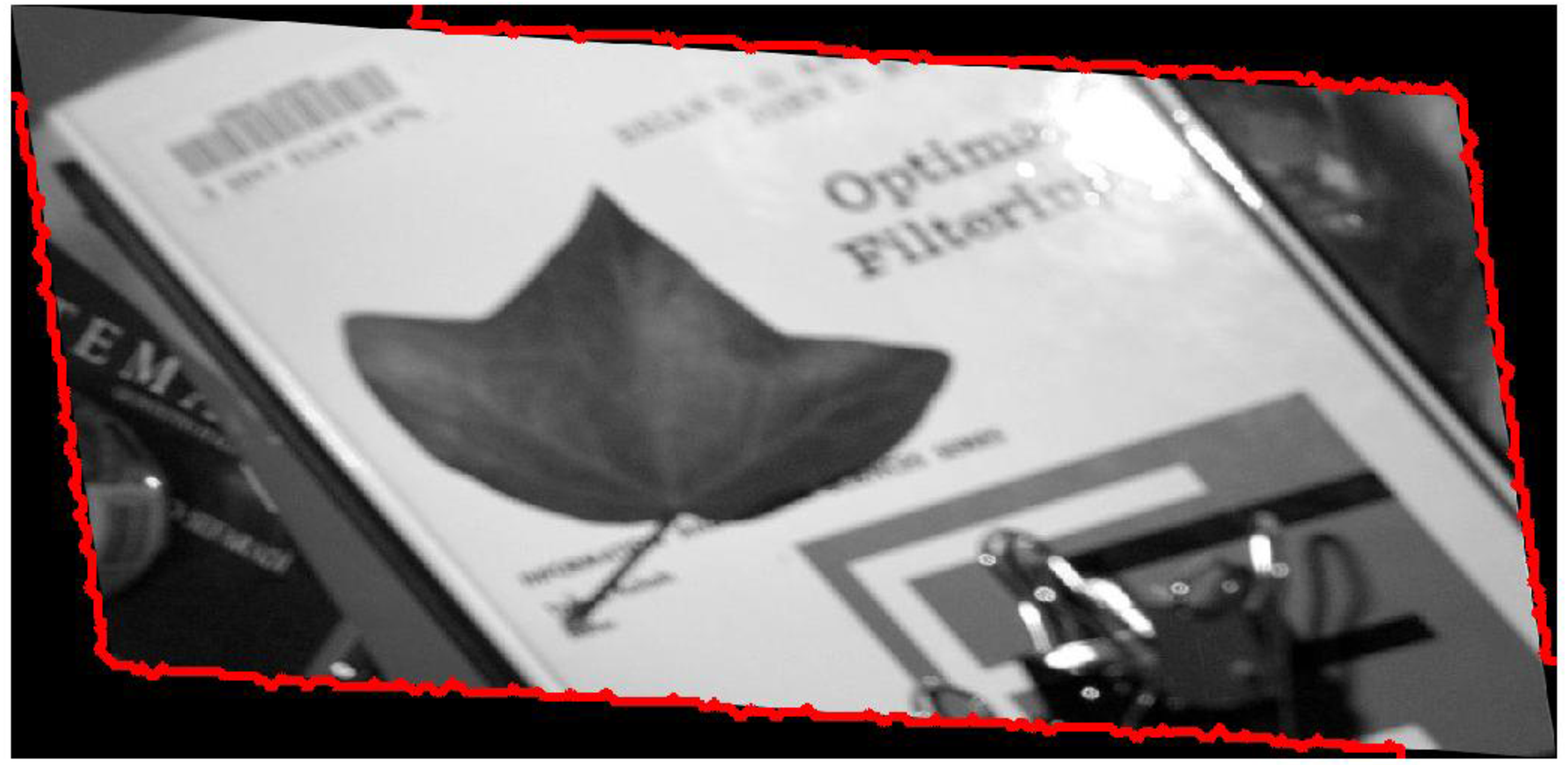}&\hspace{-10pt}
  \includegraphics[height=0.8in]{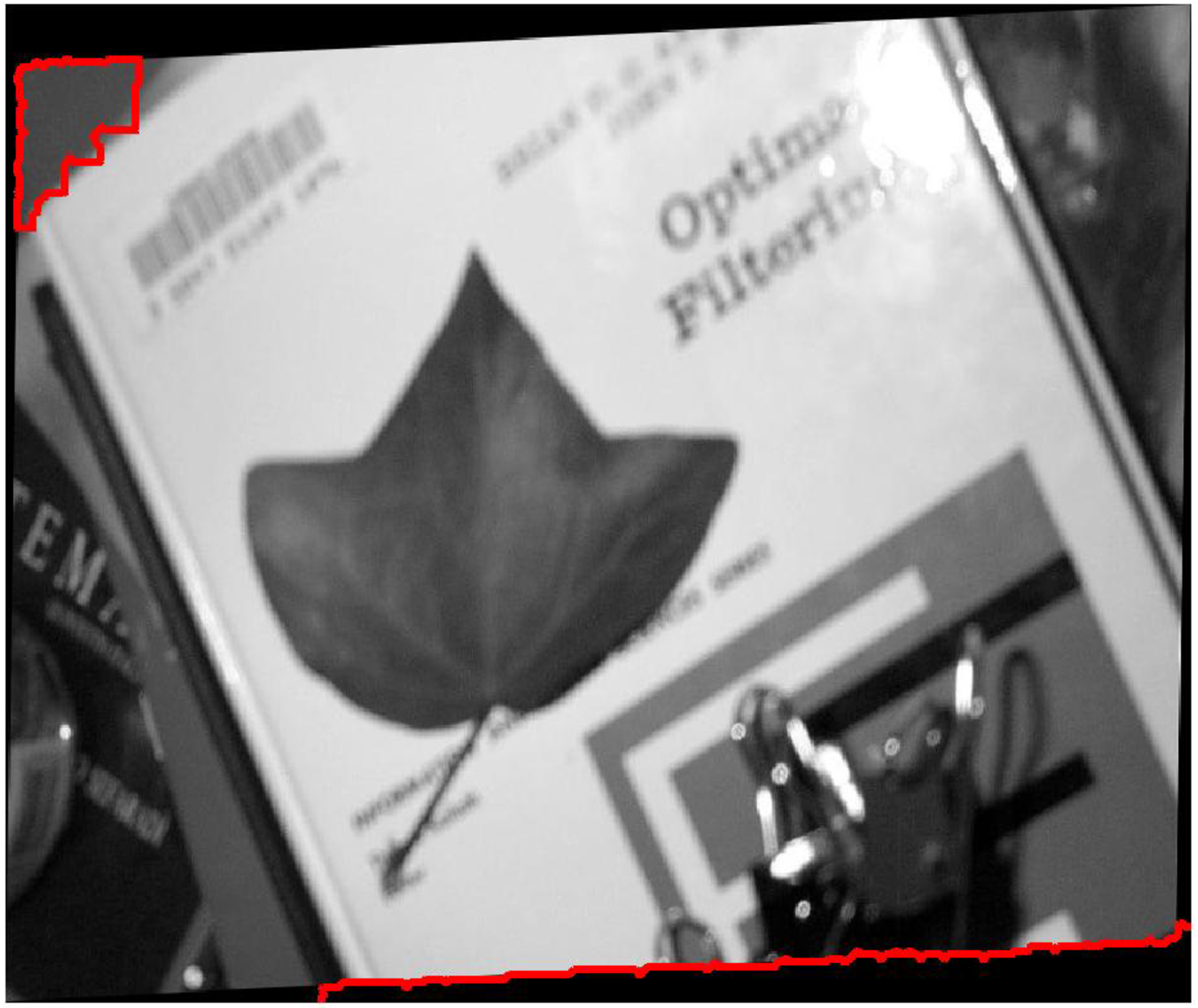}
\end{tabular}
  \caption{Visual comparison. The first row to the $4^{th}$ row shows the results of our method, global search, discriminative clustering based co-segmentation \cite{JouBacPon12MultiClassCoSeg} and submodular optimization \cite{GunheeKimEricXingLiFeiFeiKanade11SOCoSeg}}\label{FIG:visual_cmp}
\end{figure*}

We conduct a quantitative experimental comparison. To eliminate miscellaneous factors in the experiment as many as possible, we use images generated by randomly transformation of the same image which is the leaf image shown in Fig. \ref{FIG:global_good}. We use 50 random affine transformation matrices to generate 50 random images. The object example has been shown in Fig. \ref{FIG:Obj_and_Shape}. The quantitative results are shown in Fig. \ref{FIG:quant_cmp}. We can observe that our method, matching-constrained active contour (MCAC), significantly outperforms the others. Some visual results are also shown in Fig. \ref{FIG:visual_cmp}. We can observe that our method can locate the object boundary satisfactorily. However, the results of global search may deviate from the object of interest because of insufficient information on the object. The co-segmentation methods may produce a lot spurious contours and some times they may also miss the object. This is because the formulation of co-segmentation generally does not include a sensible object model. Our model integrates feature matching, shape prior modeling and object boundary modeling, all of which are practically useful object models.

\subsection{Discussions}
There is a tolerance level $\tau$ in the formulation of the matching-constrained active contour to be determined by the user. We provide some observations to help the users to select their appropriate $\tau$. Since the small $\tau$ may only allow small deviation from the initial solution even if the boundary is not very close to the initial contour. Hence, we may not need to constrain the optimization intensively. In the images for evaluation, the boundary of the objects of interest corresponds to the local optimal solutions to the active contour energy. Hence, we only need the gradient descent given the initial feasible solution produced by optimal matching without compromising to the constraint. The constraint should be more useful during the process of searching for the optimal solution if the boundary is actually not a local optimal solution to the active contour model.

Objects may deform both rigidly and non-rigidly. In this paper, we have considered a general type of deformation, i.e. the affine transformation. The more general non-rigid deformations can be estimated by following the nonrigid shape prior modeling, e.g. \cite{Etyngier07Shapepriors} \cite{Wang09PVCE} \cite{Joshi09BAC}, once the affine shape transformation is known.

Robust detection of the convergence of the gradient descent based method is a research problem in many areas such as optimization, machine learning. In our experiments, we terminate the algorithm if the average change of the affine parameters is less than a threshold or the maximum iteration is reached.

The computational cost depends on the number of selected feature points on the template object. A careful selection of the feature points can reduce the computational cost drastically while preserving the segmentation accuracy.

\section{Conclusion and future work}\label{SEC:Concl}
In this paper, we proposed a novel constrained optimization formulation of active contours. The numerical optimization of this new active contour model leads to an automatic object segmentation algorithm. This work expands the capacity of the conventional active contour approach for object segmentation, and the extension has practical significance in that the conventional semi-automatic framework has been automated.

There are several possible future directions. 1) The shape modeling in our affine-invariant interior-point-to-shape relation can be improved, and the computations can be accelerated by cleverly selecting the feature points. 2) The object matching is based on single template. However, the appearance of the object of interest could vary significantly in different images. Robust object matching is crucial to our method. 3) There could be interesting applications of this method, such as in the training phase of general object recognition tasks.

%In this paper, we propose a novel framework of automatic object segmentation based on active contour, deformable shape modeling, and object matching. The key component is a mathematical relation between the inner points and the shape contour. We show that the affine transformation of the inner points will lead to the same affine transformation of the shape contour in this relation, which we termed the affine invariance property. Based on this affine-invariant relation, we can produce an initial curve for active contours by feature matching and affine point registration. We also require the optimal solution to the segmentation to be the affine transformation of the initial shape contour. Thus, we derive the gradient descent equation for the affine parameters to minimize the active contour energy. We formulate the entire framework as a matching-constrained active contour model. We derive the projected-gradient descent equations for the optimization. Experiments show promising results on real images.

%Our method is based on reliable object matching.
%\include{App}

%%%%%%%%%%%%%%%%%%%%%%%%%%%%%%%%%%%%%%%%%%%%%
%\newpage
%\section*{APPENDIX}  % use *-form to suppress numbering
\renewcommand\thesection{APPENDIX-\Alph{section}}
\setcounter{section}{0}
\renewcommand{\theequation}{\Alph{section}-\arabic{equation}}
  % redefine the command that creates the equation no.
%\appendix
\setcounter{equation}{0}  % reset counter
\section{Gradient descent equations for shape training}\label{APD-A}
The gradient descent equations for learning the parameters $\{\alpha_i\}$ and $\beta$ in the shape decision function in Eq. (\ref{EQ:RBF_H_e}) are as follows:
\begin{equation*}
\begin{split}
{\partial\alpha_i\over\partial t} =& -\int_\mathcal{D} 2(H_e-H_o)\psi(\|\vec{z}-\vec{p}_i\|) H'dxdy,\\
&i=1,2,\ldots,N,\\
\end{split}
\end{equation*}
\begin{equation*}
{\partial\beta\over\partial t} = -\int_\mathcal{D} 2(H_e-H_o) H'dxdy,
\end{equation*}
where $H'$ is the first order derivative of $H$. The trained contour curve is defined by Eq. (\ref{EQ:Def_C}) in which $H_o$ is replaced by the trained $H_e$.

\setcounter{equation}{0}
\section{Proof of Proposition \ref{Prop:NAffInV}}\label{APD-B}
\begin{IEEEproof}
Let $\vec{a} = \vec{z}_c-\vec{p}_i$, we have that $\mathbf{A}\vec{a}=\mathbf{A}\vec{z}_c -\mathbf{A}\vec{p}_i$. Our objective is to show $\|\mathbf{A}\vec{a}\|^2\neq\|\vec{a}\|^2$ in general. Since $\|\mathbf{A}\vec{a}\|^2-\|\vec{a}\|^2 = \vec{a}^T(\mathbf{A}^T\mathbf{A}-I)\vec{a}$, and $\mathbf{A}^T\mathbf{A}-I$ is a nonzero symmetric matrix. We can write
$$
\|\mathbf{A}\vec{a}\|^2-\|\vec{a}\|^2 = \vec{b}^TD\vec{b},
$$
where $\mathbf{A}^T\mathbf{A}-I = U^TDU$ and $\vec{b} = U\vec{a}$. This is a result of eigen-decomposition.

Obviously, there exists one vector $\vec{b}$ such that the above does not equal $0$ for any nonzero $\mathbf{D}$. By almost arbitrary scaling of $\vec{b}$, we obtain infinitely many such vectors, which completes the proof.
\end{IEEEproof}

%\subsection*{Update equations for solving the joint matching-registration model in (\ref{EQ:JF_match_regi1})}
%The gradients of the energy function in (\ref{EQ:JF_match_regi1}) are as follows.
%\begin{equation}\label{EQ:FixPT_A}
%\nabla_{\mathbf{A}} E\big(\mathbf{A},\vec{b}\big) = -\sum_{ij} \hat{c}_{ij} g_{ij}(\mathbf{A}\vec{p_i}+\vec{b}-\vec{q^t_j})\vec{p_i}^T
%\end{equation}
%\begin{equation}\label{EQ:FixPT_b}
%\nabla_{\vec{b}} E\big(\mathbf{A},\vec{b}\big) = -\sum_{ij} \hat{c}_{ij} g_{ij}(\mathbf{A}\vec{p_i}+\vec{b}-\vec{q^t_j})
%\end{equation}
%where $\hat{c}_{ij} = e^{-c_{ij}}$. We adopt the Gaussian function to approximate the Dirac delta. Thus, $g_{ij}=\delta_{\epsilon}(r_{ij})$. We also normalize the Gaussian functions according to the constraint $\sum_{i}\delta_\epsilon(r_{ij})=1$.
%
%By letting (\ref{EQ:FixPT_A}) and (\ref{EQ:FixPT_b}) to be zero, we obtain the following system of equations.
%\begin{equation}\label{EQ:FXP_A}
%\mathbf{A} = \left(\sum_{ij} \hat{c}_{ij} g_{ij}(\vec{q^t_j}-\vec{b})\vec{p_i}^T\right)\left(\sum_{ij} \hat{c}_{ij} g_{ij}\vec{p_i}\vec{p_i}^T\right)^{^{-1}}
%\end{equation}
%\begin{equation}\label{EQ:FXP_b}
%\vec{b}= {\sum_{ij} \hat{c}_{ij} g_{ij}(\mathbf{A}\vec{p_i}-\vec{q^t_j})\over\sum_{ij} \hat{c}_{ij} g_{ij}}
%\end{equation}
%%\sum_{ij}\mathbf{C}_{ij}\delta_\epsilon(\hat{\vec{p^t_i}}-\vec{p^t_j})
%Some few fixed-point iterations of the above can yield a converged solution with the initial $\mathbf{A}$ and $\vec{b}$ given by the state-of-the-art-matching and registration.

\setcounter{equation}{0}
\section{Derivations of Eq. (\ref{EQ:dE/dt_dJ/dt})}\label{APD-C}
In the following, we present the derivations of Eq. (\ref{EQ:dE/dt_dJ/dt}).
\begin{equation*}
\begin{split}
{dE\over dt}=&-\nabla_{\mathbf{A}}E^T \left(\nabla_{\mathbf{A}}J-\left\langle\nabla_{\mathbf{A}}J,{\nabla_{\mathbf{A}}E\over\|\nabla_{\mathbf{A}}E\|}\right\rangle{\nabla_{\mathbf{A}}E\over\|\nabla_{\mathbf{A}}E\|}\right)\\
&-\nabla_{\vec{b}}E^T \left(\nabla_{\vec{b}}J-\left\langle\nabla_{\vec{b}}J,{\nabla_{\vec{b}}E\over\|\nabla_{\vec{b}}E\|}\right\rangle{\nabla_{\vec{b}}E\over\|\nabla_{\vec{b}}E\|}\right)\\
=&-\langle\nabla_{\mathbf{A}}J,{\nabla_{\mathbf{A}}E}\rangle+\langle\nabla_{\mathbf{A}}J,{\nabla_{\mathbf{A}}E}\rangle\\
&-\langle\nabla_{\vec{b}}J,{\nabla_{\vec{b}}E}\rangle+\langle\nabla_{\vec{b}}J,{\nabla_{\vec{b}}E}\rangle=0
\end{split}
\end{equation*}

\begin{equation*}
\begin{split}
{dJ\over dt}=&-\nabla_{\mathbf{A}}J^T \left(\nabla_{\mathbf{A}}J-\left\langle\nabla_{\mathbf{A}}J,{\nabla_{\mathbf{A}}E\over\|\nabla_{\mathbf{A}}E\|}\right\rangle{\nabla_{\mathbf{A}}E\over\|\nabla_{\mathbf{A}}E\|}\right)\\
&-\nabla_{\vec{b}}J^T \left(\nabla_{\vec{b}}J-\left\langle\nabla_{\vec{b}}J,{\nabla_{\vec{b}}E\over\|\nabla_{\vec{b}}E\|}\right\rangle{\nabla_{\vec{b}}E\over\|\nabla_{\vec{b}}E\|}\right)\\
=&-\left(\|\nabla_{\mathbf{A}}J\|^2-\left\langle\nabla_{\mathbf{A}}J,{\nabla_{\mathbf{A}}E\over\|\nabla_{\mathbf{A}}E\|}\right\rangle^2\right)\\
&-\left(\|\nabla_{\vec{b}}J\|^2-\left\langle\nabla_{\vec{b}}J,{\nabla_{\vec{b}}E\over\|\nabla_{\vec{b}}E\|}\right\rangle^2\right)\leq0
\end{split}
\end{equation*}

\setcounter{equation}{0}
\section{Derivation of Eqs. (\ref{EQ:OBJ_GD_invA}) and (\ref{EQ:OBJ_GD_b})}\label{APD-D}
Our derivation is based on the following equality for minimizing general active contour energy.
\begin{equation}\label{EQ:CE}
{\partial C\over\partial t} = -\nabla J = \alpha(p)\vec{N}.
\end{equation}
Note that the equality holds true for geometric active contours.

The differential of the shape decision function $\phi_S$ at the implicit contour leads to the following.
\begin{equation}\label{EQ:Diff_phi_S}
\begin{split}
&\left.{\partial \phi_S(\vec{z}(t),\{\mathbf{A}^{-1}(t),\vec{b}(t)\})\over\partial t}\right|_{\vec{z}\in\{\vec{z}|\phi_S=0\}}\\
&=\nabla_{\vec{z}}\phi_S\cdot{\partial C\over\partial t}+{D\phi_S\over D\mathbf{A}^{-1}}\cdot{d \mathbf{A}^{-1}\over dt}+{D\phi_S\over D \vec{b}}\cdot{d\vec{b}\over dt}=0.
\end{split}
\end{equation}
Substituting Eq. (\ref{EQ:CE}) into the above, we obtain the expression for $\alpha$.
\begin{equation}\label{EQ:alpha_Diff_phi_S}
\alpha = -{{D\phi_S\over D\mathbf{A}^{-1}}\cdot{d \mathbf{A}^{-1}\over dt}+{D\phi_S\over D \vec{b}}\cdot{d\vec{b}\over dt}\over\langle\vec{N},\nabla_{\vec{z}}\phi_S\rangle}.
\end{equation}
Substituting the (\ref{EQ:alpha_Diff_phi_S}) into (\ref{EQ:CE}) we obtain the curve evolution as follows.
%
%Substituting the implicit contour defined by $\phi_S$ into (\ref{EQ:CE}) we may obtain the following.
\begin{equation}\label{EQ:CE_phi_S}
{\partial C\over\partial t} = -{{D\phi_S\over D\mathbf{A}^{-1}}\cdot{d\mathbf{A}^{-1}\over dt}+{D\phi_S\over D \vec{b}}\cdot{d\vec{b}\over dt}\over\langle\vec{N},\nabla_{\vec{z}}\phi_S\rangle}\vec{N}.
\end{equation}

To minimize a general active contour energy $J(C)$, we require the derivative of $J(C)$ to be non-positive as
\begin{equation}\label{EQ:Diff_J1}
{dJ\over dt}= \left\langle\nabla J, {\partial C\over \partial t}\right\rangle_p\leq 0,
\end{equation}
where $\langle f,g\rangle_p=\int \langle f(p), g(p)\rangle dp$ is the inner product of two vector functions in which $\langle,\rangle$ is the vector inner product.
Substituting (\ref{EQ:CE_phi_S}) into (\ref{EQ:Diff_J1}), considering $\vec{N}=\nabla_{\vec{z}}\phi_S$, we obtain the following:
\begin{equation}\label{EQ:Diff_J2}
\begin{split}
{dJ\over dt} &= \left\langle\nabla J, -{{D\phi_S\over D\mathbf{A}^{-1}}\cdot{d\mathbf{A}^{-1}\over dt}+{D\phi_S\over D \vec{b}}\cdot{d\vec{b}\over dt}\over\langle\vec{N},\nabla_{\vec{z}}\phi_S\rangle}\vec{N}\right\rangle_p\\
&= \left\langle\nabla_{\mathbf{A}^{-1}}J,{d\mathbf{A}^{-1}\over dt}\right\rangle_p+\left\langle\nabla_{\vec{b}}J,{d\vec{b}\over dt}\right\rangle_p,
\end{split}
\end{equation}
where
\begin{equation}
\nabla_{\mathbf{A}^-1}J = -{\nabla J^T\vec{N}\over\vec{N}^T\nabla_{\vec{z}}\phi_S}{D\phi_S\over D\mathbf{A}^{-1}},
\end{equation}
\begin{equation}
{\nabla_{\vec{b}}J} = -{\nabla J^T\vec{N}\over\vec{N}^T\nabla_{\vec{z}}\phi_S}{D\phi_S\over D\vec{b}},
\end{equation}
which gives Eqs. (\ref{EQ:OBJ_GD_invA}) and (\ref{EQ:OBJ_GD_b}). In the gradient descent process, we can set ${d\mathbf{A}^{-1}\over dt}=-\nabla_{\mathbf{A}^{-1}}J$ and ${d\vec{b}\over dt}=-\nabla_{\vec{b}}J$.
%The explicit forms of $\nabla_{\mathbf{A}^{-1}}J$ and $\nabla_{\vec{b}}J$ can be written separately by rewriting , leading to Eqs. (\ref{EQ:OBJ_GD_invA}) and (\ref{EQ:OBJ_GD_b}). To validate the two equations, we substitute Eqs. (\ref{EQ:OBJ_GD_invA}) and (\ref{EQ:OBJ_GD_b}) into (\ref{EQ:Diff_J2}),  and we obtain the following:
%\begin{equation}\label{EQ:Diff_J3}
%\begin{split}
%{dJ\over dt} =& -\left\langle\nabla J, {{{D\phi_S\over D\mathbf{A}^{-1}}\cdot {D\phi_S\over D\mathbf{A}^{-1}}\nabla J^T\vec{N} }\over(\vec{N}^T\nabla_{\vec{z}}\phi_S)^2}\vec{N}\right\rangle_p-\left\langle\nabla J, {{{D\phi_S\over D\vec{b}}\cdot{D\phi_S\over D\vec{b}} \nabla J^T\vec{N} }\over(\vec{N}^T\nabla_{\vec{z}}\phi_S)^2}\vec{N}\right\rangle_p\\
%=&-\left\langle\nabla J,(\nabla J^T\vec{N}) {\left\|{D\phi_S\over D\mathbf{A}^{-1}}\right\|^2\over(\vec{N}^T\nabla_{\vec{z}}\phi_S)^2}\vec{N}\right\rangle_p-\left\langle\nabla J, (\nabla J^T\vec{N}){\left\|{D\phi_S\over D\vec{b}}\right\|^2\over(\vec{N}^T\nabla_{\vec{z}}\phi_S)^2}\vec{N}\right\rangle_p\\
%=&-\bigints_{C}\left({(\nabla J^T\vec{N})\left\|{D\phi_S\over D\mathbf{A}^{-1}}\right\|\over(\vec{N}^T\nabla_{\vec{z}}\phi_S)}\right)^2dp-\bigints_{C}\left({(\nabla J^T\vec{N})\left\|{D\phi_S\over D\vec{b}}\right\|\over(\vec{N}^T\nabla_{\vec{z}}\phi_S)}\right)^2dp.
%\end{split}
%\end{equation}
%Obviously, ${dJ\over dt}\leq0$.

%%%%%%%%%%%%%%%%%%%%%%%%%%%%%%%%%%%%%%%%%%%%%

{
\bibliographystyle{IEEEtran}
\bibliography{LevelSetActiveContours,MRFseg,MatchingDetection,Matching}

\begin{thebibliography}{10}
\providecommand{\url}[1]{#1}
\csname url@samestyle\endcsname
\providecommand{\newblock}{\relax}
\providecommand{\bibinfo}[2]{#2}
\providecommand{\BIBentrySTDinterwordspacing}{\spaceskip=0pt\relax}
\providecommand{\BIBentryALTinterwordstretchfactor}{4}
\providecommand{\BIBentryALTinterwordspacing}{\spaceskip=\fontdimen2\font plus
\BIBentryALTinterwordstretchfactor\fontdimen3\font minus
  \fontdimen4\font\relax}
\providecommand{\BIBforeignlanguage}[2]{{%
\expandafter\ifx\csname l@#1\endcsname\relax
\typeout{** WARNING: IEEEtran.bst: No hyphenation pattern has been}%
\typeout{** loaded for the language `#1'. Using the pattern for}%
\typeout{** the default language instead.}%
\else
\language=\csname l@#1\endcsname
\fi
#2}}
\providecommand{\BIBdecl}{\relax}
\BIBdecl

\bibitem{caselles97GAC}
V.~Caselles, R.~Kimmel, and G.~Sapiro, ``Geodesic active contour,''
  \emph{International Journal of Computer Vision}, vol.~22, no.~1, pp. 61--79,
  1997.

\bibitem{ChanVese01ActiveCon}
T.~Chan and L.~Vese, ``Active contours without edges,'' \emph{IEEE Transactions
  on Image Processing}, vol.~10, no.~2, pp. 266--277, 2001.

\bibitem{GVF98}
C.~Xu and J.~L. Prince, ``Snakes, shapes, and gradient vector flow,''
  \emph{IEEE Transactions on Image Processing}, vol.~7, no.~3, pp. 359--369,
  1998.

\bibitem{GVFGAC04}
N.~Paragios, O.~Mellina-Gottardo, and V.~Ramesh, ``Gradient vector flow fast
  geometric active contours,'' \emph{IEEE Transactions on Pattern Analysis and
  Machine Intelligence}, vol.~26, no.~3, pp. 402--407, 2004.

\bibitem{LiSnake05Split}
C.~Li, J.~Liu, and M.~D. Fox, ``Segmentation of edge preserving gradient vector
  flow: An approach toward automatically initializing and splitting of
  snakes,'' in \emph{IEEE Computer Society Conference on Computer Vision and
  Pattern Recognition}, 2005.

\bibitem{Xie08MAC}
X.~Xie and M.~Mirmehdi, ``Mac: Magnetostatic active contour model,'' \emph{IEEE
  Transactions on Pattern Analysis and Machine Intelligence}, vol.~30, no.~4,
  pp. 632--646, 2008.

\bibitem{Wang2008}
J.~Wang, K.~L. Chan, and Y.~Wang, ``On the stationary solution of {P}{D}{E}
  based curve evolution,'' in \emph{Proceedings of the 19th British Machine
  Vision Conference}, 2008.

\bibitem{Li08AutoInitAC}
B.~Li and S.~T. Acton, ``Automatic active model initialization via poisson
  inverse gradient,'' \emph{IEEE Transactions on Image Processing}, vol.~17,
  no.~8, pp. 1406--1420, 2008.

\bibitem{Cremers08GlobalSP}
D.~Cremers, F.~R. Schmidt, and F.~Barthel, ``Shape priors in variational image
  segmentation: Convexity, lipschitz continuity and globally optimal
  solutions,'' in \emph{IEEE Conference on Computer Vision and Pattern
  Recognition}, 2008.

\bibitem{Schoenemann10}
T.~Schoenemann and D.~Cremers, ``A combinatorial solution for model-based image
  segmentation and real-time tracking,'' \emph{IEEE Transactions on Pattern
  Analysis and Machine Intelligence}, vol.~32, pp. 1153--1164, 2010.

\bibitem{Schoenemann07}
------, ``Globally optimal image segmentation with an elastic shape prior,'' in
  \emph{Procedings of the 12th International Conference on Computer Vision},
  2007.

\bibitem{Lawler1966MRC}
E.~L. Lawler, ``Optimal cycles in doubly weighted linear graphs,'' in
  \emph{Theory of Graphs: International Symposium}, 1966, pp. 209--213.

\bibitem{Viola04RealTimeFD}
P.~Viola and M.~J. Jones, ``Robust real-time face detection,''
  \emph{International Journal of Computer Vision}, vol.~57, pp. 137--154, May
  2004.

\bibitem{Dalal05HOG}
N.~Dalal and B.~Triggs, ``Histograms of oriented gradients for human
  detection,'' in \emph{IEEE Computer Society Conference on Computer Vision and
  Pattern Recognition}, 2005.

\bibitem{Lowe04SIFT}
D.~G. Lowe, ``Distinctive image features from scale-invariant keypoints,''
  \emph{International Journal of Computer Vision}, vol.~60, pp. 91--110, 2004.

\bibitem{Jiang09ScaRotMat}
H.~Jiang and S.~Yu, ``Linear solution to scale and rotation invariant object
  matching,'' in \emph{IEEE Computer Society Conference on Computer Vision and
  Pattern Recognition}, 2009.

\bibitem{Li10ObjMatchLocalAffine}
H.~Li, E.~Kim, X.~Huang, and L.~He, ``Object matching with a locally
  affine-invariant constraint.'' in \emph{IEEE Computer Society Conference on
  Computer Vision and Pattern Recognition}, 2010.

\bibitem{OsherSethian88Fronts}
S.~Osher and J.~A. Sethian, ``Fronts propagating with curvature-dependent
  speed: Algorithms based on {H}amilton-{J}acobi formulations,'' \emph{Journal
  of Computational Physics}, vol.~79, pp. 12--49, 1988.

\bibitem{Luenberger08BookLP&NLP}
D.~G. Luenberger and Y.~Ye, \emph{Linear and Nonlinear Programming, Second
  Edition}, 3rd~ed.\hskip 1em plus 0.5em minus 0.4em\relax Springer, 2008.

\bibitem{Jiang07LinMatch}
H.~Jiang, M.~S. Drew, and Z.-N. Li, ``Matching by linear programming and
  successive convexification,'' \emph{TPAMI}, vol.~29, pp. 959--975, June 2007.

\bibitem{Ke04PCASIFT}
Y.~Ke and R.~Sukthankar, ``Pca-sift: A more distinctive representation for
  local image descriptors,'' in \emph{IEEE Computer Society Conference on
  Computer Vision and Pattern Recognition}.\hskip 1em plus 0.5em minus
  0.4em\relax Los Alamitos, CA, USA: IEEE Computer Society, 2004.

\bibitem{JouBacPon12MultiClassCoSeg}
A.~Joulin, F.~Bach, and J.~Ponce, ``Multi-class cosegmentation,'' in \emph{IEEE
  Computer Sociaty Conference on Computer Vision and Pattern Recognition},
  2012.

\bibitem{GunheeKimEricXingLiFeiFeiKanade11SOCoSeg}
G.~Kim, E.~P. Xing, L.~Fei-Fei, and T.~Kanade, ``Distributed cosegmentation via
  submodular optimization on anisotropic diffusion,'' in \emph{Proceedings of
  the 2011 International Conference on Computer Vision}, 2011, pp. 169--176.

\bibitem{Etyngier07Shapepriors}
P.~Etyngier, F.~Segonne, and R.~Keriven, ``Shape priors using manifold learning
  techniques,'' in \emph{Proceedings of the Eleventh IEEE International
  Conference on Computer Vision}, 2007.

\bibitem{Wang09PVCE}
J.~Wang and K.~L. Chan, ``Shape evolution for rigid and nonrigid shape
  registration and recovery,'' in \emph{IEEE Computer Society Conference on
  Computer Vision and Pattern Recognition}, 2009.

\bibitem{Joshi09BAC}
S.~H. Joshi and A.~Srivastava, ``Intrinsic bayesian active contours for
  extraction of object boundaries in images,'' \emph{International Journal of
  Computer Vision}, vol.~81, no.~3, pp. 331--355, 2009.

\end{thebibliography}
}

%\begin{IEEEbiography}{Junyan Wang}
%\end{IEEEbiography}
%
%\begin{IEEEbiography}{Kap Luk Chan}
%\end{IEEEbiography}

%\newpage


%\vfill


\end{document}